\documentclass[conference]{IEEEtran}

\usepackage{cite}
\usepackage{boxedminipage}
\usepackage{multirow}
\usepackage{macros}
\usepackage{times}
\usepackage{graphicx}
\usepackage{epsfig}
\usepackage{color}
\usepackage{array}
\usepackage{tabularx}
\usepackage{algorithm}
\usepackage{algorithmic}
\usepackage{subfig}
\usepackage{amsmath,amssymb,amsfonts,amsthm}
\usepackage{float}
\usepackage{adjustbox}
\usepackage[
pass,
]{geometry}

\newcommand{\bx}{\mathbf{x}}

\newcommand{\bz}{\mathbf{z}}

\newcommand{\bbR}{\mathbb{R}}

\newcommand{\bdelta}{\boldsymbol{\delta}}

\begin{document}

\title{On the Limitation of MagNet Defense against $L_1$-based Adversarial Examples}

\author{ Pei-Hsuan Lu$^*$, Pin-Yu Chen$^\dagger$, Kang-Cheng Chen$^\spadesuit$, and Chia-Mu Yu$^*$ \\
		$^*$National Chung Hsing University, Taiwan\\
    $^\dagger$IBM Research, USA\\		
		$^\spadesuit$Yuan Ze University, Taiwan}
\maketitle
\begin{abstract}\vspace{-0.0cm}
In recent years, defending adversarial perturbations to natural examples in order to build robust machine learning models trained by deep neural networks (DNNs) has become an emerging research field in the conjunction of deep learning and security. In particular, MagNet consisting of an adversary detector 
and a data reformer is by far one of the strongest defenses in the \textit{black-box} oblivious attack setting, where the attacker aims to craft transferable adversarial examples from an undefended DNN model to bypass an unknown defense module deployed on the same DNN model. Under this setting, MagNet can successfully defend a variety of attacks in DNNs, including the high-confidence adversarial examples generated by the Carlini and Wagner's attack based on the $L_2$ distortion metric.
However, in this paper, under the same attack setting we show that adversarial examples crafted based on the $L_1$ distortion metric can easily bypass MagNet and mislead the target DNN image classifiers on MNIST and CIFAR-10. We also provide explanations on why the considered approach can yield adversarial examples with superior attack performance
and conduct extensive experiments on variants of MagNet to verify its lack of robustness to $L_1$ distortion based attacks.  
Notably, our results substantially weaken the assumption of effective threat models on MagNet that require knowing the deployed defense technique when attacking DNNs (i.e., the \textit{gray-box} attack setting).

%
%

\end{abstract}

\section{Introduction}\label{sec: Introduction}
DNNs are extensively used in many machine learning and artificial intelligence tasks. However, recent studies have highlighted that well-trained DNNs, albeit achieving superior prediction accuracy on natural examples, are in fact quite vulnerable to \textit{adversarial examples}.  For example, carefully designed adversarial perturbations to natural images can cause state-of-the-art image classifiers trained by DNNs to misclassify, while the adversarial perturbations can be made visually imperceptible \cite{szegedy2013intriguing, goodfellow2014explaining}. 
Even worse, in addition to digital spaces, adversarial examples can also be crafted in physical world by means of realizing adversarial perturbations. \cite{evtimov2017robust,athalye2017synthesizing,kurakin2016adversarial}. Due to the existence and the ease of generating adversarial examples from DNNs, 
the inconsistent decision making between DNN-based machine learning models and human perception as well as its robustness implications to safety-critical applications have given rise to the emerging research field intersecting deep learning and security.   

In the context of adversarial examples in DNNs, \textit{attacks} refer to means of crafting visually indistinguishable adversarial perturbations to natural examples, whereas \textit{defenses} refer to methods of mitigating adversarial perturbations towards building a robust DNN model. For the task of image classification, \textit{targeted attacks} aim to craft adversarial perturbations to render the prediction of the target DNN model towards a specific label, while \textit{untargeted attacks} aim to find adversarial perturbations that will lead the target DNN model to a different prediction.  
Perhaps surprisingly, the adversarial perturbations can be crafted even when the model details of the target DNN are totally unknown to an attacker, known as the (restricted) black-box attack setting \cite{papernot2017practical,CPY17_zoo_2}.
 
An important and perhaps surprising property of adversarial examples in DNNs is their attack transferability - adversarial examples generated from one DNN can also successfully fool another DNN, which we call transfer attacks \cite{szegedy2013intriguing,liu2016delving,papernot2016transferability}.  
Transfer attacks are widely used for evaluating the performance of attacks and defenses against adversarial examples in the black-box setting.
In the defender's perspective, the (oblivious) transfer attacks from an undefended DNN to a defended DNN of the same model serve as the baseline evaluation of the deployed defense techniques. In the attacker's foothold, executing a transfer attack is a preferable and practical option, as one can easily craft transferable adversarial examples from a DNN at hand to attack the target DNN without any prior knowledge of the target model.  Although various defense and adversarial subspace analysis methods have been proposed to defend transfer attacks, they have been continuously broken or bypassed by the subsequent attacks (but possibly with increased attack strengths) \cite{carlini2017towards,carlini2017adversarial,sharma2017breaking,lu2018limitation,athalye2018obfuscated,sharma2018bypassing}.

\begin{figure*}
	\centering
	\subfloat[MNIST]{
		\label{visual_MNIST}
		\includegraphics[scale=0.53]{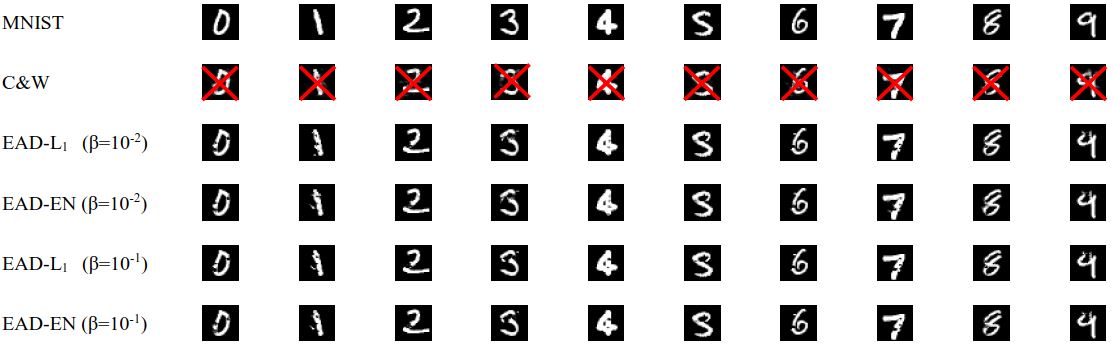}}
	\\
	\subfloat[CIFAR-10]{
		\label{visual_CIFAR}
		\includegraphics[scale=0.52]{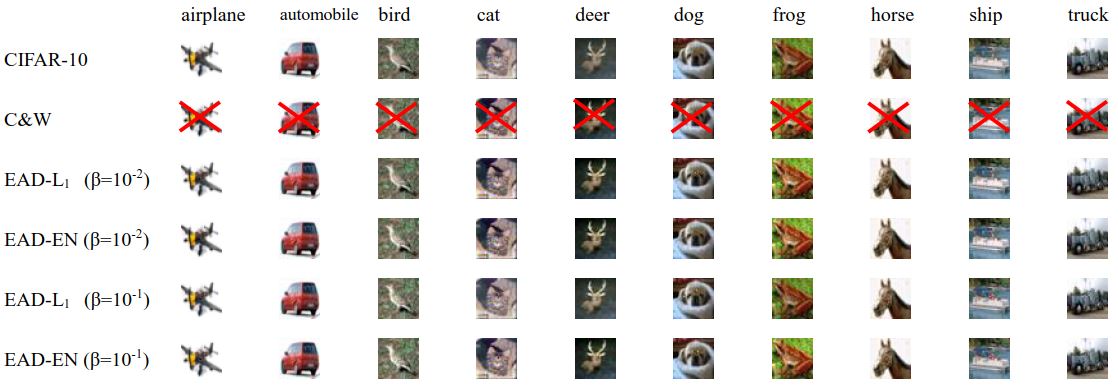}}
	\caption{Visual illustration of transferable adversarial examples crafted by different attack methods from undefended DNNs to MagNet in the oblivious (black-box) attack setting \cite{meng2017MagNet}. Unsuccessful attacks  are marked by red cross sign. EAD \cite{chen2017ead} can yield highly transferable and visually similar adversarial examples, whereas C\&W attack \cite{carlini2017towards} fails to bypass MagNet.}
	\label{fig:visual}
	\vspace*{-2mm}
\end{figure*}

Notably, the attack framework established by Carlini and Wagner in \cite{carlini2017towards}, which we call \textit{C\&W attack} for short, is a powerful attack that is capable of crafting highly transferable adversarial examples by tuning the \textit{confidence} parameter. However, in the oblivious attack setting a recent defense method called MagNet \cite{meng2017MagNet}, proposed by Meng and Chen, has demonstrated robust defense performance  against C\&W transfer attack under different confidence levels. 
In addition, MagNet can also defend other attacking methods including the fast gradient sign method (FGSM) \cite{goodfellow2014explaining}, iterative FGSM \cite{kurakin2016adversarial_ICLR}, and DeepFool \cite{moosavi2016deepfool}. The success of MagNet in defending adversarial examples roots in its two complementary defense modules: (i) a \textit{detector} that compares the statistical difference between an input image and the training data; and (ii) a \textit{reformer} trained by an auto-encoder that regulates an input image to the data manifold of training examples. Generally speaking, the detector module declares an input image as an adversarial example if its statistical distribution is significantly different from the training data. Otherwise, the input image further undergoes the reformer module, and the DNN will use the reformed example for label prediction. We defer the details of MagNet to Section \ref{subsec_magnet}.

In this paper, we demonstrate the limitation of MagNet on defending against the adversarial examples generated by the elastic-net attacks to DNNs (EAD) \cite{chen2017ead} in the oblivious attack setting. The major difference of C\&W attack and EAD is the distortion metric when crafting adversarial examples. C\&W attack is a pure $L_2$ distortion based method, whereas EAD is a hybrid attack using both $L_1$ and $L_2$ distortion metrics. As will be explained in Section \ref{subsec_EAD}, the use of $L_1$ distortion metric is able to filter out unnecessary perturbations to insignificant pixels and hence yielding adversarial examples 
with better attack performance. Specifically, our experimental results show that when using using EAD to attack MagNet with the default defense setting, about 90\% of adversarial examples on MNIST and 80\% of those on CIFAR-10 can successfully bypass MagNet, whereas only 10\% and 52\% of the adversarial examples from C\&W attack can bypass the same defense, respectively. For visual illustration, Figure \ref{fig:visual} shows some adversarial examples that successfully bypass MagNet using EAD. These adversarial examples are still visually similar to the natural examples but will cause the DNN to misclassify.   
Furthermore, to corroborate that MagNet is indeed not robust to $L_1$ distortion based adversarial examples, we also conduct extensive experiments  to evaluate the defense performance of MagNet under different settings, 
including tweaking the parameters of the detector module and changing the form of the reconstruction error when training the reformer. 

It is also worth mentioning that this paper is the first work to identify the lack of robustness of MagNet to adversarial examples in the oblivious (\textit{black-box}) attack setting \cite{meng2017MagNet}, where the attacker is completely unaware of the deployed defense mechanisms. In contrast, the recent work in \cite{carlini2017MagNet} also claims to break MagNet by using a much stronger threat model: the \textit{gray-box} setting where the attacker knows the deployed defense technique but not the exact parameters. Specifically, in order to craft transferable adversarial examples, Carlini and Wagner modified their attack by leveraging the knowledge that auto-encoder is the primary defense technique used in MagNet \cite{carlini2017MagNet}. Nonetheless, in the oblivious attack setting MagNet is still effective in defending the original C\&W attack in \cite{carlini2017towards}. Consequently, our results substantially weaken the assumption of the threat model required to bypass MagNet.

\section{Background and related work}\label{Background and related work}

\subsection{MagNet: Defending Adversarial Examples using Reformer and Detector \cite{meng2017MagNet}}
\label{subsec_magnet}
The essential component used in both the detector and the reformer of MagNet is the auto-encoder, denoted by $AE(\bx)$.  The auto-encoder $AE(\bx)$ takes an image $\bx \in \bbR^p$ as an input, compresses its information to a lower dimension, and then reconstructs the image $\bx$ back to the original dimension. The default MagNet setting learns a AE by minimizing the mean squared error $\|\bx - AE(\bx)\|_2^2$ averaged over all training examples. For an input image $\bx$, let $F(\bx) \in [0,1]^K$ denote a DNN image classifier of $K$ classes, which outputs a $K$-dimensional probability distribution of class predictions. The defense of MagNet is a serial two-stage process. First,
MagNet computes the Jensen-Shannon divergence (JSD) between $F(\bx)$ and $F(AE(\bx))$ with a temperature parameter  $T$, denoted by $JSD(F(\bx)/T),F(AE(\bx)/T)$.  The input $\bx$ is deemed adversarial if its JSD is greater than a certain threshold. Similarly, the reconstruction loss $\|\bx-AE(\bx)\|$ can also be used as a detector.
Otherwise, $\bx$ then undergoes the reformer $AE(\bx)$ before passing down to the DNN for classification. The reformer is responsible for projecting the input example to the data manifold learned by the auto-encoder such that the DNN is expected to yield correct label prediction after reforming the input example.
Overall, MagNet uses the detector to filter out adversarial examples with statistically significant perturbations and relies on the reformer to rectify the adversarial examples with small perturbations (those that are not rejected by the detector) towards correct class prediction. In Section \ref{sec: Experiments} we will evaluate the defense performance of the default MagNet setting and its robust variants.


\subsection{EAD: Elastic-Net Attacks to DNNs \cite{chen2017ead}}
\label{subsec_EAD}
Let $(\bx_0,t_0)$ denote a natural example with a associated class label $t_0$ and let $(\bx,t)$ denote its adversarial example with a target attack label $t \neq t_0$.   
The $L_q$ norm of the image difference  $\bdelta=\bx-\bx_0$, defined as $\| \bdelta \|_q=(\sum_{i=1}^p | \bdelta_i |^q)^{1/q}$  when $q \geq 1$, is a widely used distortion metric between natural and adversarial examples. For targeted attacks, EAD finds an effective adversarial example by solving the following optimization problem:
\begin{align}
\label{eqn_EAD_formulation}
&\textnormal{minimize}_{\bx}~~c \cdot f(\bx,t) +  \|\bx- \bx_0\|_2^2 + \beta \|\bx-\bx_0\|_1 \nonumber \\
&\textnormal{subject~to}~~\bx \in [0,1]^p,
\end{align}
where the box constraint $\bx \in [0,1]^p$ ensures every pixel value of $\bx$ lies within a valid normalized image space, 
$c,\beta \geq 0$ are regularization parameters for $f$ and $L_1$ distortion, respectively, and $f(\bx,t)$ is the attack loss function defined as 
\begin{align}
\label{eqn_loss_f}
f(\bx,t)=\max \{ \max_{j \neq t} [\textbf{Logit}(\bx)]_j - [\textbf{Logit}(\bx)]_t, - \kappa   \},
\end{align}
where $\textbf{Logit}(\bx)=[[\textbf{Logit}(\bx)]_1,\ldots,[\textbf{Logit}(\bx)]_K] \in \bbR^K$ is the logit of $\bx$ (the internal layer representation prior to the softmax layer)
in the considered DNN, also known as the \textit{unnormalized probabilities}. 
The parameter $\kappa \geq 0$ is called the \textit{confidence} that accounts for attack transferability. The hinge-like loss in (\ref{eqn_loss_f}) implies that the attack loss $f$ is minimized when its unnormalized probability of being the target class $t$ is $\kappa$ lager that that of being the next possible class prediction.  Similarly, for untargeted attacks, EAD uses the following attack loss function (dropping the notation $t$):
\begin{align}
\label{eqn_loss_f_untargeted}
f(\bx)=\max \{  [\textbf{Logit}(\bx)]_{t_0} - \max_{j \neq t} [\textbf{Logit}(\bx)]_j, - \kappa   \}.
\end{align}

Notable, C\&W attack \cite{carlini2017towards} is a special case of EAD when $\beta=0$, resulting in a pure $L_2$ distortion based attack. We argue that considering the $L_1$ distortion (i.e., set $\beta > 0$) is crucial in crafting transferable adversarial examples, which can be explained by the fact that $\beta$ plays the role of nulling unnecessary perturbations to insignificant pixels and shrinking the perturbation to important pixels, as indicated by C\&W attack. Specifically, let $g(\bx)=c \cdot f(\bx,t) +  \|\bx- \bx_0\|_2^2$ be the C\&W attack objective function from (\ref{eqn_EAD_formulation}) by setting $\beta=0$. 
When solving (\ref{eqn_EAD_formulation}) via projected gradient descent, EAD uses the iterative shrinkage-thresholding algorithm (ISTA) \cite{beck2009fast}:
\begin{align}
\label{eqn_ISTA}
\bx^{(k+1)}=S_{\beta}(\bx^{(k)}-\alpha_k \nabla g(\bx^{(k)}) ),
\end{align}
where  $\bx^{(k)}$ is the $k$-th iterate with $\bx^{(0)}=\bx_0$, $\nabla g(\bx^{(k)})$ denotes the gradient of $g$ at  $\bx^{(k)}$,
$\alpha_k$ denotes the step size, and $S_{\beta}: \bbR^{p} \mapsto \bbR^{p}$ is an pixel-wise projected shrinkage-thresholding function defined as
\begin{align}
\label{eqn_ISTA_S}
[S_{\beta}(\bz)]_i= \left\{
\begin{array}{ll}
\min \{\bz_i - \beta,1\}, & \text{~if~}\bz_i - {\bx_0}_i > \beta ; \\
{\bx_0}_i, & \text{~if~} |\bz_i - {\bx_0}_i| \leq \beta ; \\
\max \{\bz_i + \beta,0\}, & \text{~if~}\bz_i - {\bx_0}_i < -\beta,
\end{array}
\right.  
\end{align}
for any $i \in \{1,\ldots,p\}$. 
Therefore, with the use of ISTA, at each iteration EAD retains the original pixel value $[\bx_0]_i$ if the level of perturbation, indicated by 
$|[\bx^{(k)} - \bx_0 - \alpha_k \nabla g(\bx^{(k)})]_i |$, is no greater than $\beta$. Otherwise, it shrinks the level of perturbation by $\beta$ and projects the resulting pixel value to the box $[0,1]$
if $|[\bx^{(k)} - \bx_0 - \alpha_k \nabla g(\bx^{(k)})]_i |>\beta$. Furthermore, since $g$ is the attack objective function of C\&W attack, EAD can be interpreted as a sparsity-induced C\&W attack, where the ISTA step at each iteration adds zero perturbation to the $i$-th pixel if its C\&W attack gradient $[\nabla g]_i$ is small (i.e., the pixel is deemed insignificant for attack), or reduces the perturbation by $\beta$ if $[\nabla g]_i$ is large, leading to sharp adversarial examples with better attack performance.


\section{Experiments}\label{sec: Experiments}
In this section, we demonstrate EAD can bypass MagNet on two popular image classification datasets - MNIST and CIFAR-10. MNIST is a popular handwritten digit dataset. Each image in the dataset represents a number from 0 to 9. For CIFAR-10, there are 10 image categories: airplanes, cars, birds, cats, deer, dogs, frogs, boats, trucks. 
 The visual illustrations of adversarial examples crafted by different attack methods on the default MagNet are displayed in Figure \ref{fig:visual}.


\begin{table}
	\caption{Comparison of different attacks on MagNet (default setting) with different confidence $\kappa$ on MNIST and CIFAR-10. ASR means attack success rate (\%). The distortion metrics are averaged over successful examples. EAD yields high ASR. }
	\label{tab:ASR}	
	\centering
\begin{adjustbox}{max width=\columnwidth}	
	\begin{tabular}{ll|llll|llll}
		\hline
		& & \multicolumn{4}{c}{MNIST}& \multicolumn{4}{c}{CIFAR-10} \\
		\hline
		Attack method &$\beta$& $\kappa$& ASR&$L_1$&$L_2$& $\kappa$& ASR&$L_1$&$L_2$\\
		\hline
	    C\&W ($L_2$)  & NA&15&10 (best) &3.553&1.477&20&52&3.675&0.126\\
		\hline
		
		\multirow{4}{33pt}{EAD\quad(EN~rule)}&$10^{-3}$ & 20&46.2&3.116&2.165&15&69.2&3.024&0.242\\
		&$10^{-2}$&15& 87.8 & 0.531 & 2.509 &15&74.5&2.73&0.380\\
		&$5\cdot10^{-2}$&15&90.1&0.266&2.730&15&77&2.810&0.544 \\
		&$10^{-1}$&15& 90.2&0.433&2.803&15&78.6&3.234&0.681 \\
		\hline

		\multirow{4}{33pt}{EAD\quad($L_1$~rule)}&$10^{-3}$ & 20&70.2&1.89&2.507&15&60.5&1.1718&0.327\\
		&$10^{-2}$&15& 84.5 & 0.449 & 2.701 &15&66.7&1.646&0.495\\
		&$5\cdot10^{-2}$&15&80.5&0.351&2.876&15&75.9&2.258&0.678 \\
		&$10^{-1}$&15& 83.8&0.381&2.922&15&79.8&2.883&0.805 \\
		\hline
	\end{tabular}
	\end{adjustbox}
	\vspace{-2mm}
\end{table}

	\subsection{Experiment Setup, Parameter Setting and Threat Model}\label{Experiment Setup and Parameter Setting}
We follow the oblivious attack setting used in MagNet \cite{meng2017MagNet} to implement untargeted attacks from an undefended DNN to the same DNN protected by MagNet, where the attacker is unaware of MagNet's existence.
 The same DNN architecture and training parameters in \cite{meng2017MagNet} are used to train the image classifiers on MNIST and CIFAR-10.
The defense performance against adversarial examples (which we call \textit{classification accuracy}) of MagNet is measured by the percentage of adversarial examples that are either detected by the MagNet's detector, or correctly classified by the DNN after reforming, which complements the attack success rate (ASR). That is, higher ASR implies weaker defense.
 We focus on the comparison between C\&W attack\footnote[1]{https://github.com/carlini/nn robust attacks.} ($L_2$ based attack) and EAD\footnote[2]{https://github.com/ysharma1126/EAD-Attack} ($L_1$ based attack) when evaluating the defense capability of MagNet.
The best regularization parameter $c$ is obtained  via 9 binary search steps (starting from 0.001) and 1000 iterations are used for each attack with the same initial learning rate 0.01. For EAD, we report the attack results using different $L_1$ regularization parameter $\beta$ and decision rules (elastic-net  (EN) or $L_1$ distortion) for selecting the final adversarial example.
On MNIST and CIFAR-10, we craft adversarial examples with different confidence level $\kappa$ picked in the range of [0, 40] and [0,100]
with an increment of 5, respectively.  The default MagNet setting\footnote[3]{https://github.com/Trevillie/MagNet} and our implemented robust variants are used for defense evaluation.  On both MNIST and CIFAR-10, we randomly selected 1000 correctly classified images from the test sets to attack MagNet. All experiments are conducted using an Intel Xeon E5-2620v4 CPU, 125 GB RAM and a NVIDIA TITAN Xp GPU with 12 GB RAM.

	Note that although Meng et al. provide their training parameters and DNN model used in their best experimental results, we cannot reproduce such effectively defensive results displayed in the MagNet paper \cite{meng2017MagNet}. All the reported results in this paper correspond to our self-trained MagNet.


	\subsection{Performance Evaluation of Oblivious Attacks on MagNet}\label{Overall performance breaking MagNet}
	
	 When crafting adversarial examples, the attack strength can be adjusted by changing the confidence level $\kappa$. The higher the confidence, the stronger the attack strength, but also the greater the distortion.
     Table \ref{tab:ASR} summarizes the statistics of adversarial examples crafted by different attacks under different confidence $\kappa$ on MNIST and CIFAR-10 against the default MagNet.
	 It is apparent that considering $L_1$ distortion when crafting adversarial examples indeed greatly improves attack performance when compared to merely using $L_2$ distortion, as discussed in Section \ref{subsec_EAD}. 
	
	In addition to showing the attack results under the default MagNet setting, we also adjusted the defense parameters used in MagNet to make it more robust, which we call robust MagNet. However, we find that even MagNet's defensive capability can be improved, EAD can still effectively attack robust MagNet. In what follows,
	Section \ref{MNIST} and Section \ref{CIFAR-10} discuss the experimental results of EAD attacking MagNet, and Section \ref{Adujust parameters MNIST} and Section \ref{Adujust parameters CIFAR} discuss the experimental results of EAD attacking robust MagNet.

		\subsubsection{MagNet with default setting on MNIST}\label{MNIST}
\begin{figure}[t]
	\vspace{-7mm}
	\centering
	\subfloat[Default (D)]{
		\includegraphics[scale=0.29]{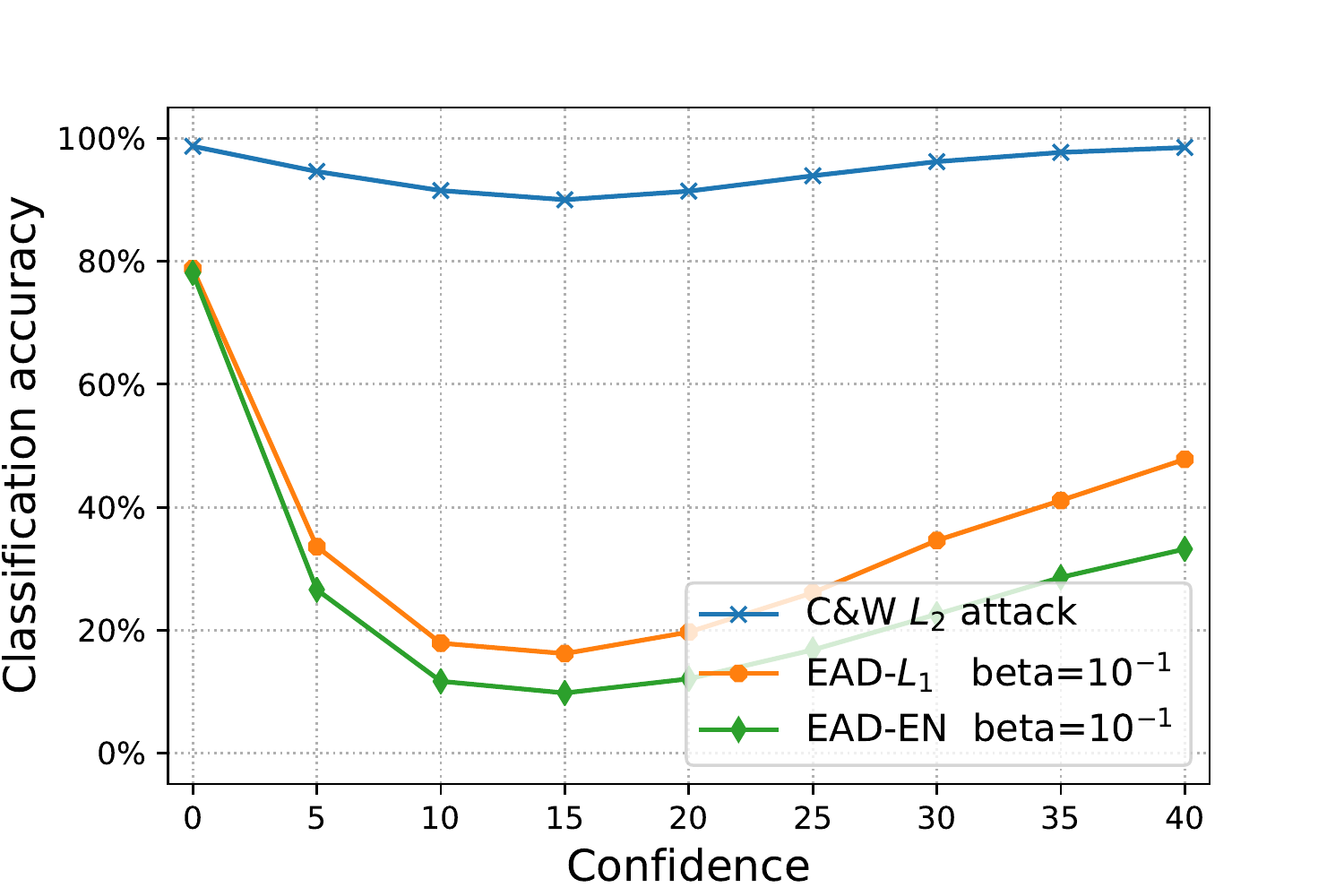}}
	\subfloat[D+JSD]{
		\includegraphics[scale=0.29]{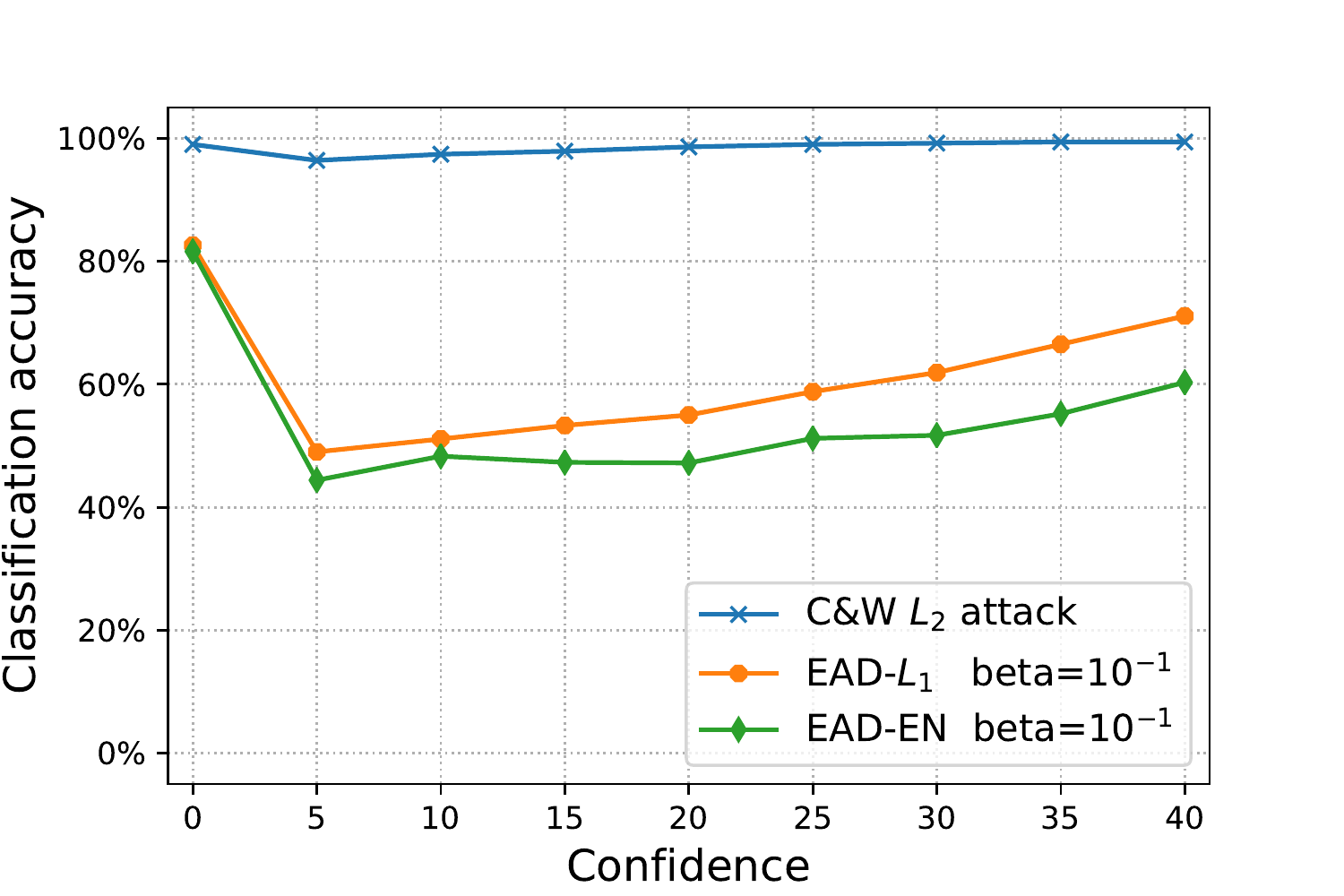}}	
	\\
	\vspace{-4mm}
	\subfloat[D+256]{
		\includegraphics[scale=0.29]{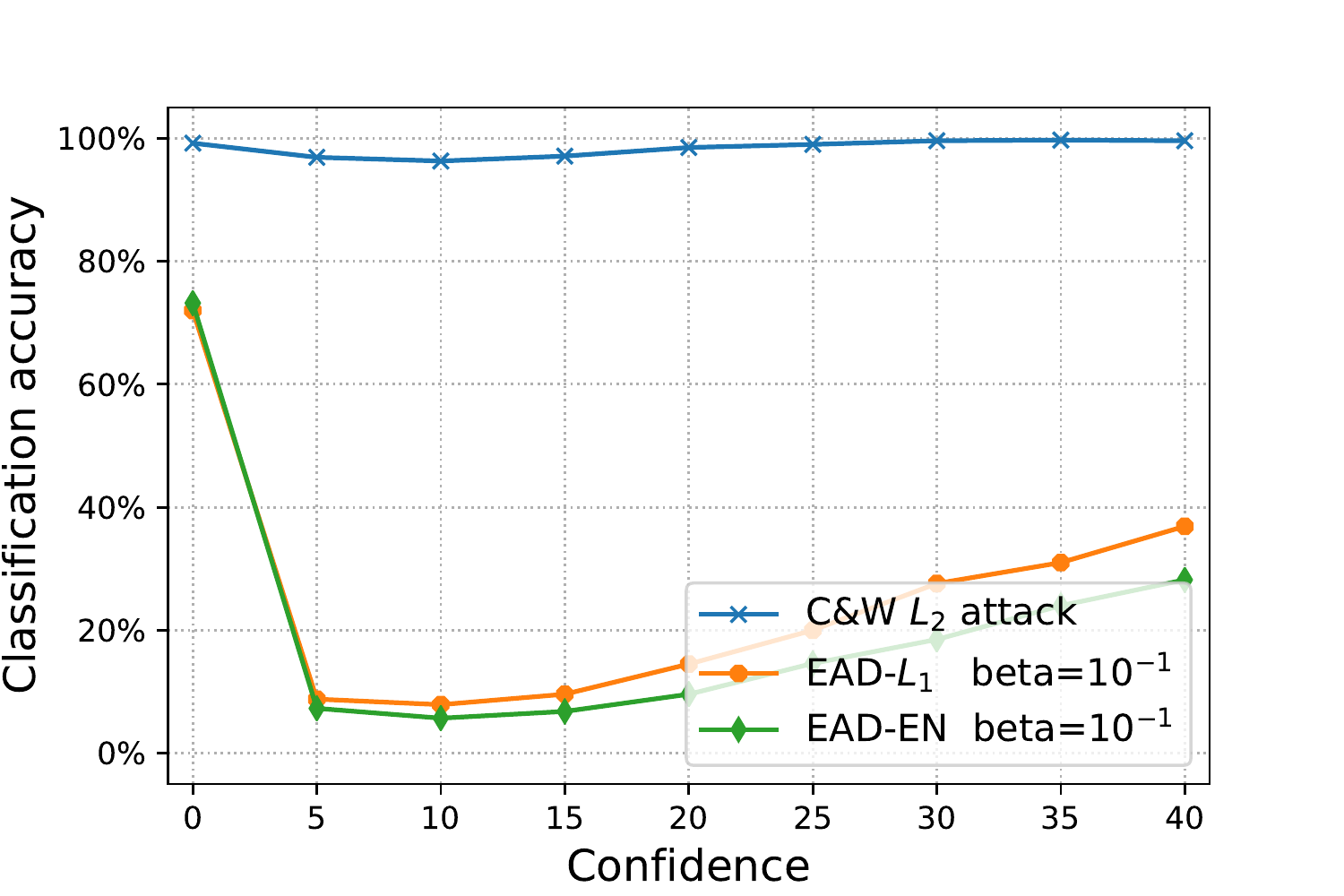}}
	\subfloat[D+256+JSD]{
		\includegraphics[scale=0.29]{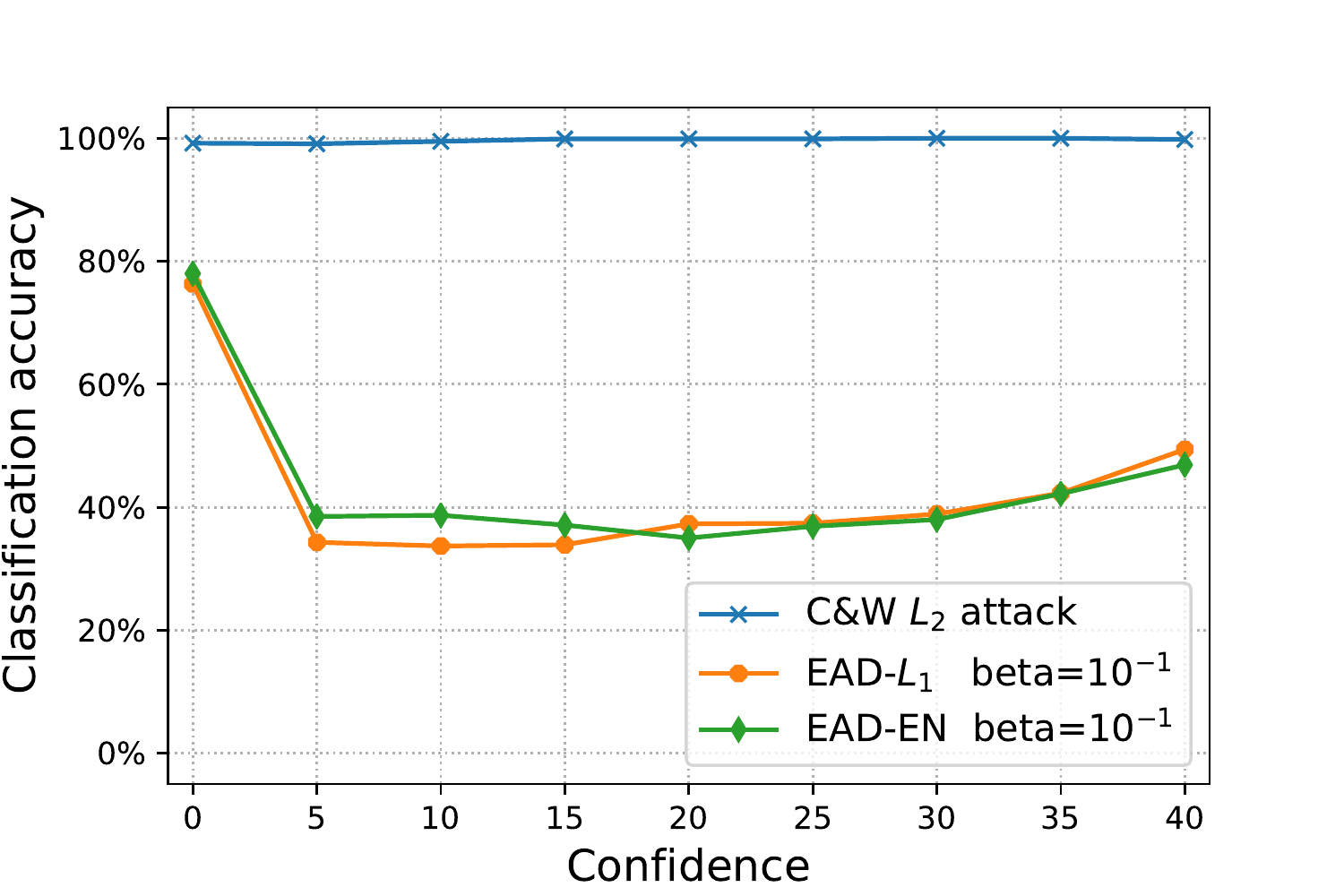}}
	\caption{Defense performance of default and robust MagNets against C\&W and EAD attacks on MNIST. Higher accuracy means better defense. Different from C\&W attack, EAD significantly degrades MagNet's defense performance. }
	\label{fig:MNIST_main}
	\vspace{-2mm}
\end{figure}

		Because MNIST is a simple image classification task, Meng et al. \cite{meng2017MagNet} use only two detectors based on  $L_1$ and $L_2$ reconstruction errors. Figure \ref{fig:MNIST_main} (a) shows the defense performance of the default MagNet. It is observed that  the detector rejects more adversarial examples and hence becomes more effective as the confidence increases, which can be explained by the fact that the detector is designed to filter out the input example when it is too far away from the data manifold of natural training examples.
		On the other hand, when the input example is close to the data manifold, the reformer is in charge of rectifying the input via the trained auto-encoder. The detector and reformer hence compliment each other and constitute MagNet.	   Interestingly,			
		there is a dip when the confidence levels are between 10 and 20 because in this range, the effectiveness of the reformer is diminishing and the detectors are yet  ineffective. 
		
		Remarkably, the classification accuracy of C\&W attack is above $90\%$ at all confidence levels, meaning that C\&W attack to MagNet is not effective in the oblivious attack setting. Indeed, in \cite{carlini2017MagNet} Carlini and Wagner impose a much stronger threat model assumption than the oblivious attack setting 
		in order to bypass the default MagNet, which requires knowing the existence of auto-encoder as a defense module.	On the other hand, we find that in the oblivious attack setting, the default MagNet fails to defend a majority of adversarial examples crafted by EAD. For instance,
		comparing C\&W attack with EAD when setting $\beta=10^{-1}$ at the confidence 15, MagNet's classification accuracy reduces significantly from 90\% to 9.7\%, suggesting that approximately 90$\%$ of adversarial examples crafted by EAD with confidence 15 can bypass MagNet. Comparing to \cite{carlini2017MagNet},
		this result suggests that a weaker threat model that is completely unaware of MagNet defense is sufficient for successful adversarial attacks on MNIST.		
		It it also worth noting that since C\&W attack only uses $L_2$ distortion while EAD uses both $L_1$ and $L_2$ distortion,
		the significant difference in defense performance degradation corroborates the effect of involving the $L_1$ distortion when crafting effective adversarial examples.

		We also investigate the effects of the $L_1$ regularization parameter $\beta$ and decision rules in EAD on attacking the default MagNet. Due to space limitations, we summarize our findings below and display the visual diagrams in the supplementary material.
		 When $\beta$ is small (e.g., $\beta=10^{-3}$), we obtain a better attack performance under the $L_1$ decision rule than that under the EN rule because in this case $L_2$ distortion dominates the EN distortion.
	    As $\beta$ becomes larger, the attack performance of EAD under the EN rule is better than that under the $L_1$ rule, which can be explained by the potential over-contraction and aggressive thresholding for large $\beta$ in the ISTA step of EAD.

		\subsubsection{EAD attack on robust MagNet on MNIST}\label{Adujust parameters MNIST}
		The default setting of MagNet on MNIST contains only two detectors based on $L_1$ and $L_2$ reconstruction errors, respectively. We added two JSD detectors with temperature $T$ of 10 and 40 to MagNet. In Figure \ref{fig:MNIST_main} (b), this MagNet can achieve above 96$\%$ classification accuracy on adversarial examples generated by C\&W $L_2$ attack at all tested confidence levels. Comparing to  Figure \ref{fig:MNIST_main} (a), the defense performance is
		indeed improved by including JSD detectors. However, it is still not robust to EAD, since 
	 approximately 40 $\%$ of adversarial examples crafted by EAD  can still bypass this MagNet.

\begin{table}[t]
	\caption{Defensive architecture of robust MagNet on MNIST, including both encoders and decoders.}	
	\centering
	\begin{tabular}{llll}	\hline
		\multicolumn{2}{c}{Detector I \& Reformer} & \multicolumn{2}{c}{Detector II}\\
		\hline
		Conv.Sigmoid   &$3 \times 3 \times 256 $& Conv.Sigmoid& $3 \times 3 \times 256 $\\
		AveragePooling &$ 2 \times 2$ & Conv.Sigmoid& $3 \times 3 \times 256 $\\
		Conv.Sigmoid   &$3 \times 3 \times 256 $ & Conv.Sigmoid& $3 \times 3 \times 1 $ \\
		Conv.Sigmoid   &$3 \times 3 \times 256 $ &  &\\
		Upsampling     &$ 2 \times 2$ & &\\
		Conv.Sigmoid   &$3 \times 3 \times 256$ &&\\
		Conv.Sigmoid   &$3 \times 3 \times 1$ &&\\
		\hline
	\end{tabular}
	\label{tab:robust MagNet on MNIST}
			\vspace{-2mm}
\end{table}

\begin{table}[t]
	\caption{ MNIST test accuracy (\%).}	
	\centering
	\begin{tabular}{|p{30pt}|c|c|c|c|}
		\hline
		& Default (D)  & D+JSD & D+256 & D+256+JSD \\
		\hline
		Without MagNet & 99.42 & 99.42 & 99.42 & 99.42 \\
		\hline
		With MagNet & 99.13 & 97.75 & 99.24 & 97.55 \\
		\hline
	\end{tabular}
	\label{tab:MNIST}
			\vspace{-4mm}
\end{table}
		
		We also changed the number of filters used in a auto-encoder's convolution layer from 3 to 256 (see Table \ref{tab:robust MagNet on MNIST}). We find that auto-encoders can be more stable and achieve better performance on encoding and decoding by increasing the number of filters within a reasonable range, and this factor actually improves the robustness of MagNet. As shown in Figure \ref{fig:MNIST_main} (c), this change indeed leads to better defense against C\&W attack, particularly in the confidence level ranging from 5 to 25. However, approximately 70 $\%$ of adversarial examples crafted by EAD can still bypass this robust MagNet.

	Figure \ref{fig:MNIST_main} (d) shows that MagNet can further improves its defense performance by jointly changing the number of filter to 256 and adding the two JSD detectors. Nonetheless, it still fails to defense against approximately 50 $\%$ of adversarial examples crafted by EAD, suggesting limited robustness in MagNet against $L_1$ distortion based adversarial examples. 

		To justify the vulnerability of MagNet to $L_1$ distortion based adversarial examples is not caused by the use of $L_2$ reconstruction error when training the auto-encoders in MagNet, we also trained auto-encoders using  the mean absolute error ($L_1$ loss) on MNIST.
	    We find that these auto-encoders used in MagNet can defend C\&W $L_2$ attacks but are still susceptible to EAD. The visual diagrams are given in the supplementary material.    
		
		Given different MagNet models on MNIST, Table \ref{tab:robust MagNet on MNIST} and Table \ref{tab:MNIST} summarize the prediction accuracy on the test dataset and the best ASR of adversarial examples among the tested confidence levels, respectively.
		We conclude that the default and robust MagNet are able to defeat $L_2$ attacks, while they are still vulnerable to $L_1$ attacks using EAD.

\begin{table}[t]
	\caption{Best attack success rate (ASR)  ($\%$) of  EAD on MNIST. Higher ASR means weaker defense. }	
	\centering
\begin{adjustbox}{max width=\columnwidth}	
	\begin{tabular}{p{43pt}l|l|l|l|l}
		\hline
		 Decision rule&$\beta$&Default (D)  & D+JSD & D+256 & D+256+JSD \\
		\hline
		\multirow{4}{33pt}{EAD\quad(EN~rule)}&$10^{-3}$ & 46.2 & 7.5 & 31.2 & 1.9 \\
		  &$10^{-2}$& 87.8 & 34 & 90.1 & 39.5 \\
		  &$5\cdot10^{-2}$&90.1&51.6&93.6&60 \\
		  &$10^{-1}$& 90.2&55.6&94.3&65.1 \\
		 \hline
		\multirow{4}{33pt}{EAD\quad($L_1$~rule)}&$10^{-3}$ &70.2&18.9&72.9&14.1\\
		  &$10^{-2}$ & 84.5&38.8&92.6&49.5 \\
		  &$5\cdot10^{-2}$ & 80.5&48.8&90.3&62.6\\
		  &$10^{-1}$ & 83.8&51&92.1&66.3\\
		\hline
	\end{tabular}
\end{adjustbox}
	\label{tab:MNIST_ASR}
		\vspace{-8mm}	
\end{table}

\begin{figure}[t]
	\centering
	\subfloat[Default (D)]{
		\includegraphics[scale=0.29]{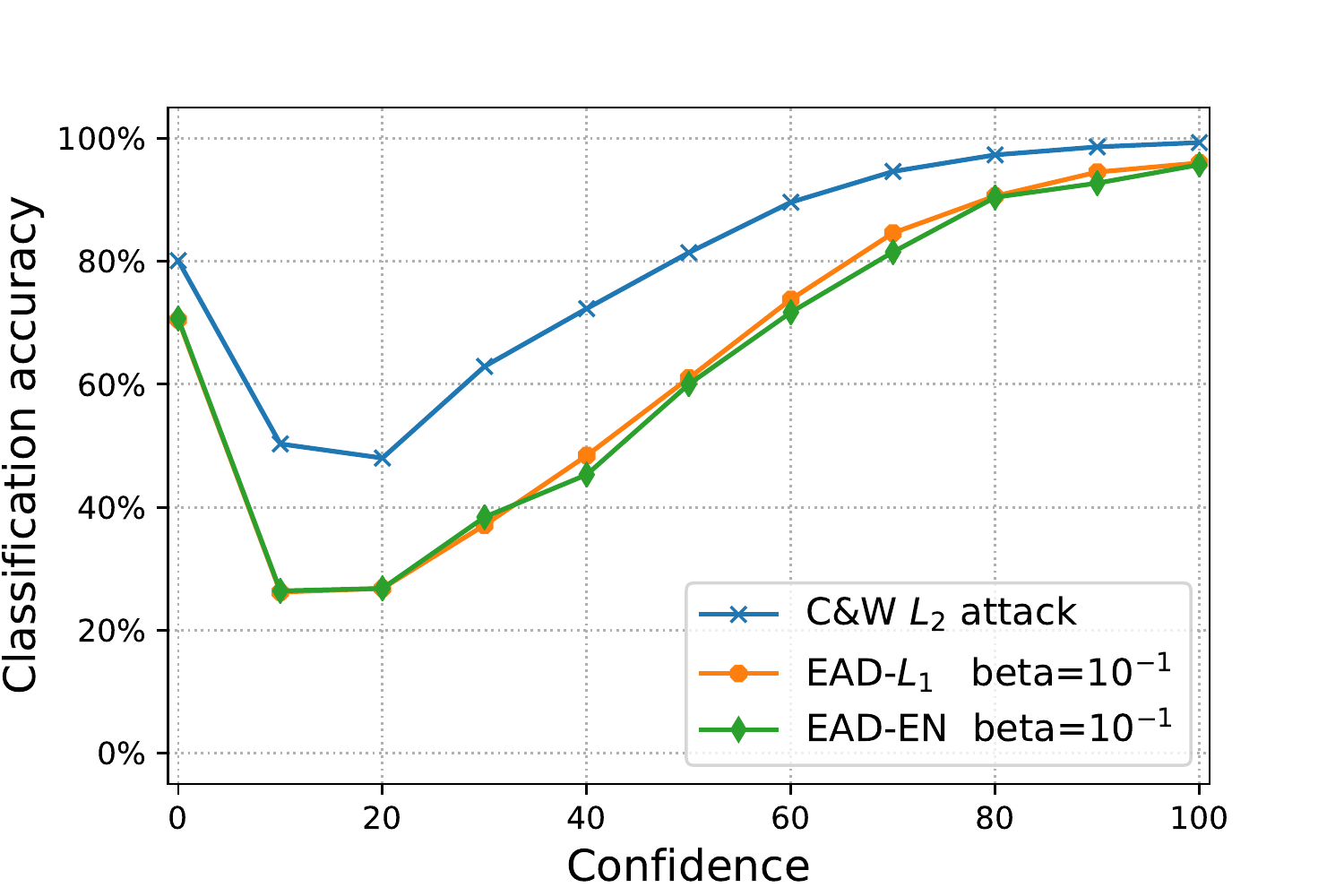}}
	\subfloat[D+256]{
		\includegraphics[scale=0.29]{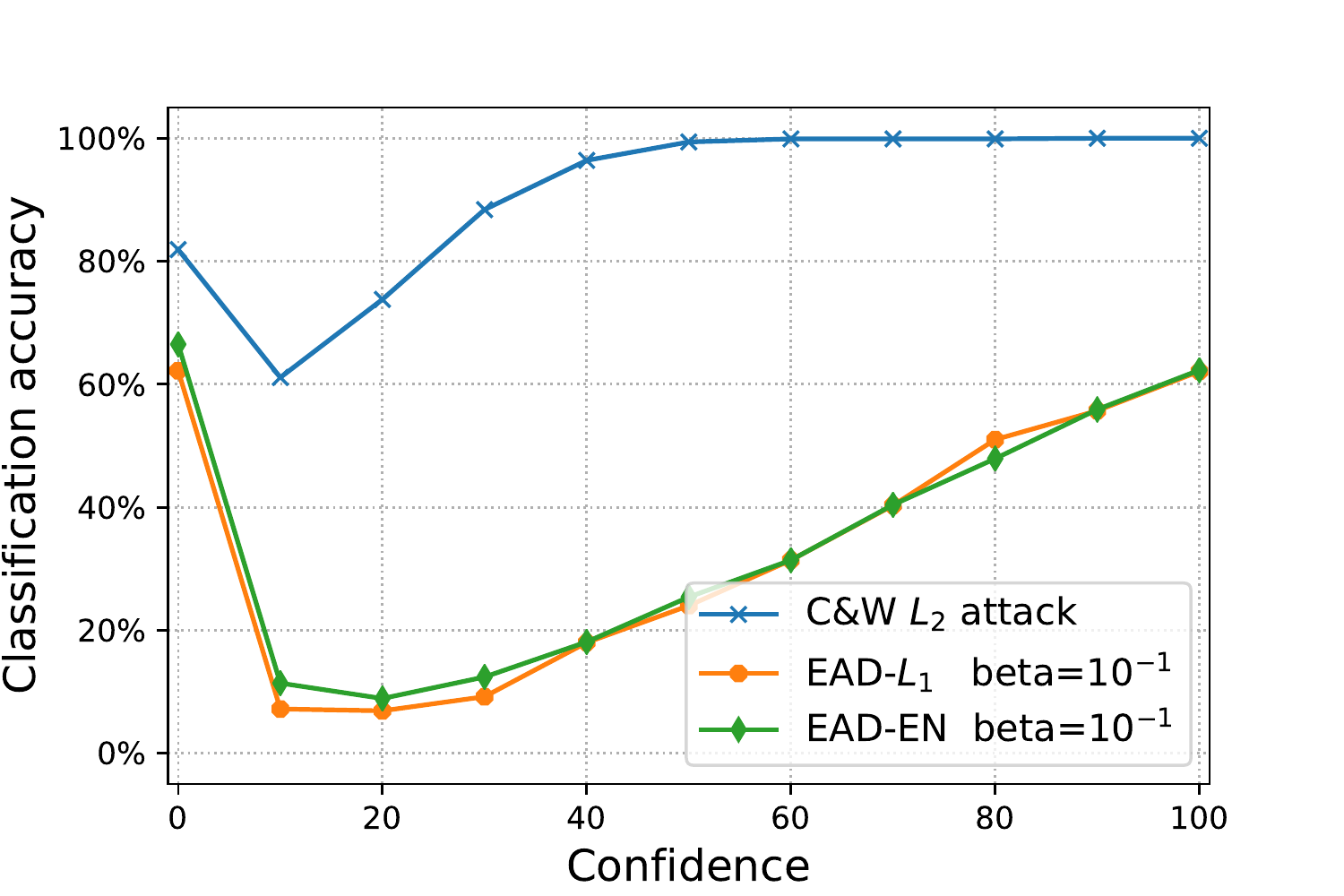}}
	\caption{Defense performance of default and robust MagNets against C\&W and EAD attacks on CIFAR-10. Higher accuracy means better defense. Different from C\&W attack, EAD significantly degrades MagNet's defense performance. }
	\label{fig:CIFAR_main}	
	\vspace{-4mm}	
\end{figure}

	\subsubsection{Default MagNet on CIFAR-10}\label{CIFAR-10}
	Training and securing a classifier on CIFAR-10 is more challenging than that on MNIST. In Magnet, Meng et al. use two types of detectors  based on $L_1$ and $L_2$ reconstruction errors as well as two JSD detectors with the temperature $T$ of 10 and 40. Specifically, the JSD detectors  are shown to be effective in detecting adversarial examples with large reconstruction errors.

    We adjusted the strength of attack by changing the confidence in the range from 0 to 100 and show the defense performance in Figure \ref{fig:CIFAR_main} (a). Using the default MagNet,
    approximately 70$\%$ of the adversarial examples crafted by EAD will not be detected or corrected by MagNet at confidence 10. Moreover, despite using effective detectors on CIFAR-10,  MagNet's classification accuracy is particularly vulnerable to EAD at the confidence levels ranging from 10 to 20.

	\subsubsection{EAD attack on robust MagNet on CIFAR-10}\label{Adujust parameters CIFAR}
	We changed the number of filters used in a auto-encoder's convolution layer from 3 to 256 (see Table \ref{tab:robust MagNet on CIFAR}) and summarize the resulting CIFAR-10 test accuracy in Table \ref{tab:CIFAR}. Comparing  Figure \ref{fig:CIFAR_main} (a) and  Figure \ref{fig:CIFAR_main} (b), this robust MagNet aids in more effective defense against C\&W $L_2$ attack at all confidence levels than the default MagNet on CIFAR-10.
	 However, we find that EAD can still attain high attack success rate as $\beta$ increases in both defense settings, as summarized in Table \ref{tab:CIFAR_ASR}, which  suggests limited defense capability of MagNet against $L_1$ distortion based adversarial examples on CIFAR-10.

\begin{table}[t]
	\caption{Defensive architecture of robust MagNet on CIFAR-10, including both encoders and decoders.}	
	\centering	
	\begin{tabular}{ll}	\hline
		\multicolumn{2}{c}{Detectors \& Reformer}\\
		\hline
		Conv.Sigmoid& $3 \times 3 \times 256 $\\
		Conv.Sigmoid& $3 \times 3 \times 256 $\\
		Conv.Sigmoid& $3 \times 3 \times 3 $ \\
		\hline
	\end{tabular}
	\label{tab:robust MagNet on CIFAR}
	\vspace{-2mm}
\end{table}

\begin{table}[t]
	\caption{CIFAR-10 test accuracy (\%).}	
	\centering
	\begin{tabular}{|p{30pt}|l|l|}
		\hline
		& Default (D) & D + 256\\
		\hline
		Without MagNet & 86.91 & 86.91 \\
		\hline
		With MagNet & 83.33 & 83.4 \\
		\hline
	\end{tabular}
	\label{tab:CIFAR}
		\vspace{-4mm}
\end{table}	

\begin{table}[t]
	\caption{Best attack success rate (ASR) ($\%$) of  EAD on CIFAR-10. Higher ASR means weaker defense. }	
	\centering
	\begin{tabular}{p{43pt}l|l|l}
		\hline
		Decision rule&$\beta$& Default (D) & D+256 \\
		\hline
		\multirow{4}{33pt}{EAD (EN~rule)}&$10^{-3}$ &69.2&55.6\\
		&$10^{-2}$&74.5&72\\
		&$5\cdot10^{-2}$&77&86.3\\
		&$10^{-1}$& 78.6&91.5\\
		\hline
		\multirow{4}{33pt}{EAD ($L_1$~rule)}&$10^{-3}$ &60.5&49.2\\
		&$10^{-2}$ &66.7&71.8\\
		&$5\cdot10^{-2}$ &75.9&90.9\\
		&$10^{-1}$ &79.8&93.7\\
		\hline
	\end{tabular}
	\label{tab:CIFAR_ASR}
	\vspace{-4mm}
\end{table}

	We also investigated the defense performance of MagNet with different reconstruction errors for training the auto-encoders on the CIFAR-10 training set. Similar to MNIST, we find that  on CIFAR-10, replacing the mean squared error with the mean absolute error  can defend C\&W $L_2$ attacks but not the $L_1$ based attacks using EAD. The visual diagrams are given in the supplementary material.

\section{Conclusion and Discussion}\label{sec: Conclusion}
We summarize the main results of this paper as follows:

\begin{itemize}
	\item On MNIST, the default MagNet using the detectors merely based on reconstruction errors is not robust.
	\item The setting of auto-encoders has a great influence on MagNet's defensive ability.
	\item Despite its success in defending $L_2$ distortion based adversarial examples on MNIST and CIFAR-10, MagNet and its robust variants are ineffective against $L_1$ distortion based adversarial examples crafted by EAD. 
	Furthermore, even though we implemented improved  detectors in MagNet, EAD can still easily craft adversarial examples that bypass MagNet in the oblivious attack setting, which substantially weakens the existing attack assumption of knowing the deployed defense technique when attacking MagNet in the existing literature.
\end{itemize}

Based on our analysis, we have the following suggestions for evaluating and designing future defense models:
\begin{itemize}
	\item $L_1$ and $L_2$ based adversarial examples have distinct characteristics. Future defense models should test their robustness against both cases.
	\item In the oblivious attack setting, neither the reformer nor the detector in MagNet can effectively defend against adversarial examples at the medium confidence levels, which calls for additional defense mechanisms. 	
\end{itemize}

\bibliographystyle{IEEEtran}
\bibliography{IEEEabrv,bibfile,adversarial_learning}

\begin{thebibliography}{10}
\providecommand{\url}[1]{#1}
\csname url@samestyle\endcsname
\providecommand{\newblock}{\relax}
\providecommand{\bibinfo}[2]{#2}
\providecommand{\BIBentrySTDinterwordspacing}{\spaceskip=0pt\relax}
\providecommand{\BIBentryALTinterwordstretchfactor}{4}
\providecommand{\BIBentryALTinterwordspacing}{\spaceskip=\fontdimen2\font plus
\BIBentryALTinterwordstretchfactor\fontdimen3\font minus
  \fontdimen4\font\relax}
\providecommand{\BIBforeignlanguage}[2]{{%
\expandafter\ifx\csname l@#1\endcsname\relax
\typeout{** WARNING: IEEEtran.bst: No hyphenation pattern has been}%
\typeout{** loaded for the language `#1'. Using the pattern for}%
\typeout{** the default language instead.}%
\else
\language=\csname l@#1\endcsname
\fi
#2}}
\providecommand{\BIBdecl}{\relax}
\BIBdecl

\bibitem{szegedy2013intriguing}
C.~Szegedy, W.~Zaremba, I.~Sutskever, J.~Bruna, D.~Erhan, I.~Goodfellow, and
  R.~Fergus, ``Intriguing properties of neural networks,'' \emph{arXiv preprint
  arXiv:1312.6199}, 2013.

\bibitem{goodfellow2014explaining}
I.~J. Goodfellow, J.~Shlens, and C.~Szegedy, ``Explaining and harnessing
  adversarial examples,'' \emph{ICLR'15; arXiv preprint arXiv:1412.6572}, 2015.

\bibitem{evtimov2017robust}
I.~Evtimov, K.~Eykholt, E.~Fernandes, T.~Kohno, B.~Li, A.~Prakash, A.~Rahmati,
  and D.~Song, ``Robust physical-world attacks on machine learning models,''
  \emph{arXiv preprint arXiv:1707.08945}, 2017.

\bibitem{athalye2017synthesizing}
A.~Athalye and I.~Sutskever, ``Synthesizing robust adversarial examples,''
  \emph{arXiv preprint arXiv:1707.07397}, 2017.

\bibitem{kurakin2016adversarial}
A.~Kurakin, I.~Goodfellow, and S.~Bengio, ``Adversarial examples in the
  physical world,'' \emph{arXiv preprint arXiv:1607.02533}, 2016.

\bibitem{papernot2017practical}
N.~Papernot, P.~McDaniel, I.~Goodfellow, S.~Jha, Z.~B. Celik, and A.~Swami,
  ``Practical black-box attacks against machine learning,'' in \emph{ACM Asia
  Conference on Computer and Communications Security}, 2017, pp. 506--519.

\bibitem{CPY17_zoo_2}
P.-Y. Chen, H.~Zhang, Y.~Sharma, J.~Yi, and C.-J. Hsieh, ``Zoo: Zeroth order
  optimization based black-box attacks to deep neural networks without training
  substitute models,'' in \emph{ACM Workshop on Artificial Intelligence and
  Security}, 2017, pp. 15--26.

\bibitem{liu2016delving}
Y.~Liu, X.~Chen, C.~Liu, and D.~Song, ``Delving into transferable adversarial
  examples and black-box attacks,'' \emph{arXiv preprint arXiv:1611.02770},
  2016.

\bibitem{papernot2016transferability}
N.~Papernot, P.~McDaniel, and I.~Goodfellow, ``Transferability in machine
  learning: from phenomena to black-box attacks using adversarial samples,''
  \emph{arXiv preprint arXiv:1605.07277}, 2016.

\bibitem{carlini2017towards}
N.~Carlini and D.~Wagner, ``Towards evaluating the robustness of neural
  networks,'' in \emph{IEEE Symposium on Security and Privacy (SP)}, 2017, pp.
  39--57.

\bibitem{carlini2017adversarial}
------, ``Adversarial examples are not easily detected: Bypassing ten detection
  methods,'' \emph{arXiv preprint arXiv:1705.07263}, 2017.

\bibitem{sharma2017breaking}
Y.~Sharma and P.-Y. Chen, ``Attacking the {Madry} defense model with
  {$L_1$}-based adversarial examples,'' \emph{ICLR Workshop; arXiv:1710.10733},
  2018.

\bibitem{lu2018limitation}
P.-H. Lu, P.-Y. Chen, and C.-M. Yu, ``On the limitation of local intrinsic
  dimensionality for characterizing the subspaces of adversarial examples,''
  \emph{ICLR Workshop; arXiv:1803.09638}, 2018.

\bibitem{athalye2018obfuscated}
A.~Athalye, N.~Carlini, and D.~Wagner, ``Obfuscated gradients give a false
  sense of security: Circumventing defenses to adversarial examples,''
  \emph{arXiv preprint arXiv:1802.00420}, 2018.

\bibitem{sharma2018bypassing}
Y.~Sharma and P.-Y. Chen, ``Bypassing feature squeezing by increasing adversary
  strength,'' \emph{arXiv preprint arXiv:1803.09868}, 2018.

\bibitem{meng2017MagNet}
D.~Meng and H.~Chen, ``Magnet: a two-pronged defense against adversarial
  examples,'' \emph{ACM CCS}, 2017.

\bibitem{chen2017ead}
P.-Y. Chen, Y.~Sharma, H.~Zhang, J.~Yi, and C.-J. Hsieh, ``Ead: Elastic-net
  attacks to deep neural networks via adversarial examples,'' \emph{AAAI;
  arXiv:1709.04114}, 2018.

\bibitem{kurakin2016adversarial_ICLR}
A.~Kurakin, I.~Goodfellow, and S.~Bengio, ``Adversarial machine learning at
  scale,'' \emph{ICLR'17; arXiv preprint arXiv:1611.01236}, 2016.

\bibitem{moosavi2016deepfool}
S.-M. Moosavi-Dezfooli, A.~Fawzi, and P.~Frossard, ``Deepfool: a simple and
  accurate method to fool deep neural networks,'' in \emph{Proceedings of the
  IEEE Conference on Computer Vision and Pattern Recognition}, 2016, pp.
  2574--2582.

\bibitem{carlini2017MagNet}
N.~Carlini and D.~Wagner, ``Magnet and" efficient defenses against adversarial
  attacks" are not robust to adversarial examples,'' \emph{arXiv preprint
  arXiv:1711.08478}, 2017.

\bibitem{beck2009fast}
A.~Beck and M.~Teboulle, ``A fast iterative shrinkage-thresholding algorithm
  for linear inverse problems,'' \emph{SIAM journal on imaging sciences},
  vol.~2, no.~1, pp. 183--202, 2009.

\end{thebibliography}

\newpage

\section*{Acknowledgement}
We thank Mr. Dongyu Meng for helping us understand MagNet training procedures and for suggesting possible ways to improve robustness, resulting in the robust MagNet variants used in this paper.

\section*{Supplementary Material: Complete Defense Performance Plots of MagNet}
Here we plot the defense performance of different variants of MagNet on MNIST and CIFAR-10. For each variant we also show four defense schemes: (1) no defense - plain DNN; (2) DNN with only MagNet detector(s); (3) DNN with only MagNet reformer; and (4) DNN with detector(s) and reformer. All experiments are performed in the oblivious attack setting.

\begin{figure*}
	\centering
	\subfloat[default]{
		\label{fig:CW}
		\includegraphics[scale=0.5]{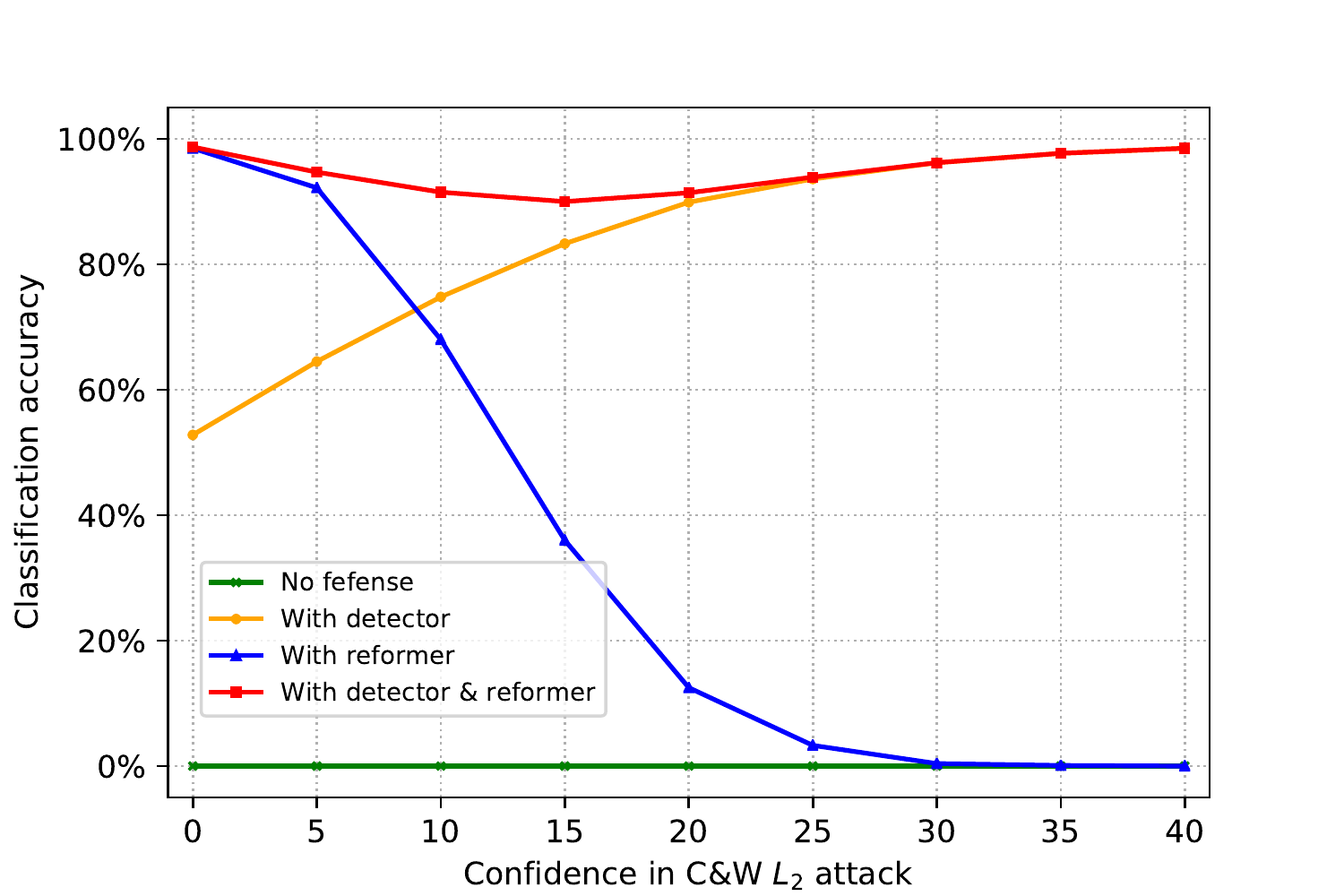}}
	\subfloat[JSD]{
		\label{fig:CW_JSD}
		\includegraphics[scale=0.5]{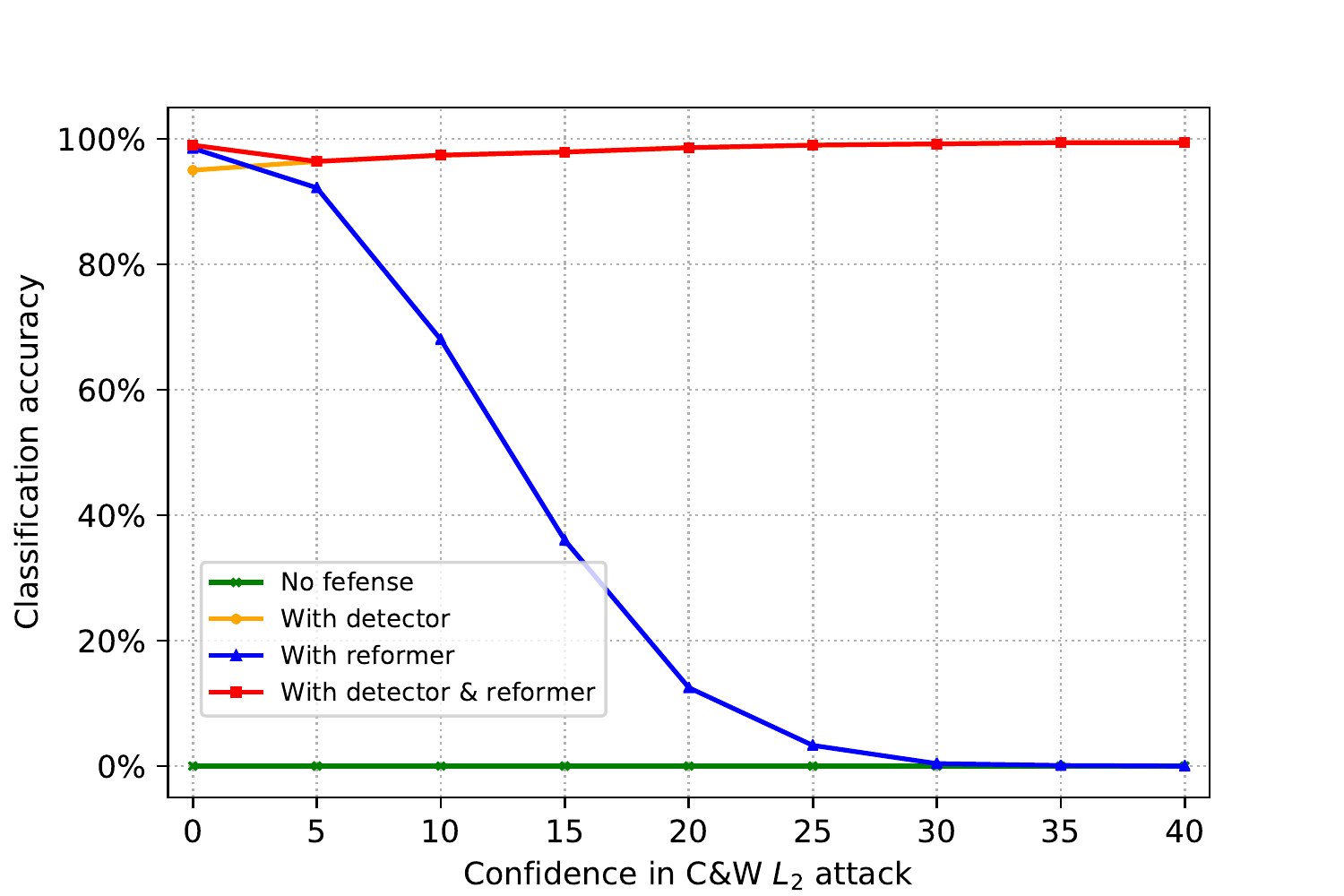}}	
	\\
	\subfloat[256]{
		\label{fig:CW_256}
		\includegraphics[scale=0.5]{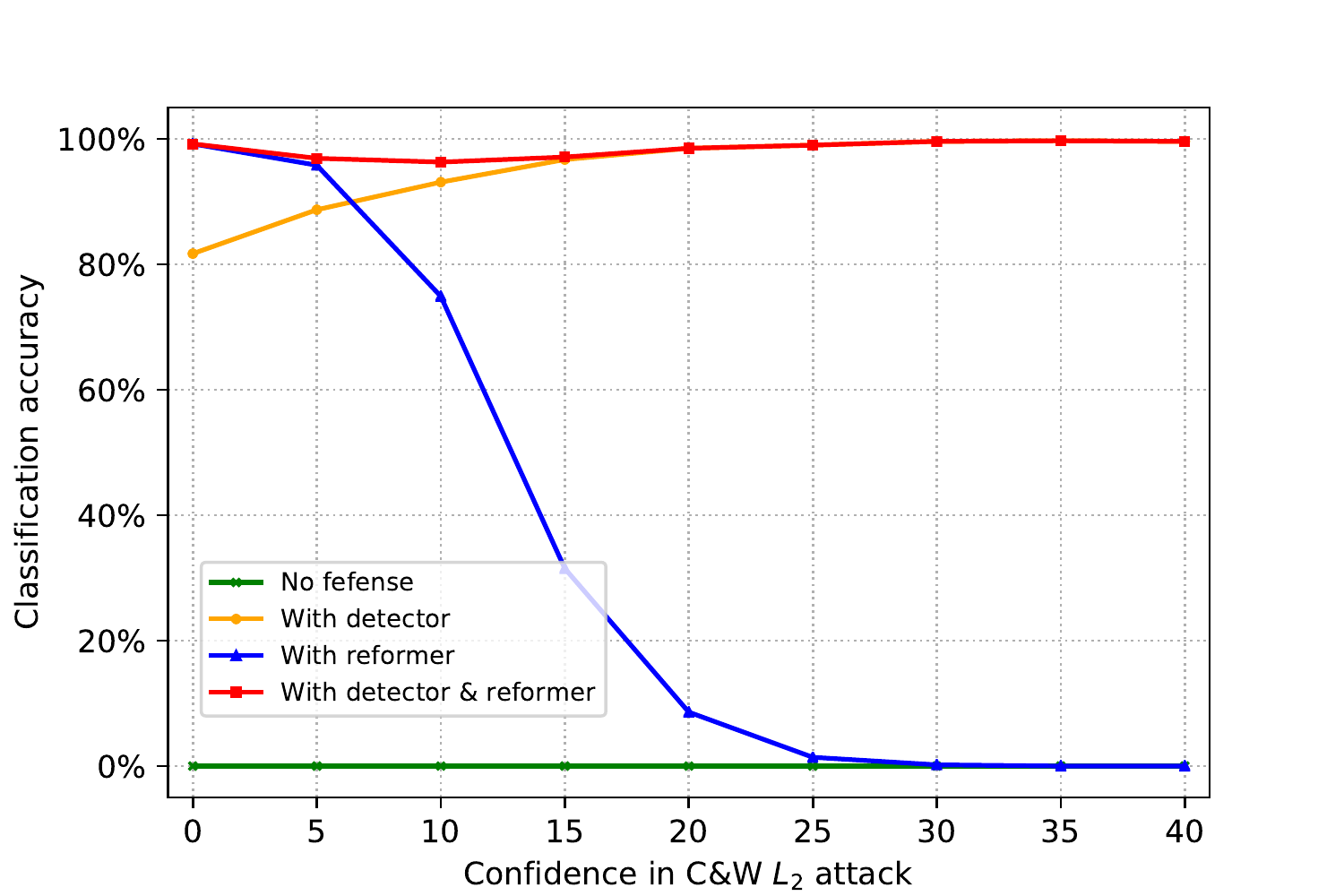}}
	\subfloat[256+JSD]{
		\label{fig:CW_256_JSD}
		\includegraphics[scale=0.5]{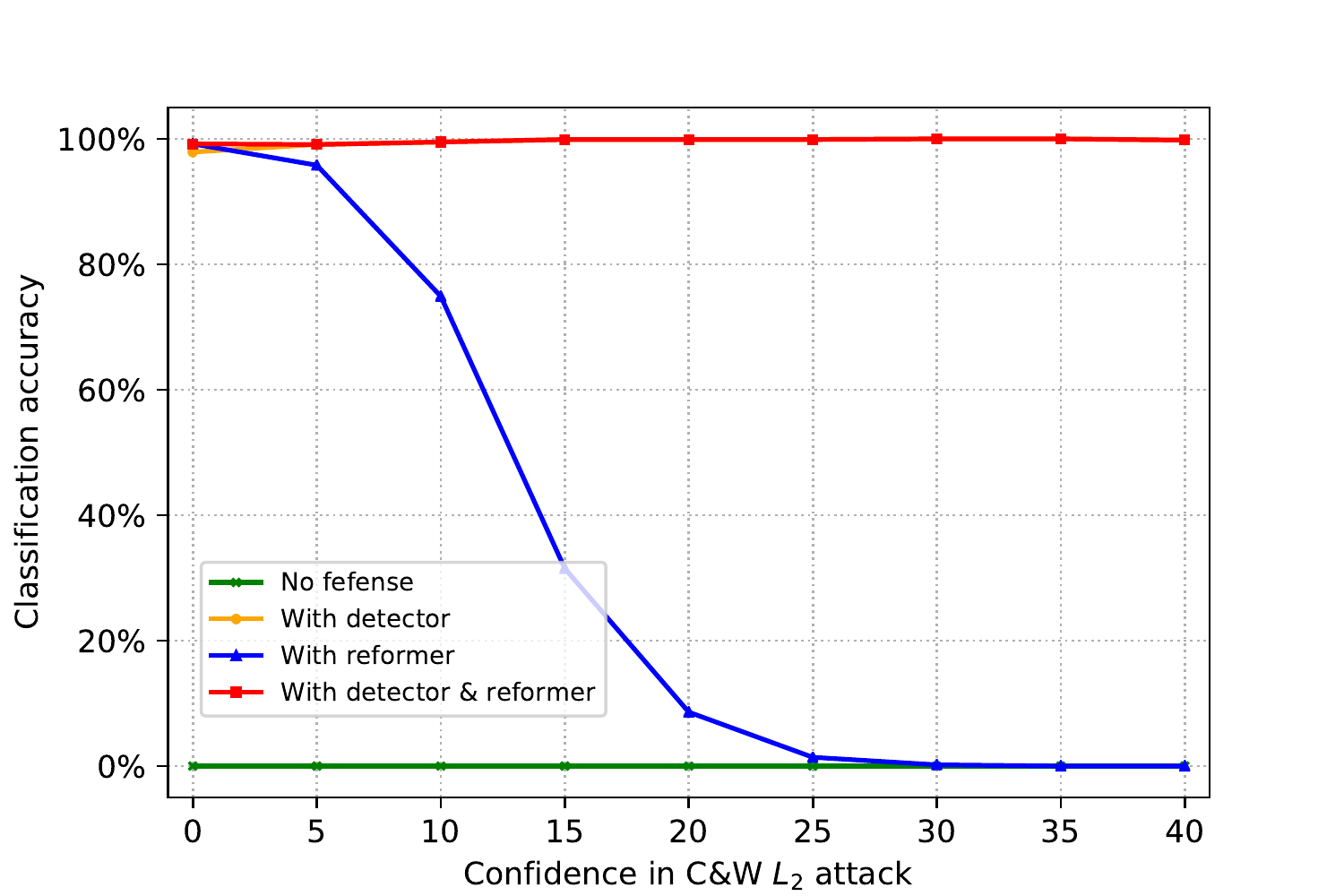}}
	\caption{C\&W $L_2$ attack to MagNet under different auto-encoder structure on MNIST dataset with varying confidence.}
	\label{fig:MNIST}
\end{figure*}

\begin{figure*}
	\centering
	\subfloat[default]{
		\label{fig:CIFAR_CW}
		\includegraphics[scale=0.5]{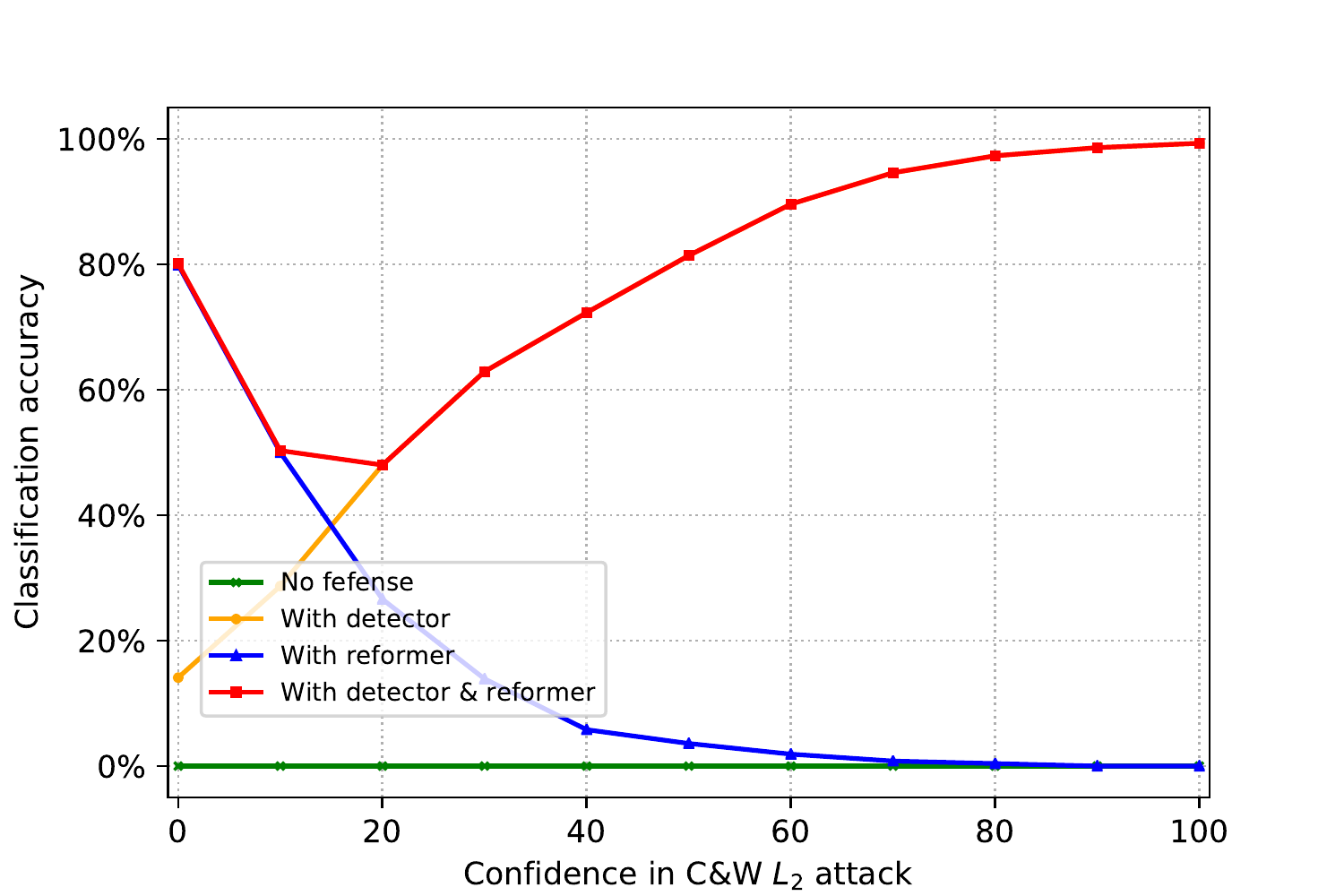}}
	\subfloat[256]{
		\label{fig:CIFAR_CW_256}
		\includegraphics[scale=0.5]{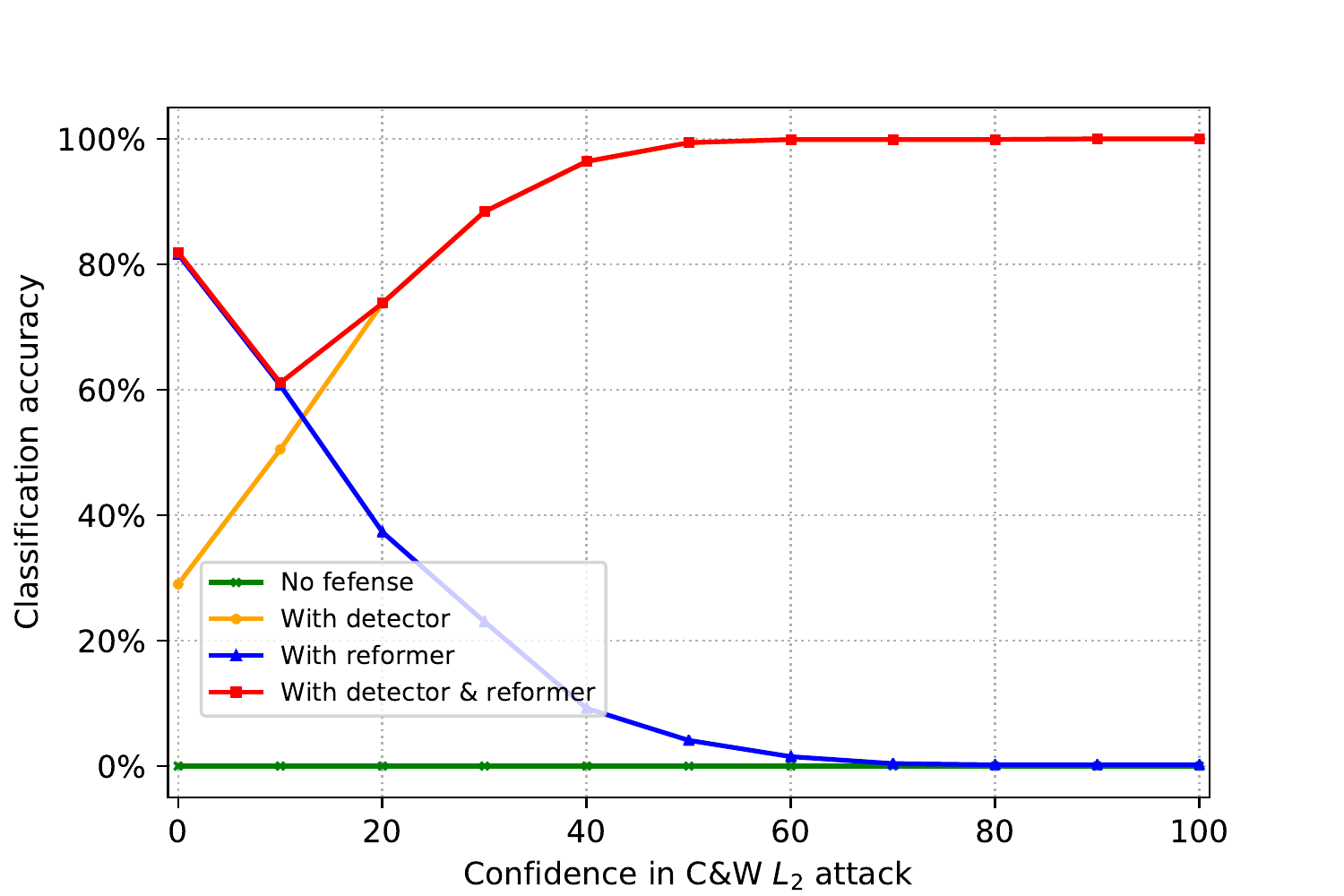}}
	\caption{C\&W $L_2$ attack to MagNet under different auto-encoder structure on CIFAR-10 dataset with varying confidence.}
	\label{fig:CIFAR}
\end{figure*}

\begin{figure*}
	\centering
	\subfloat[$L_1$ decision rule $\beta=10^{-3}$]{
		\label{fig:L1_e-3}
		\includegraphics[scale=0.5]{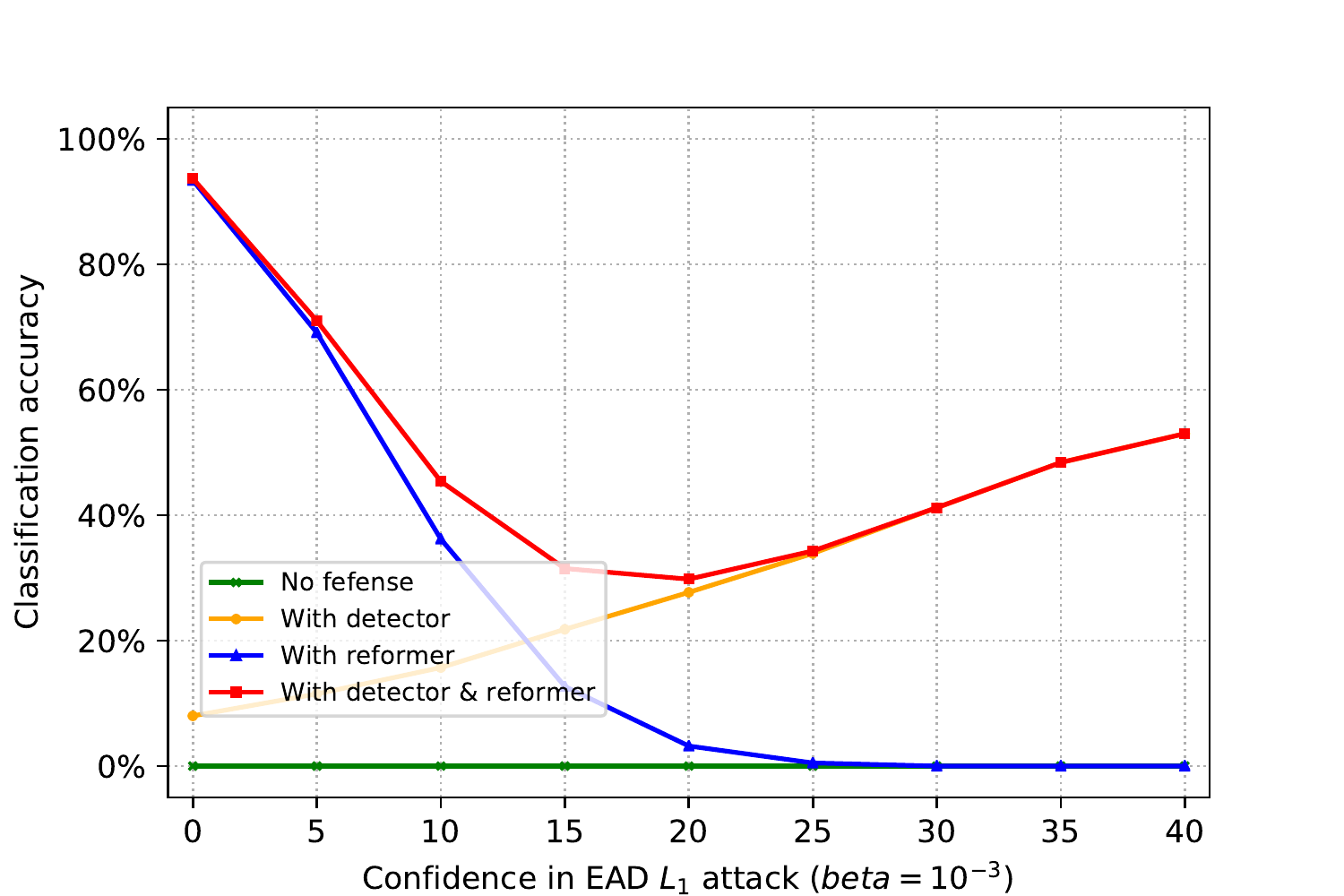}}
	\subfloat[EN decision rule $\beta=10^{-3}$]{
		\label{fig:EN_e-3}
		\includegraphics[scale=0.5]{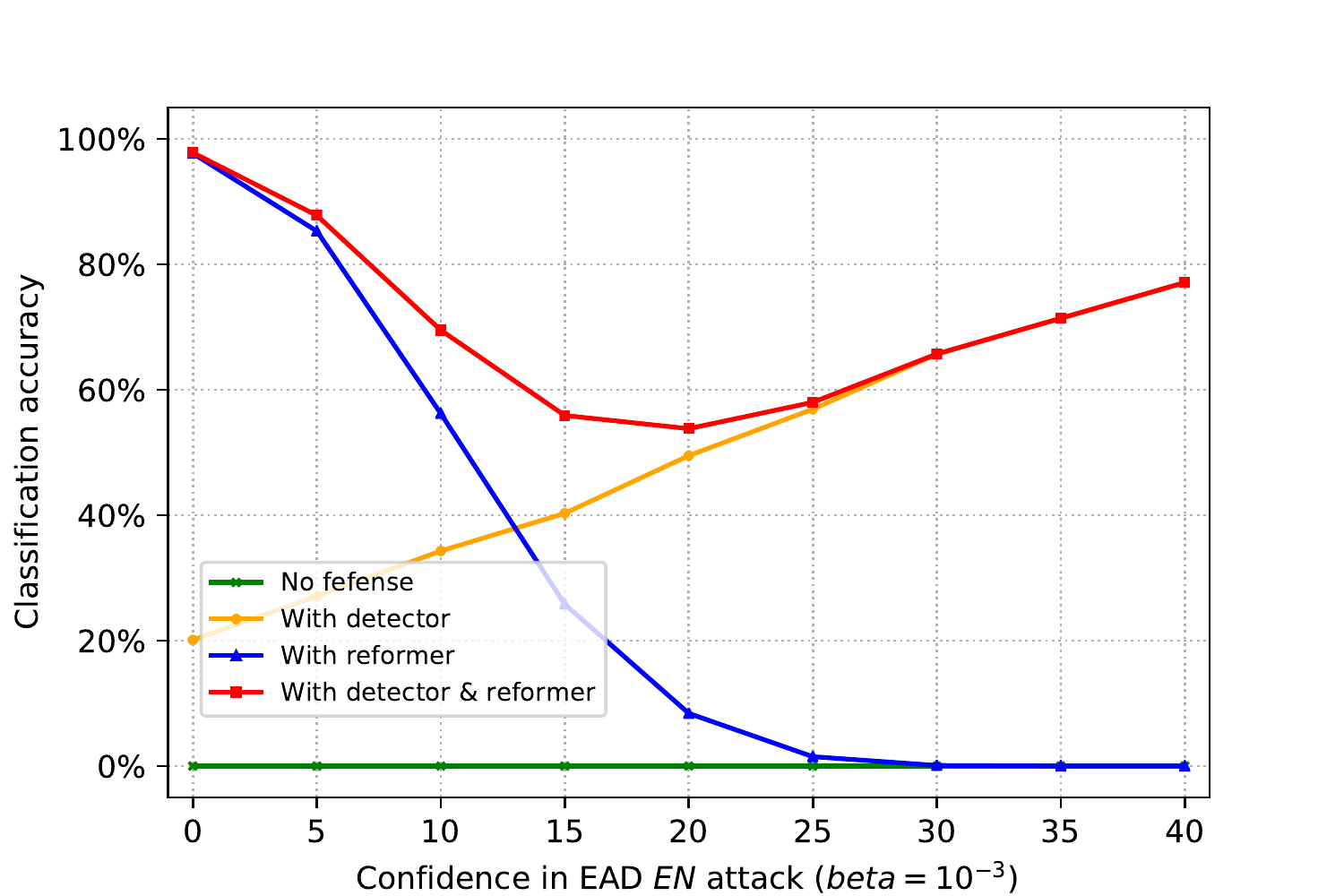}}
	\\
	\subfloat[$L_1$ decision rule $\beta=10^{-2}$]{
		\label{fig:L1_e-2}
		\includegraphics[scale=0.5]{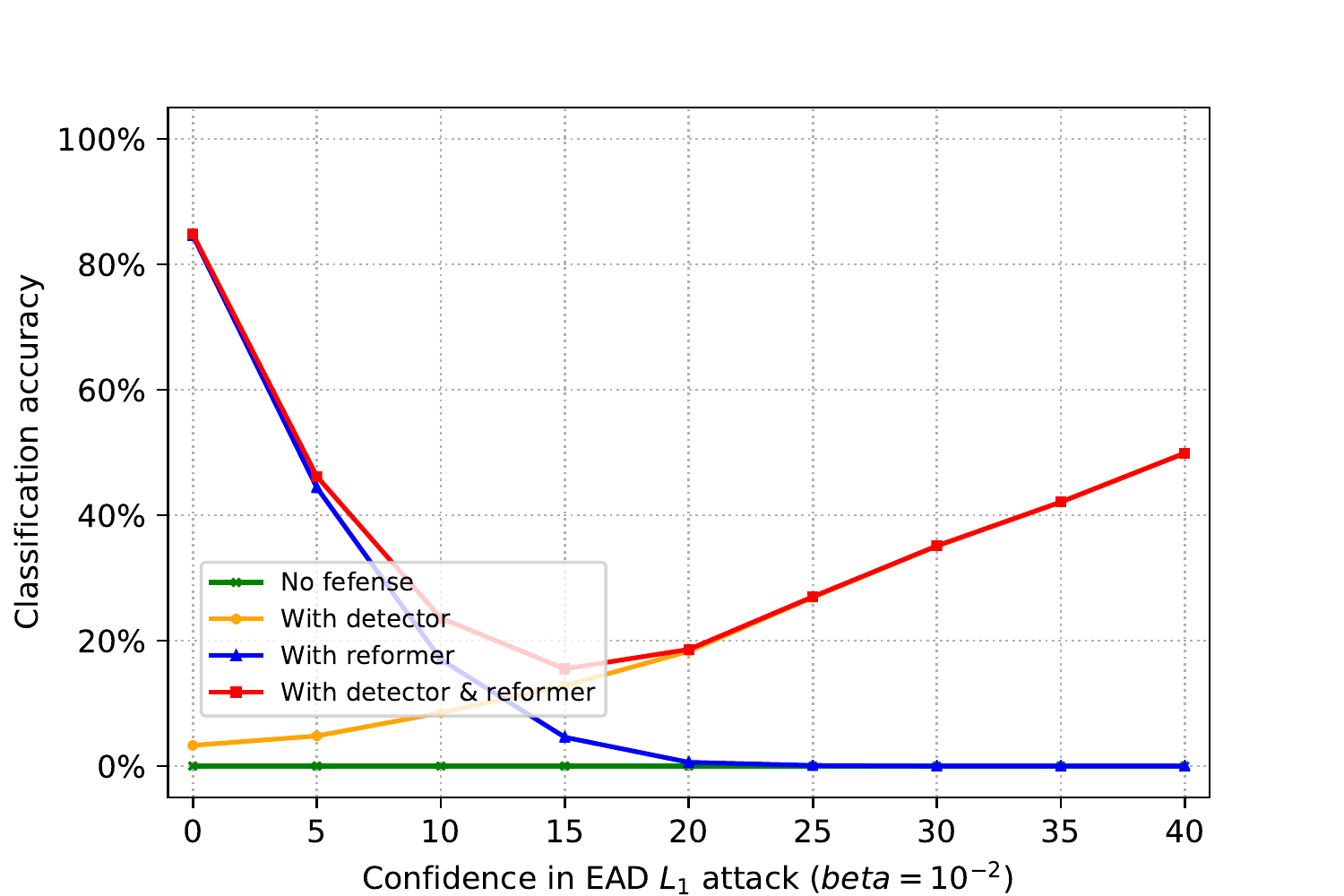}}
	\subfloat[EN decision rule $\beta=10^{-2}$]{
		\label{fig:EN_e-2}
		\includegraphics[scale=0.5]{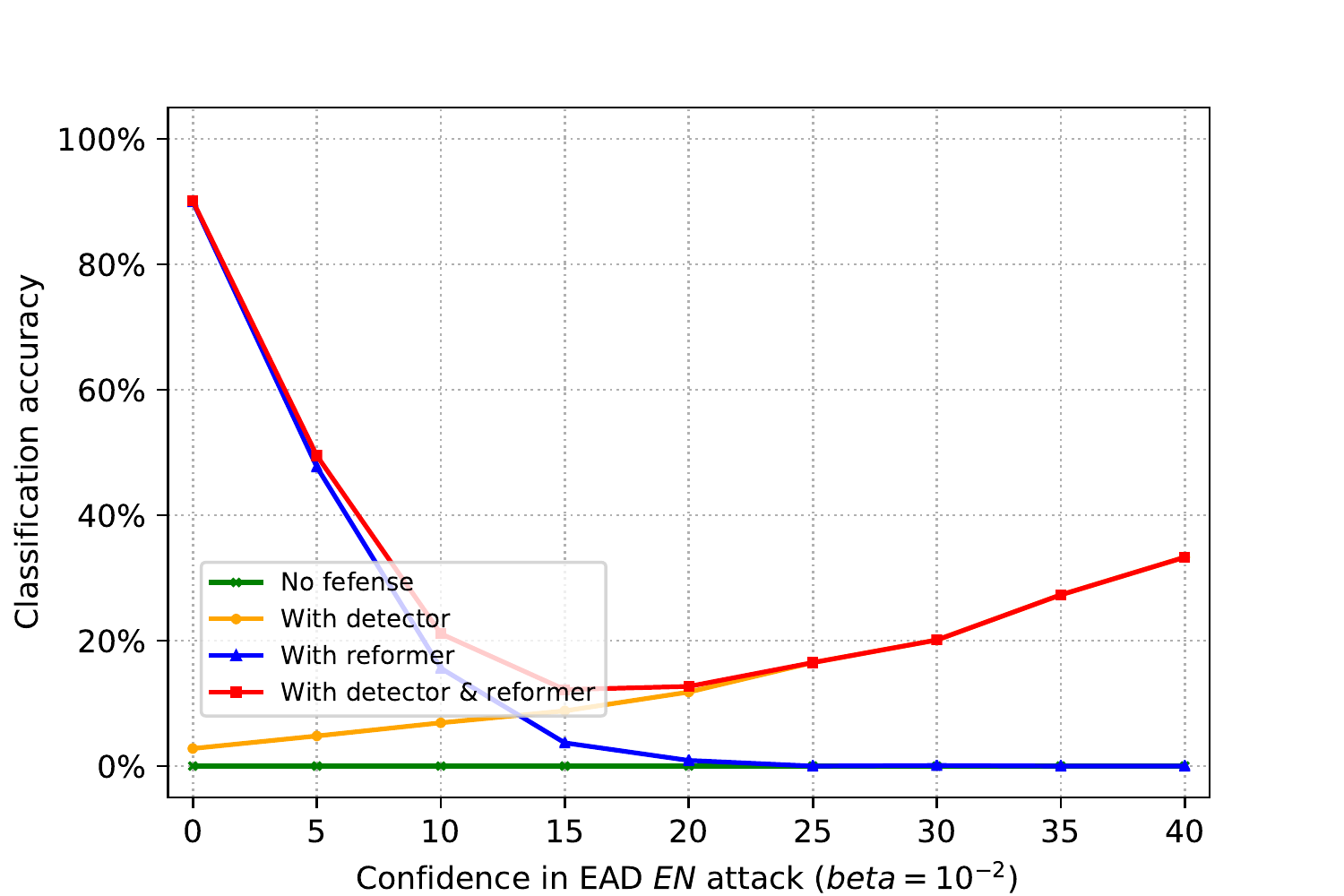}}
	\\
	\subfloat[$L_1$ decision rule $\beta=5\cdot10^{-2}$]{
		\label{fig:L1_5e-2}
		\includegraphics[scale=0.5]{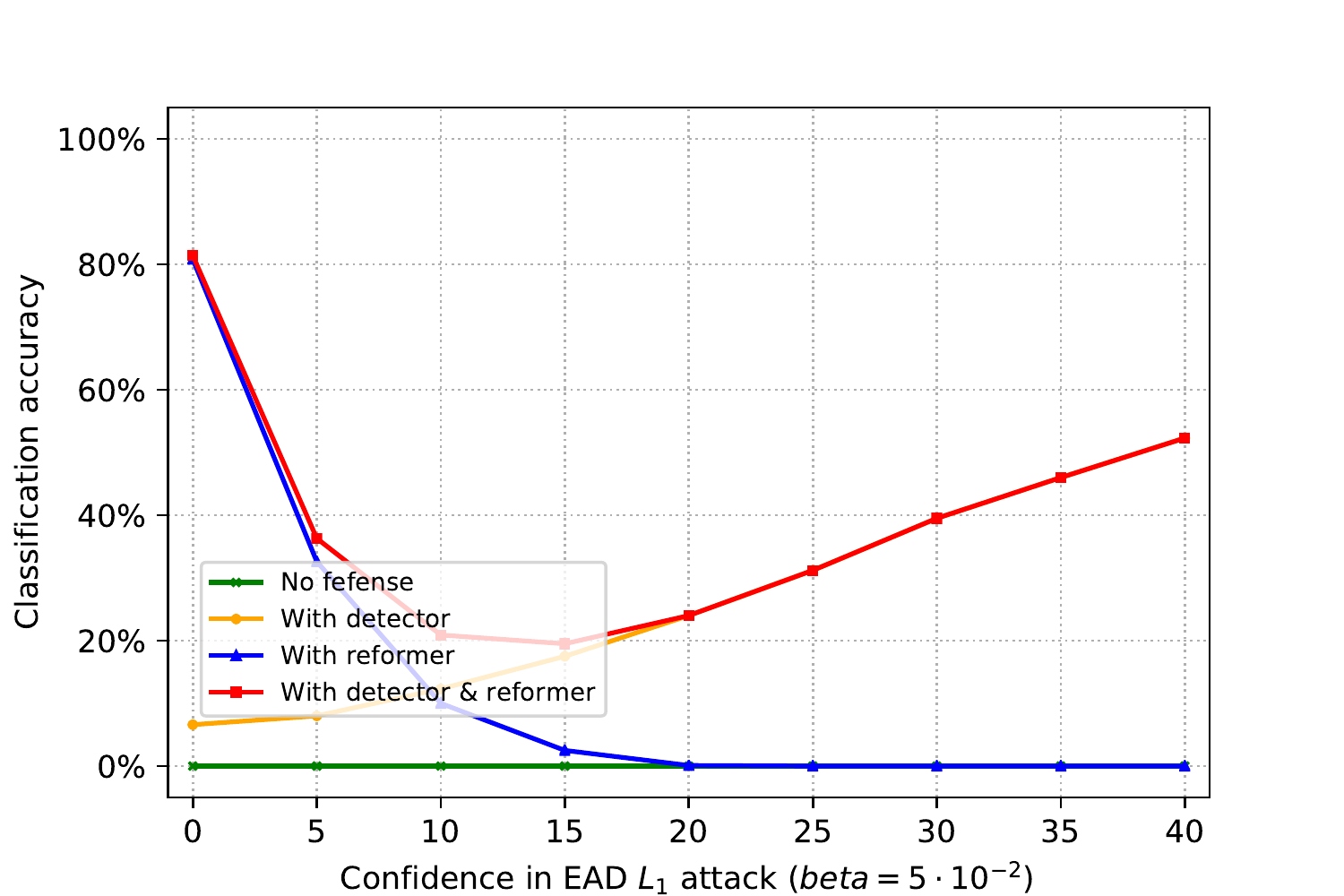}}
	\subfloat[EN decision rule $\beta=5\cdot10^{-2}$]{
		\label{fig:EN_5e-2}
		\includegraphics[scale=0.5]{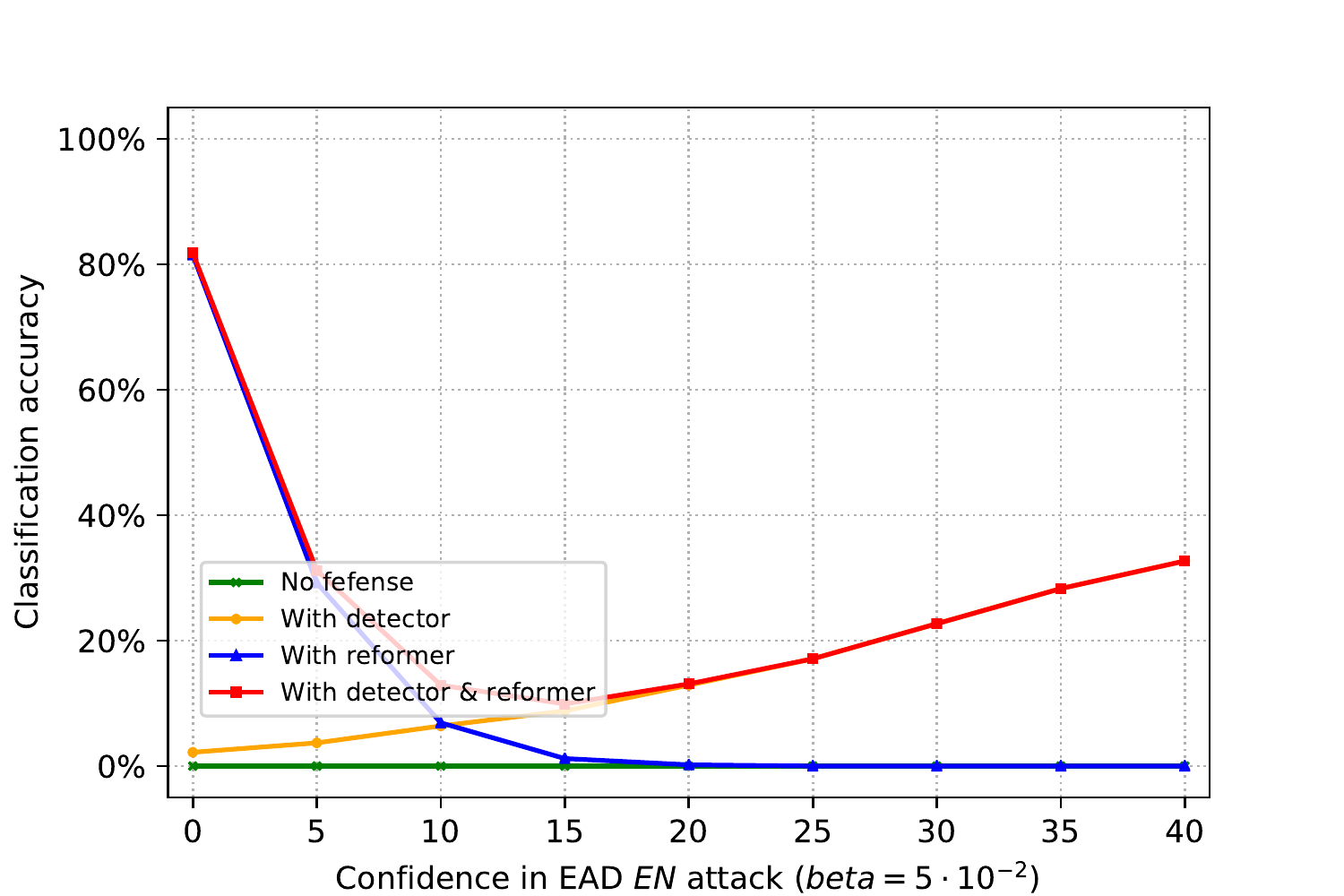}}
	\\	
	\subfloat[$L_1$ decision rule $\beta=10^{-1}$]{
		\label{fig:L1_e-1}
		\includegraphics[scale=0.5]{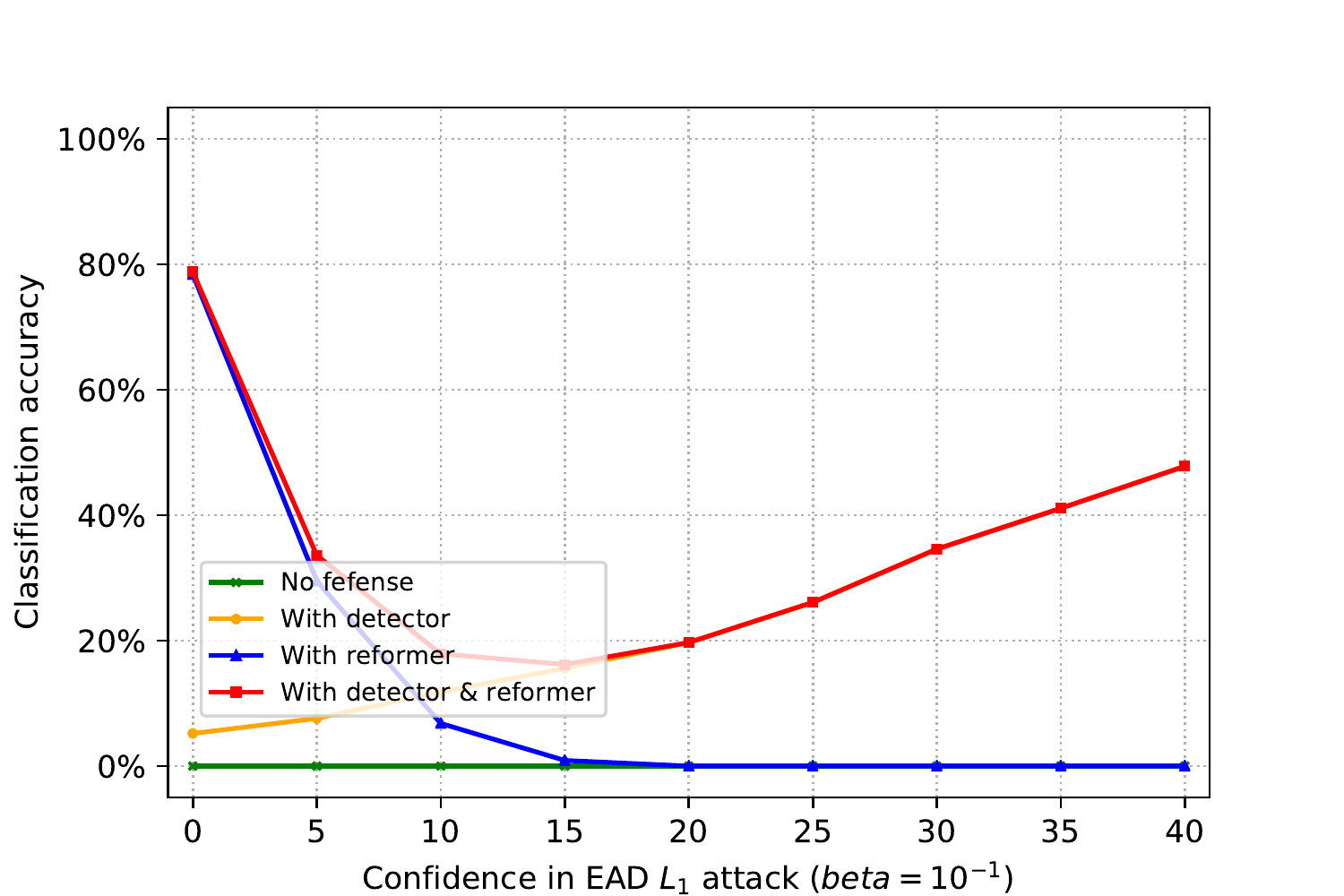}}
	\subfloat[EN decision rule $\beta=10^{-1}$]{
		\label{fig:EN_e-1}
		\includegraphics[scale=0.5]{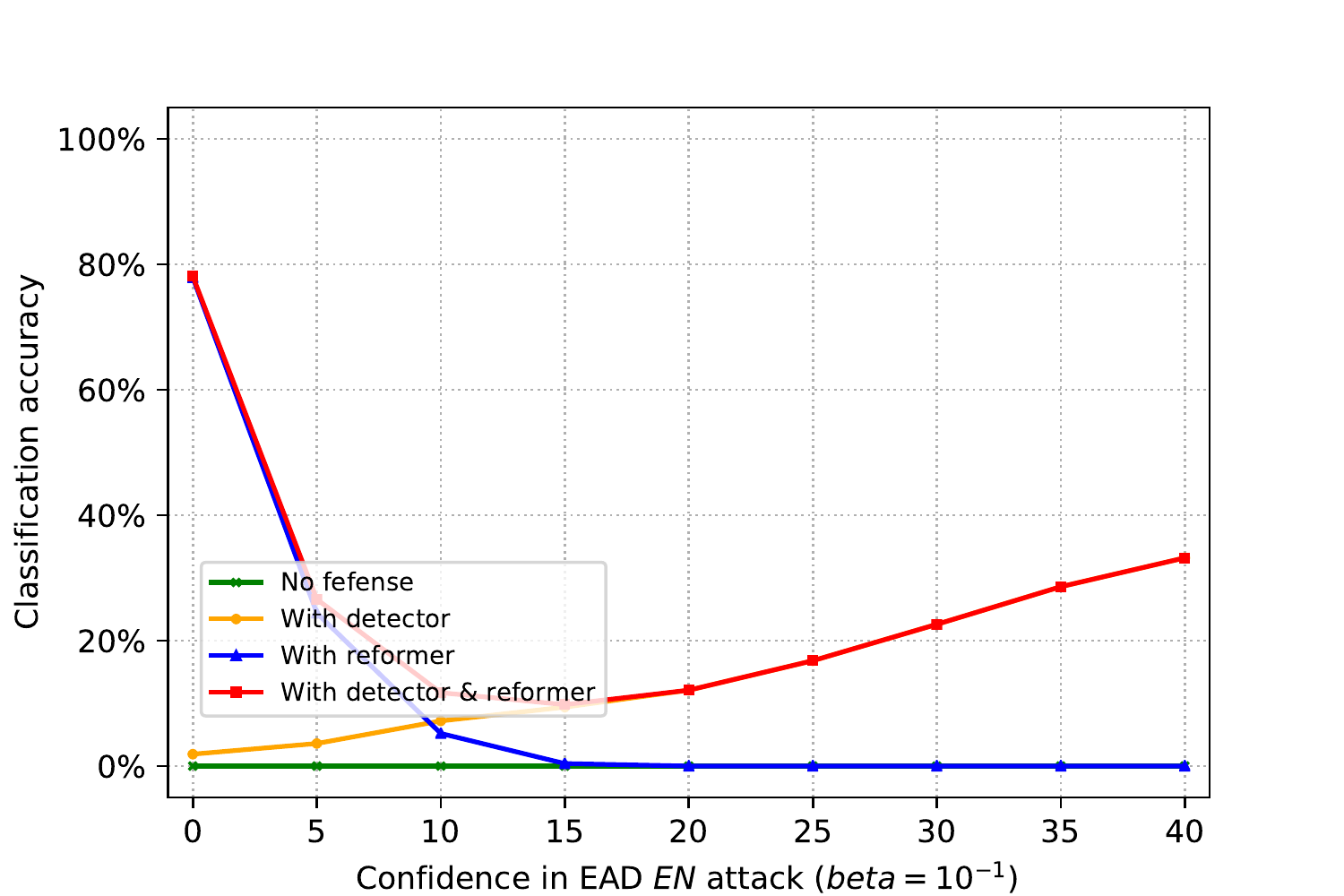}}
	\caption{EAD attacks on default MagNet under different $\beta$ and different decision rules on MNIST with varying confidence.}
	\label{fig:MNIST_default}
\end{figure*}

\begin{figure*}
	\centering
	\subfloat[$L_1$ decision rule $\beta=10^{-3}$]{
		\label{fig:CIFAR_L1_e-3}
		\includegraphics[scale=0.5]{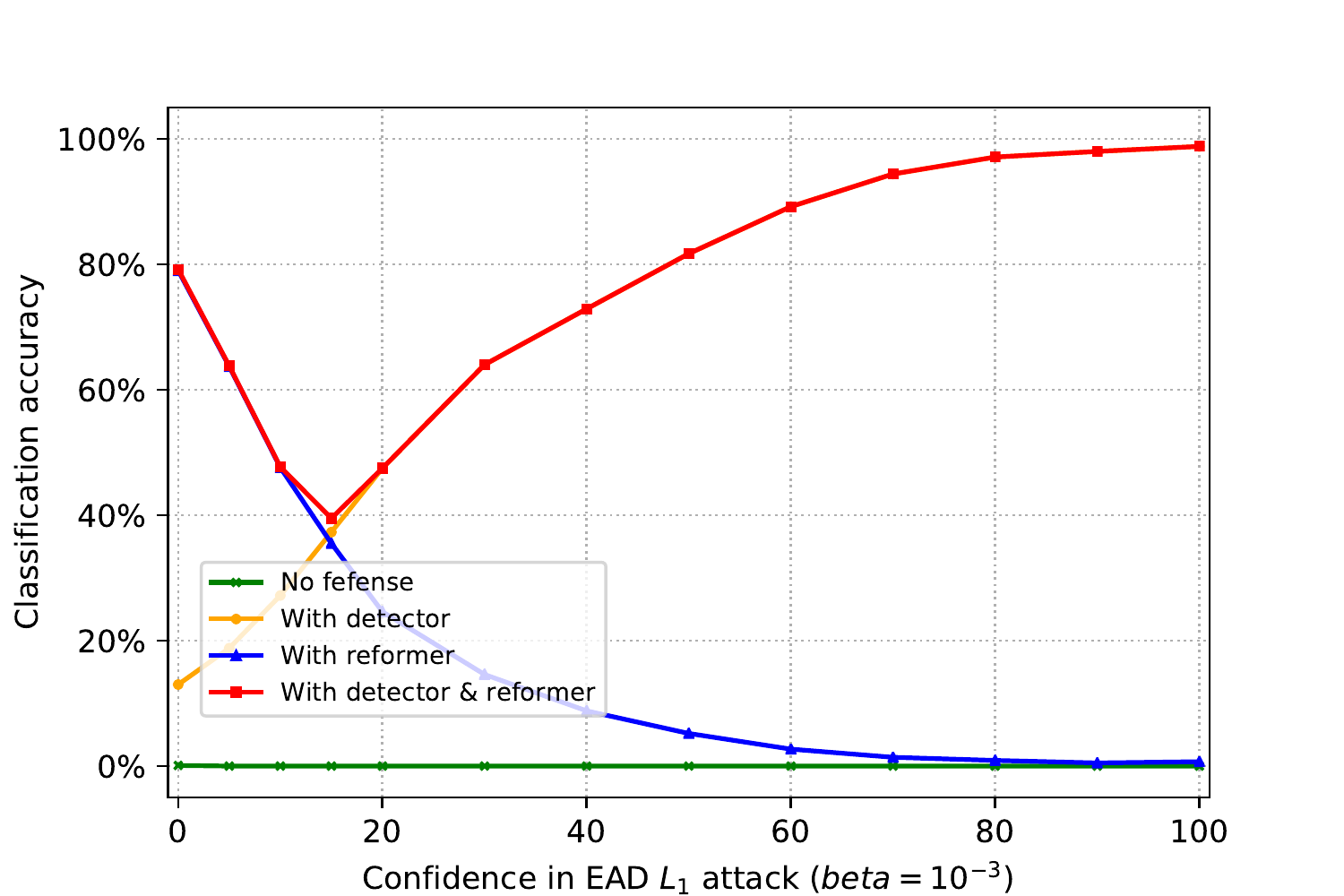}}
	\subfloat[EN decision rule $\beta=10^{-3}$]{
		\label{fig:CIFAR_EN_e-3}
		\includegraphics[scale=0.5]{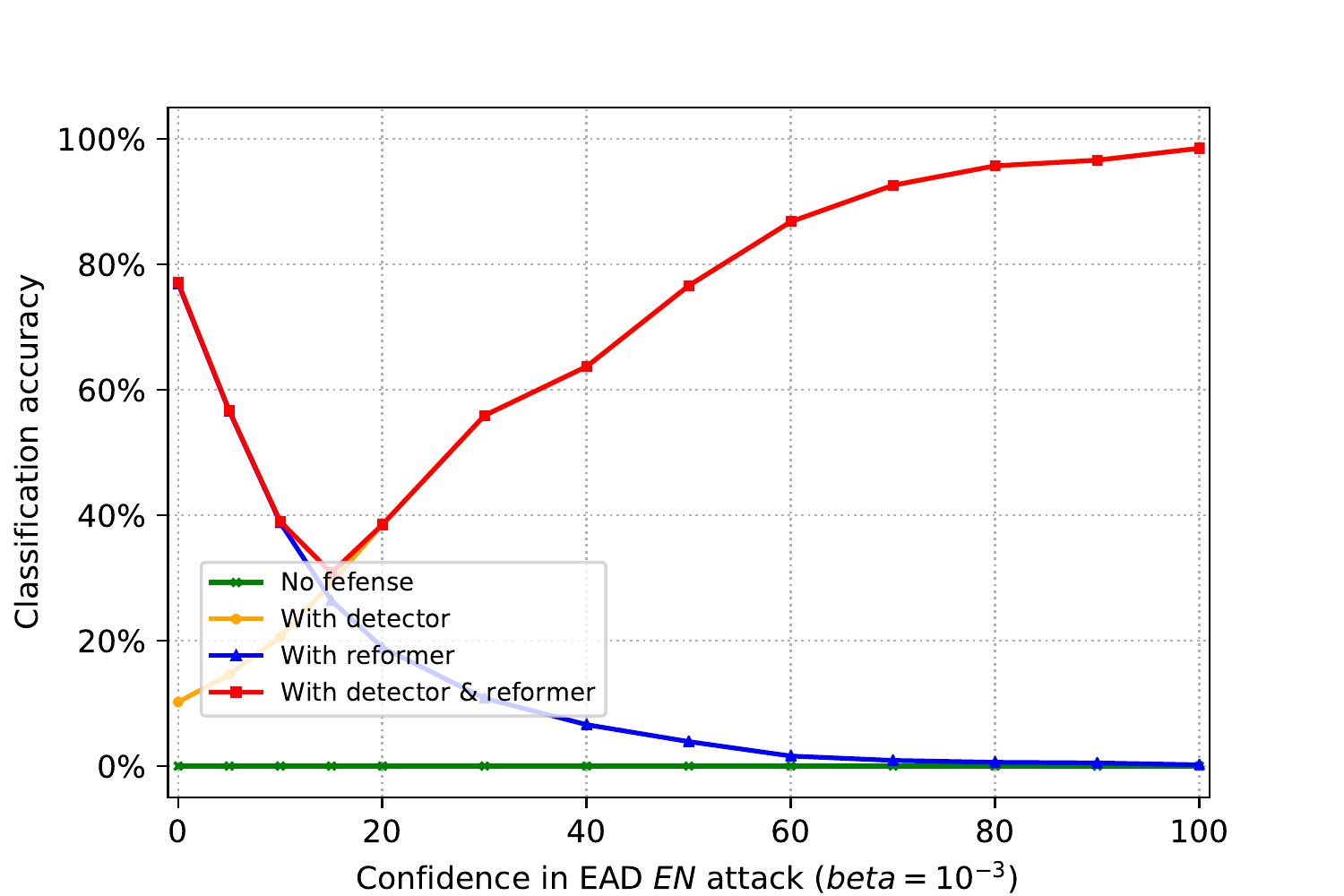}}
	\\
	\subfloat[$L_1$ decision rule $\beta=10^{-2}$]{
		\label{fig:CIFAR_L1_e-2}
		\includegraphics[scale=0.5]{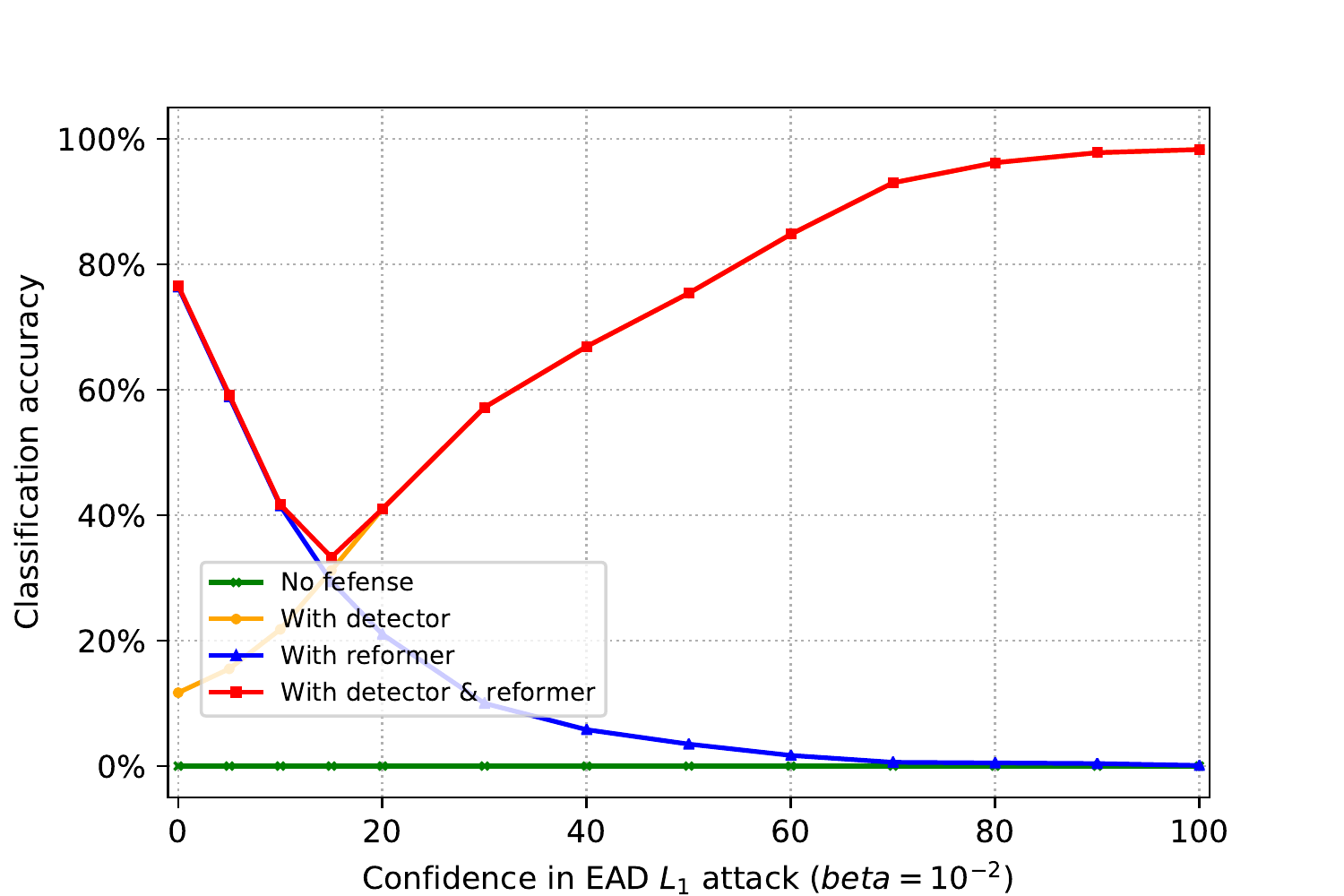}}
	\subfloat[EN decision rule $\beta=10^{-2}$]{
		\label{fig:CIFAR_EN_e-2}
		\includegraphics[scale=0.5]{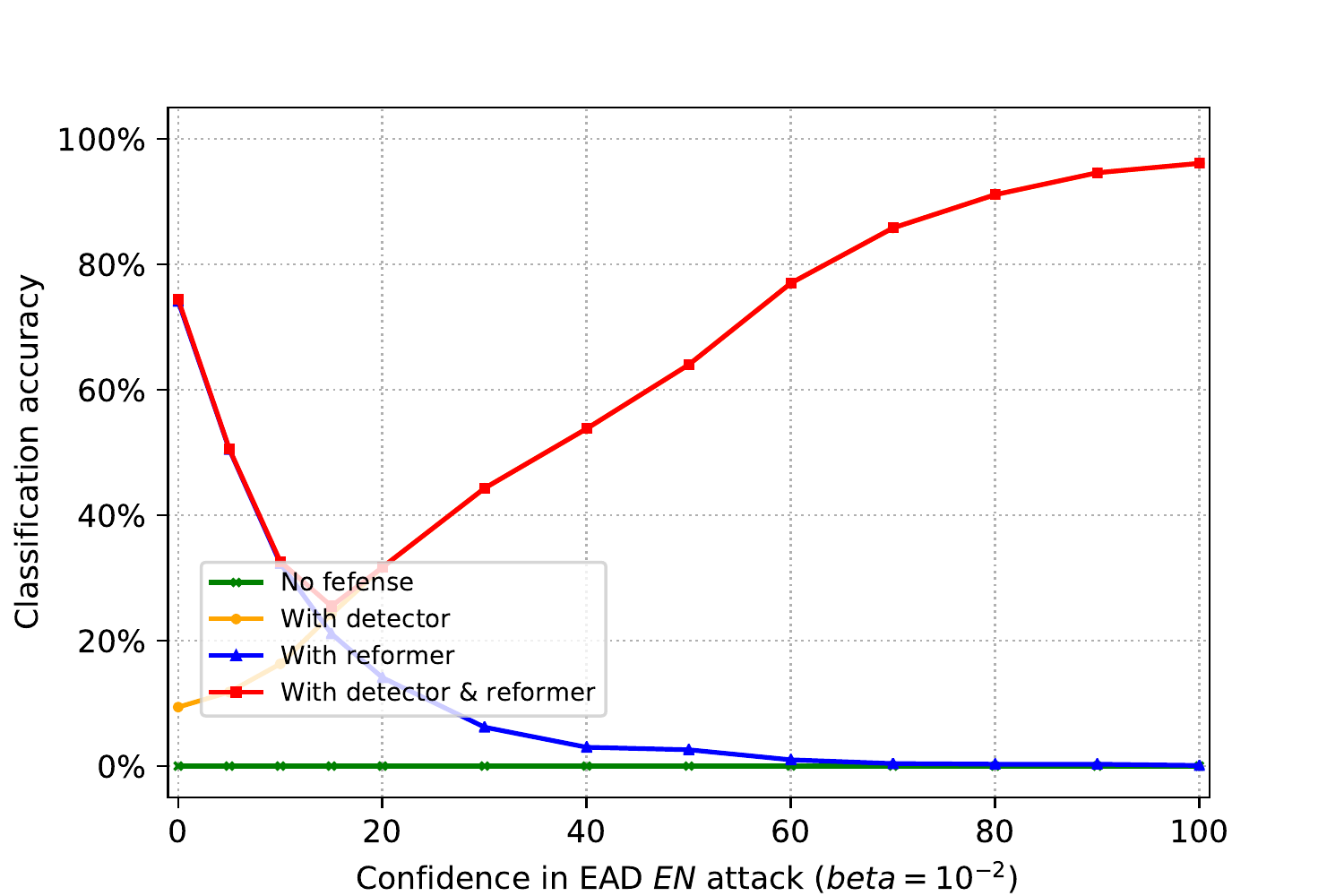}}
	\\
	\subfloat[$L_1$ decision rule $\beta=5\cdot10^{-2}$]{
		\label{fig:CIFAR_L1_5e-2}
		\includegraphics[scale=0.5]{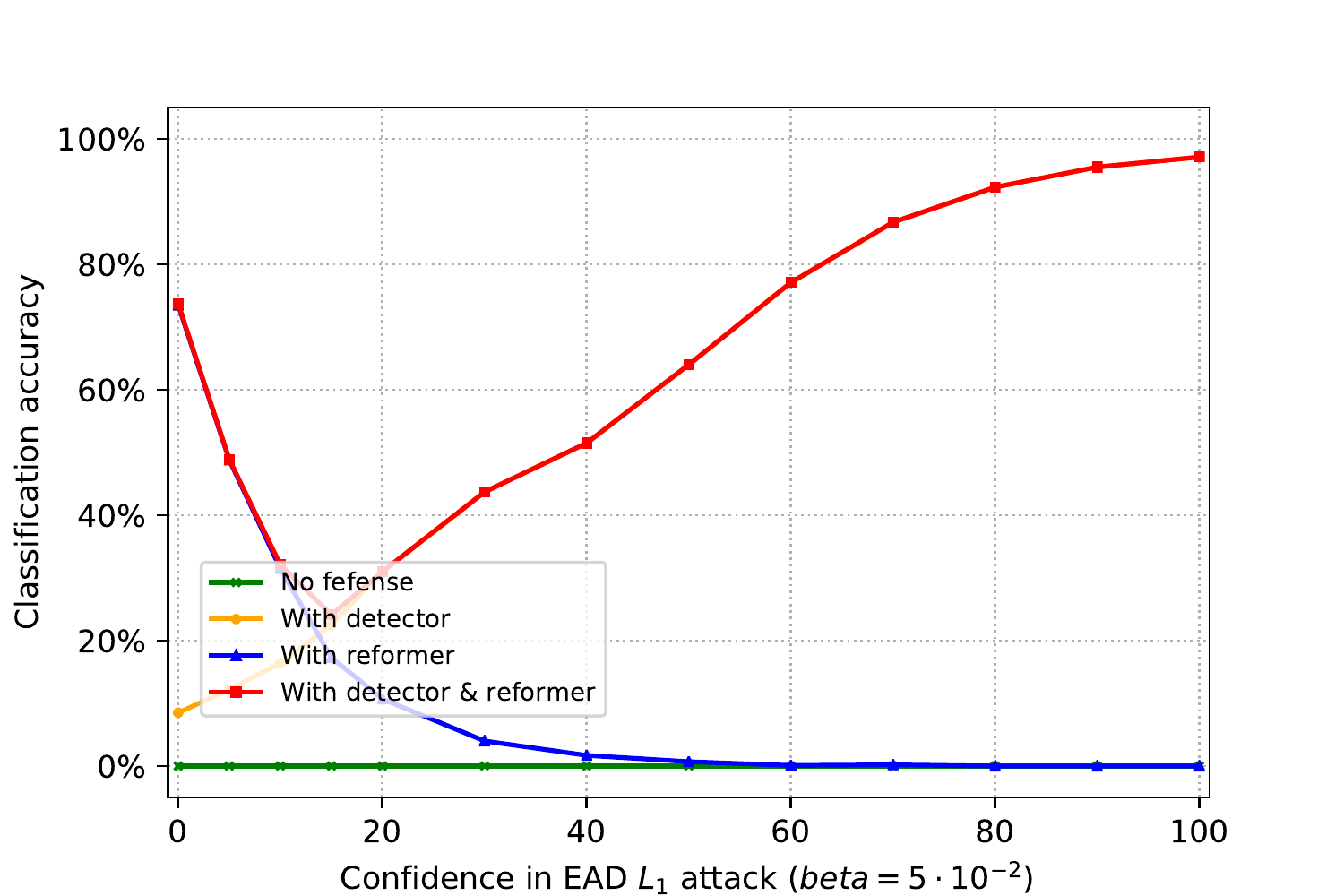}}
	\subfloat[EN decision rule $\beta=5\cdot10^{-2}$]{
		\label{fig:CIFAR_EN_5e-2}
		\includegraphics[scale=0.5]{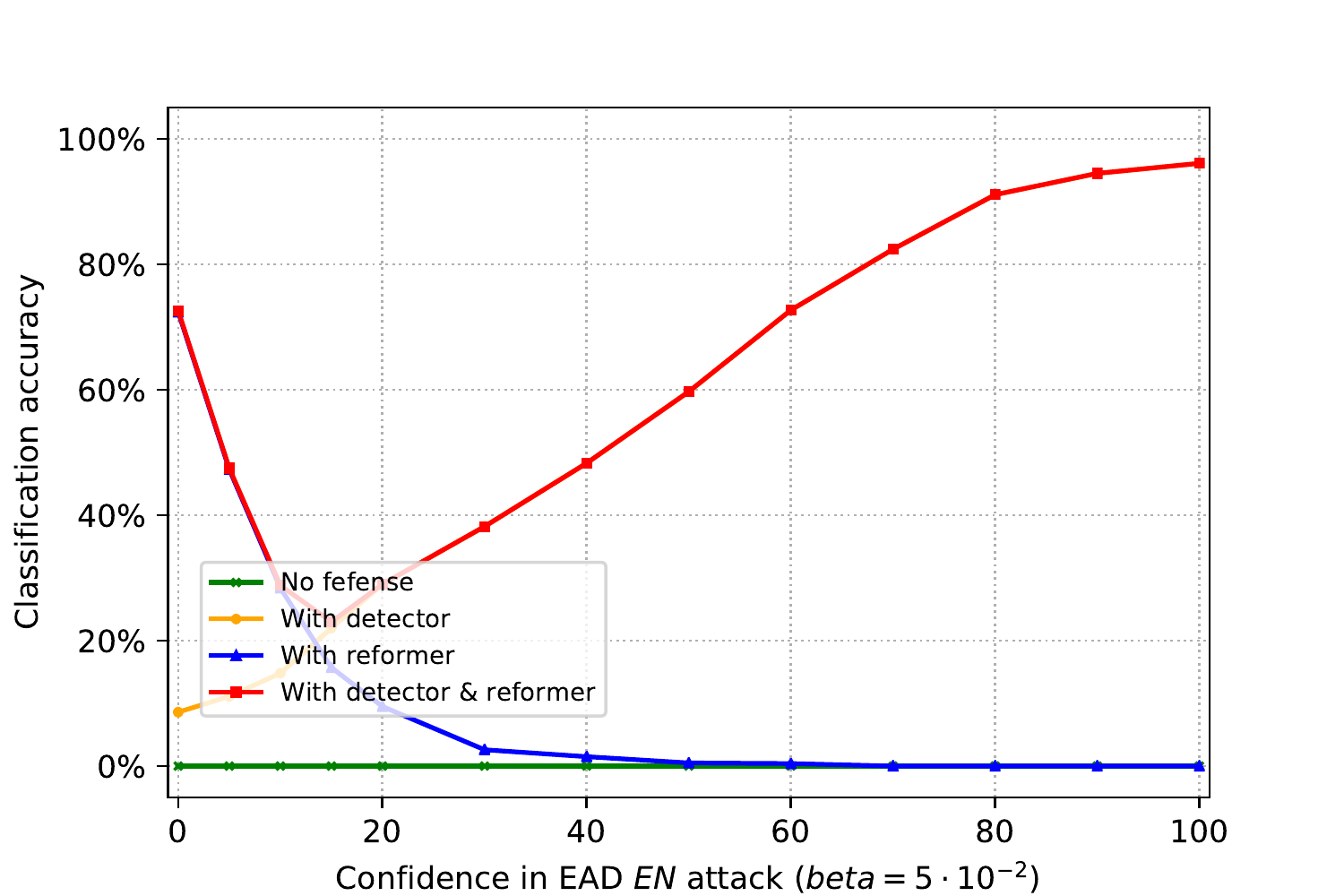}}
	\\	
	\subfloat[$L_1$ decision rule $\beta=10^{-1}$]{
		\label{fig:CIFAR_L1_e-1}
		\includegraphics[scale=0.5]{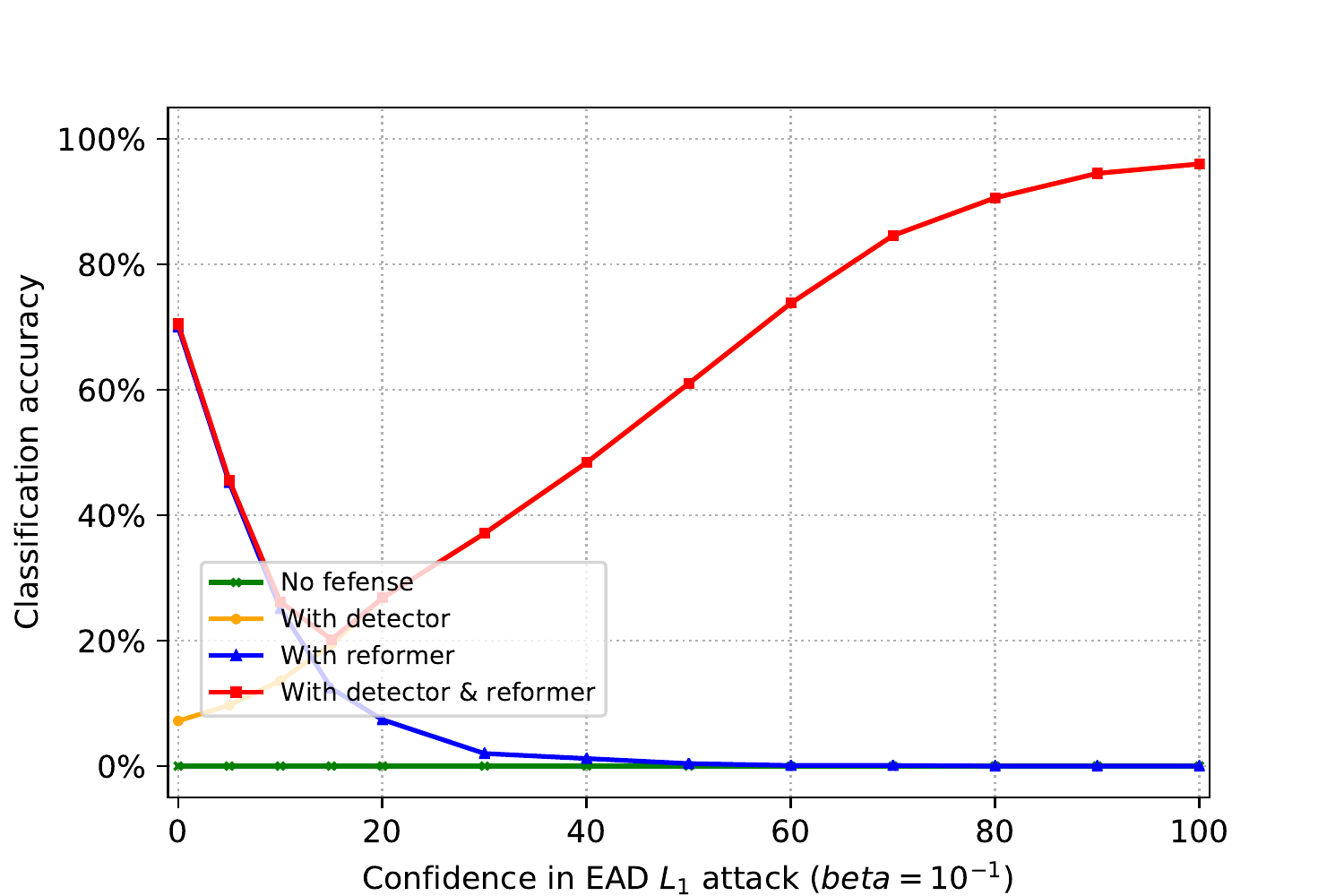}}
	\subfloat[EN decision rule $\beta=10^{-1}$]{
		\label{fig:CIFAR_EN_e-1}
		\includegraphics[scale=0.5]{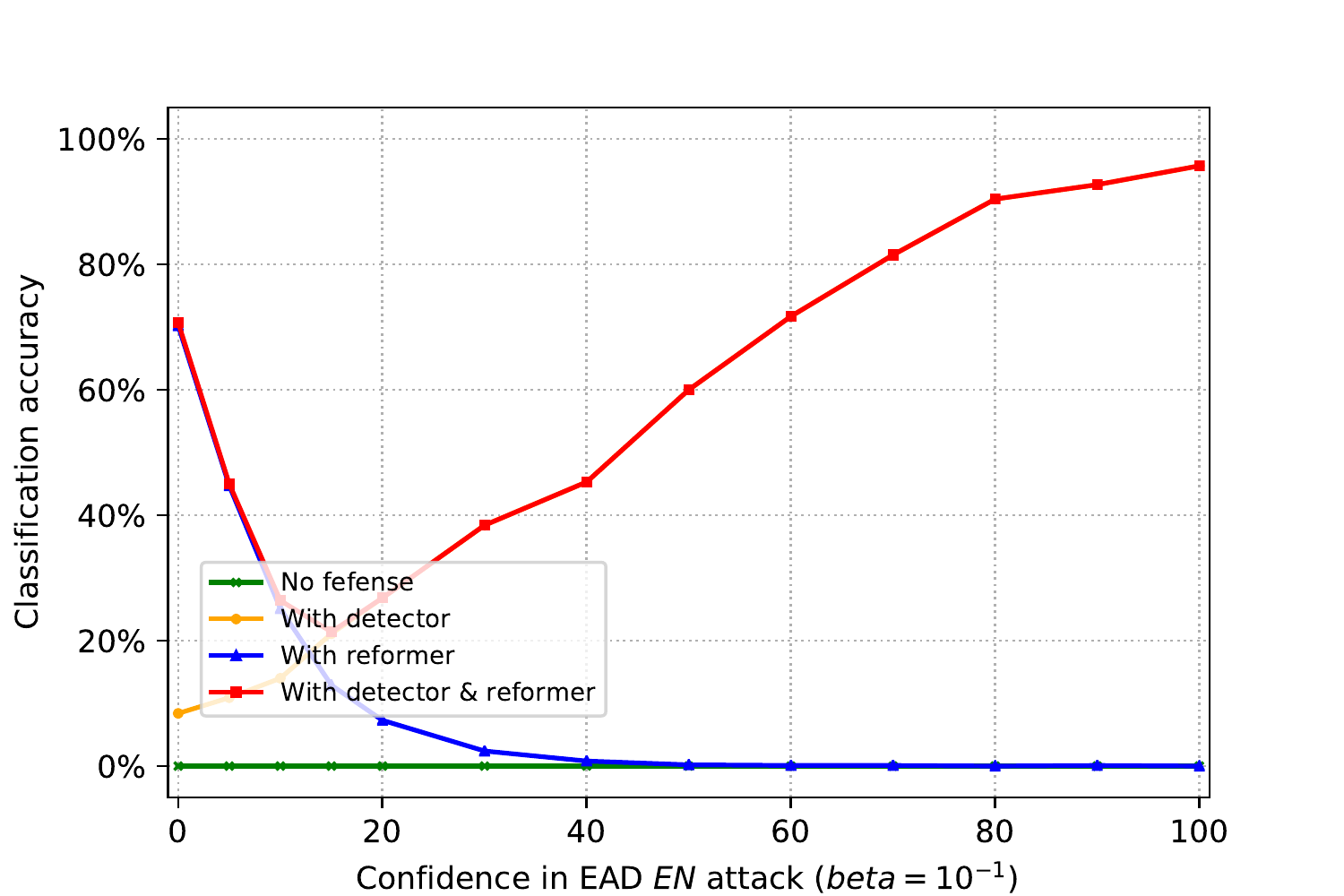}}
	\caption{EAD attacks on default MagNet under different $\beta$ and different decision rules on CIFAR-10  with varying confidence.}
	\label{fig:CIFAR_default}
\end{figure*}

\begin{figure*}
	\centering
	\subfloat[$L_1$ decision rule $\beta=10^{-3}$]{
		\label{fig:L1_e-3_JSD}
		\includegraphics[scale=0.5]{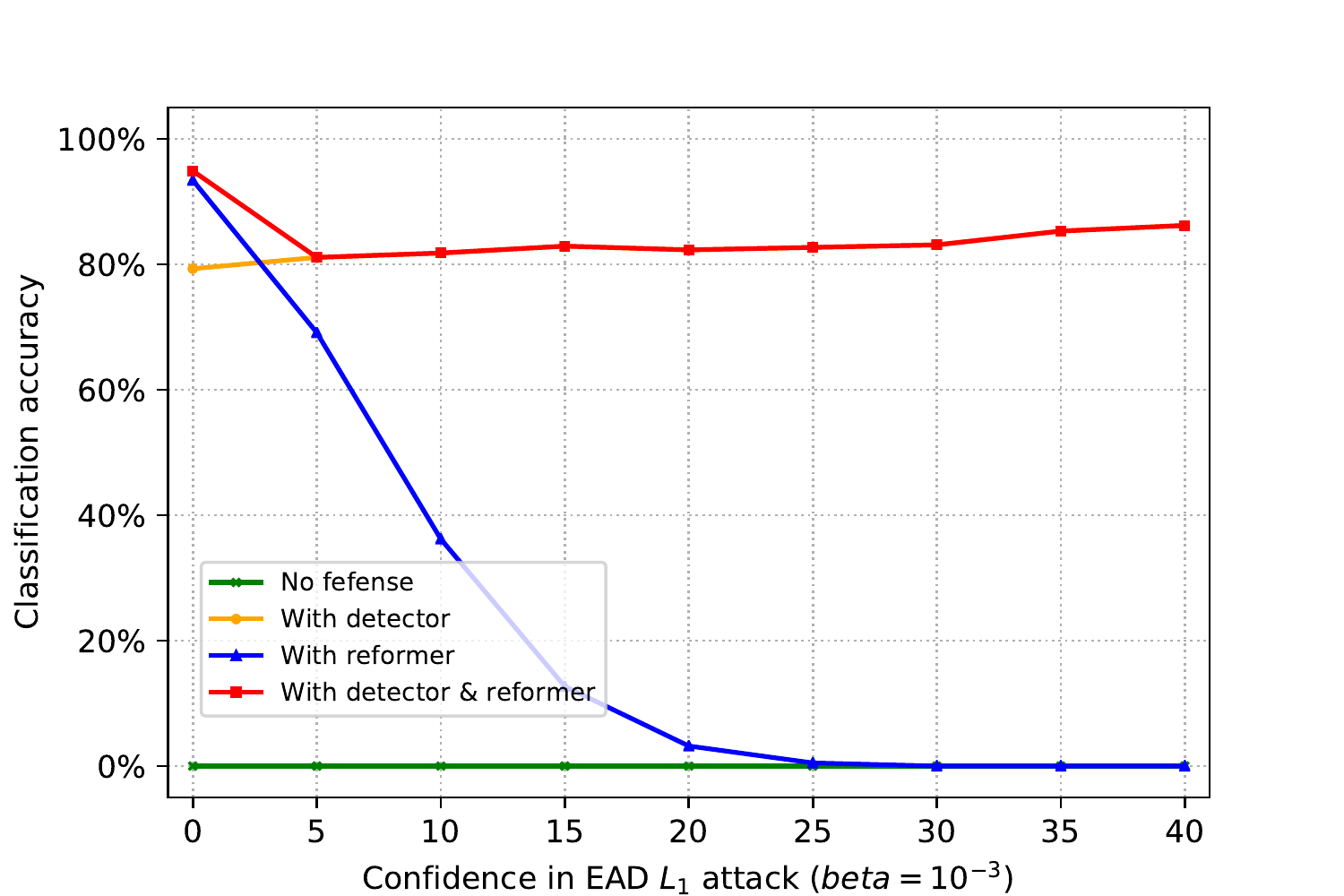}}
	\subfloat[EN decision rule $\beta=10^{-3}$]{
		\label{fig:EN_e-3_JSD}
		\includegraphics[scale=0.5]{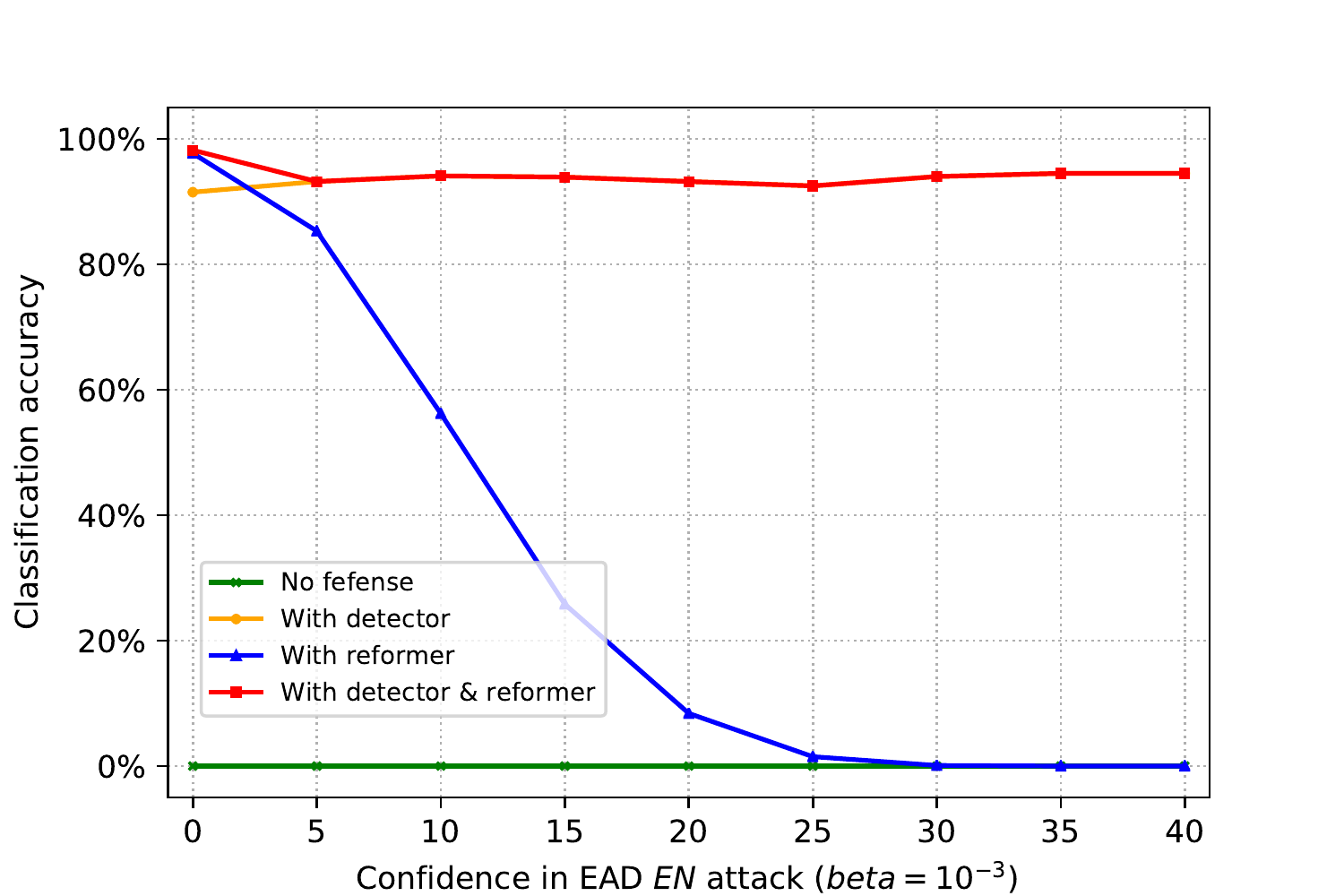}}
	\\
	\subfloat[$L_1$ decision rule $\beta=10^{-2}$]{
		\label{fig:L1_e-2_JSD}
		\includegraphics[scale=0.5]{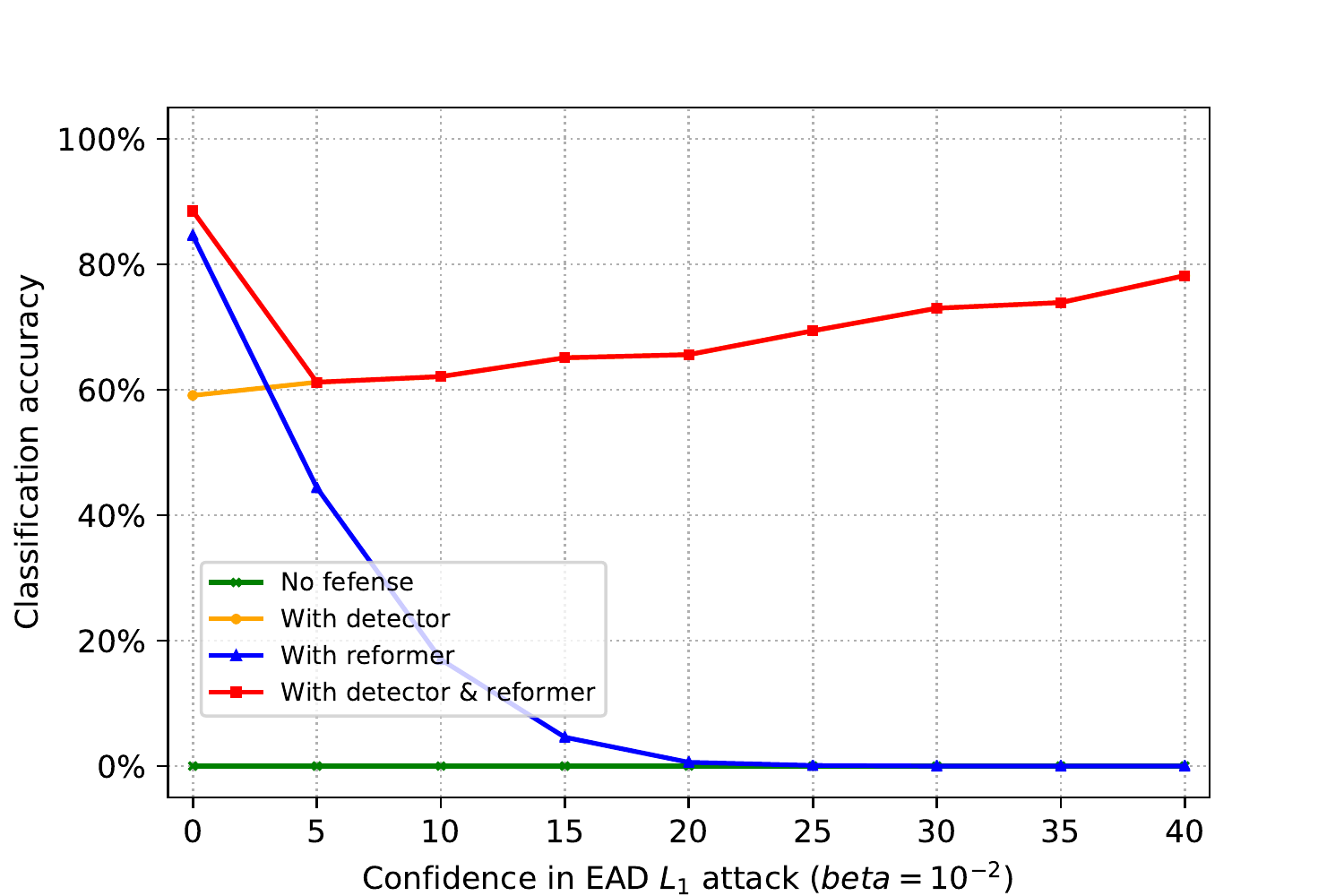}}
	\subfloat[EN decision rule $\beta=10^{-2}$]{
		\label{fig:EN_e-2_JSD}
		\includegraphics[scale=0.5]{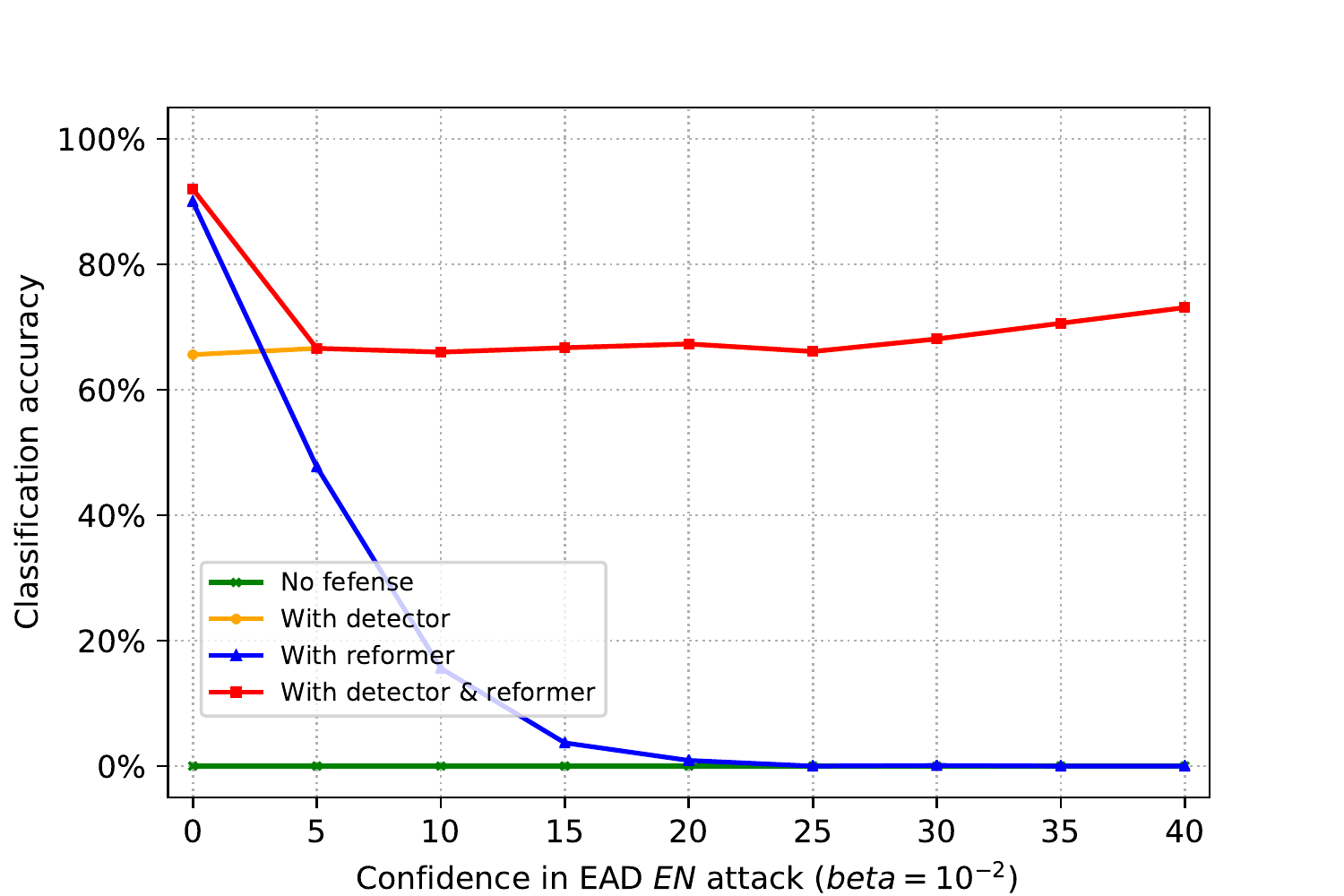}}
	\\
	\subfloat[$L_1$ decision rule $\beta=5\cdot10^{-2}$]{
		\label{fig:L1_5e-2_JSD}
		\includegraphics[scale=0.5]{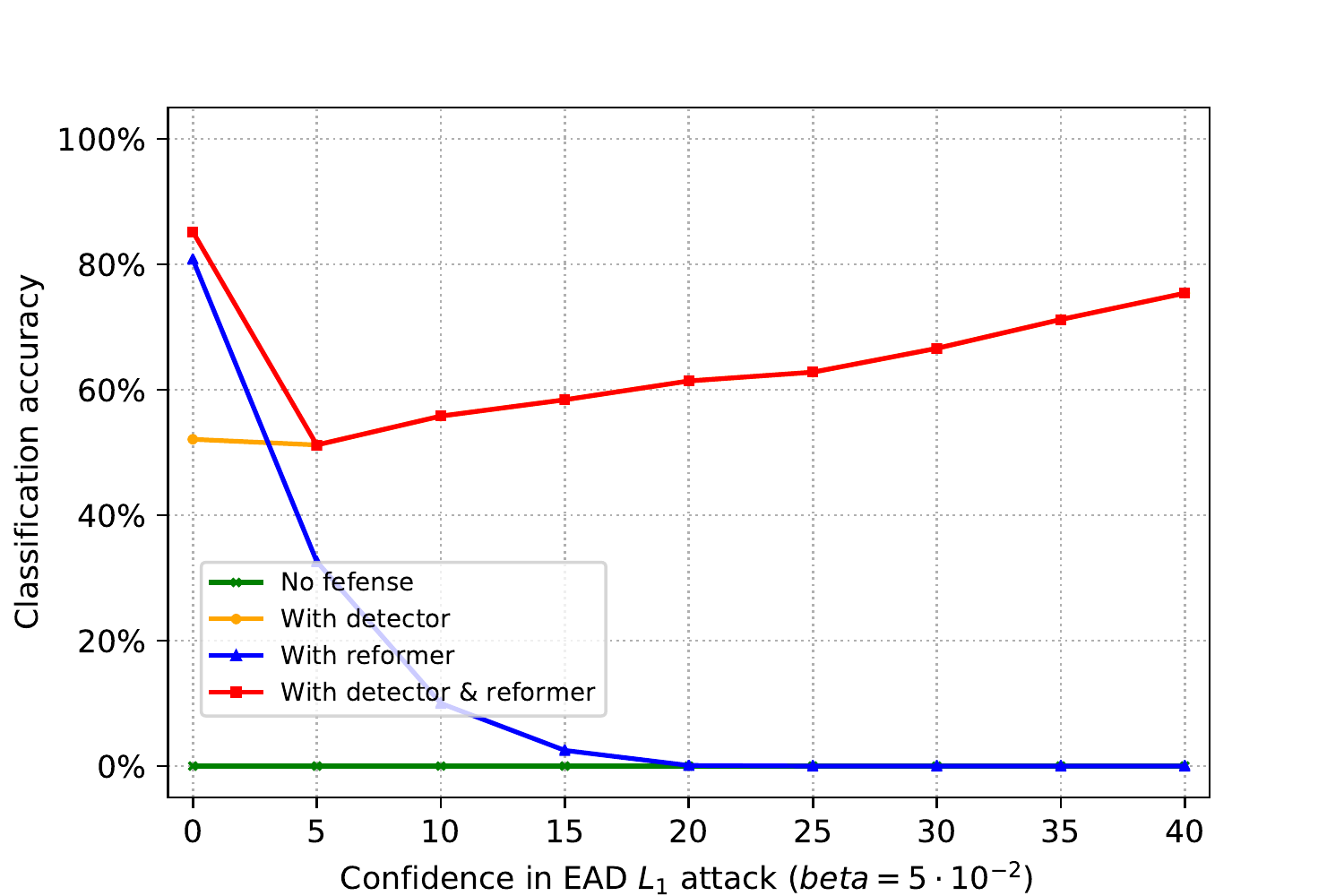}}
	\subfloat[EN decision rule $\beta=5\cdot10^{-2}$]{
		\label{fig:EN_5e-2_JSD}
		\includegraphics[scale=0.5]{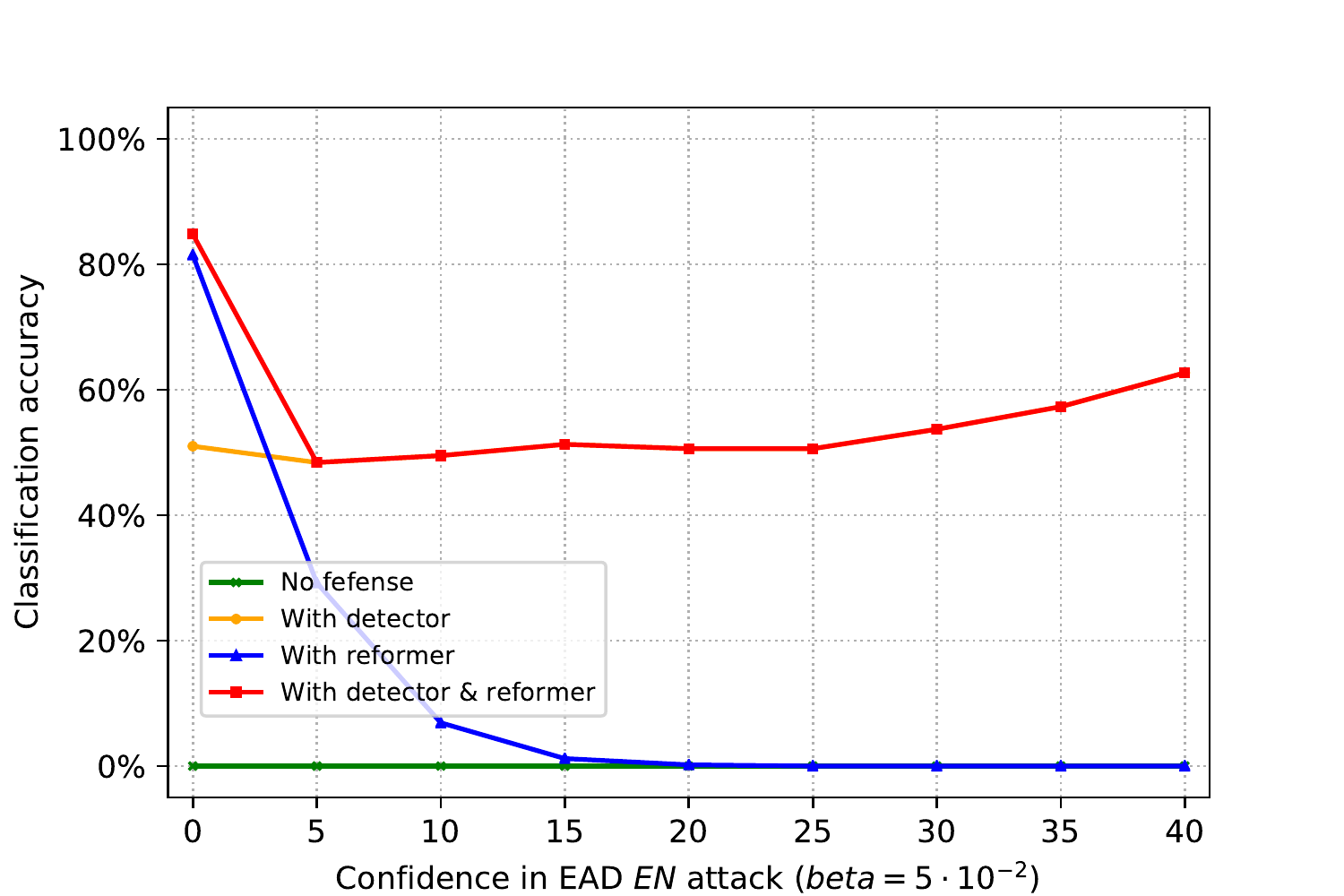}}
	\\	
	\subfloat[$L_1$ decision rule $\beta=10^{-1}$]{
		\label{fig:L1_e-1_JSD}
		\includegraphics[scale=0.5]{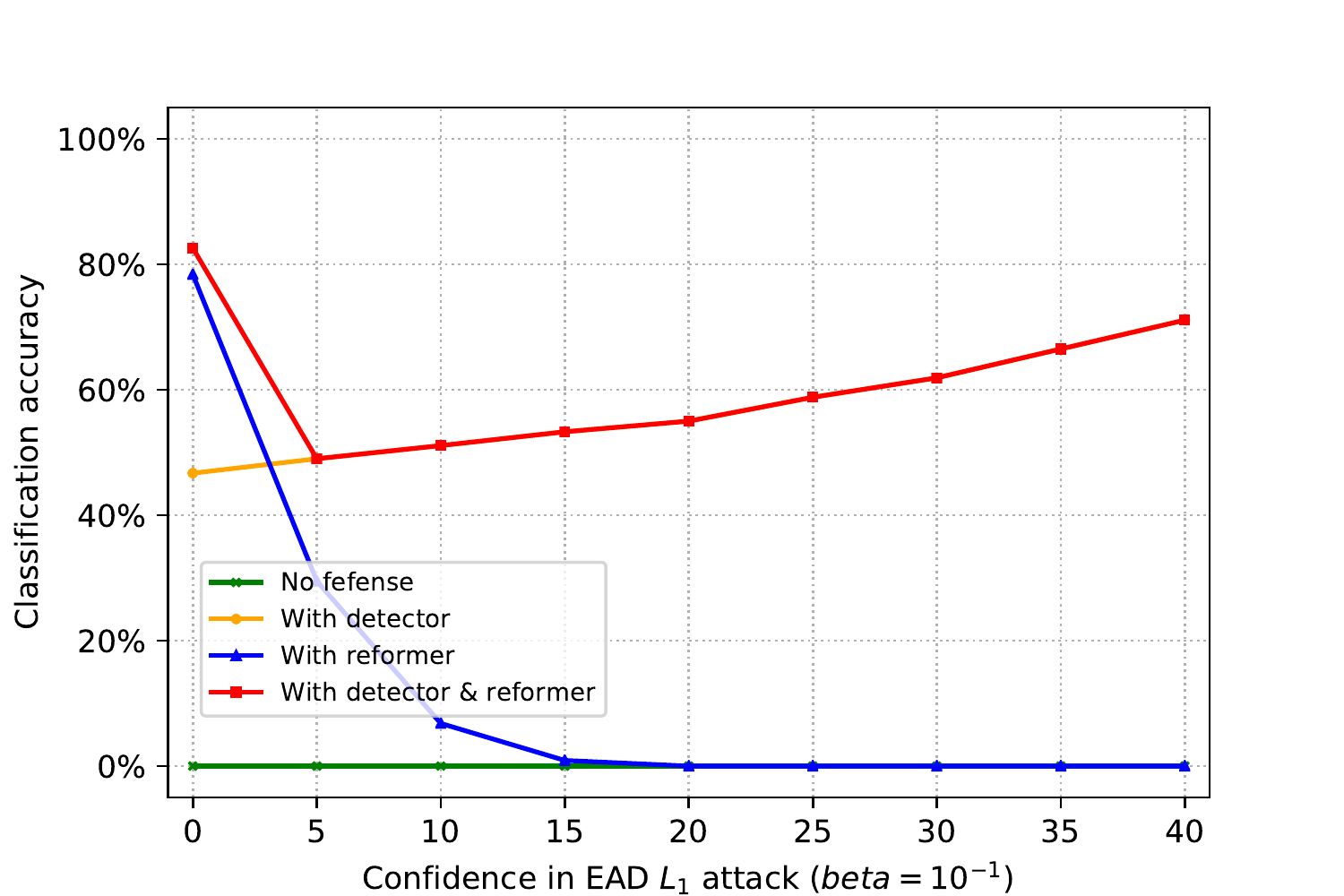}}
	\subfloat[EN decision rule $\beta=10^{-1}$]{
		\label{fig:EN_e-1_JSD}
		\includegraphics[scale=0.5]{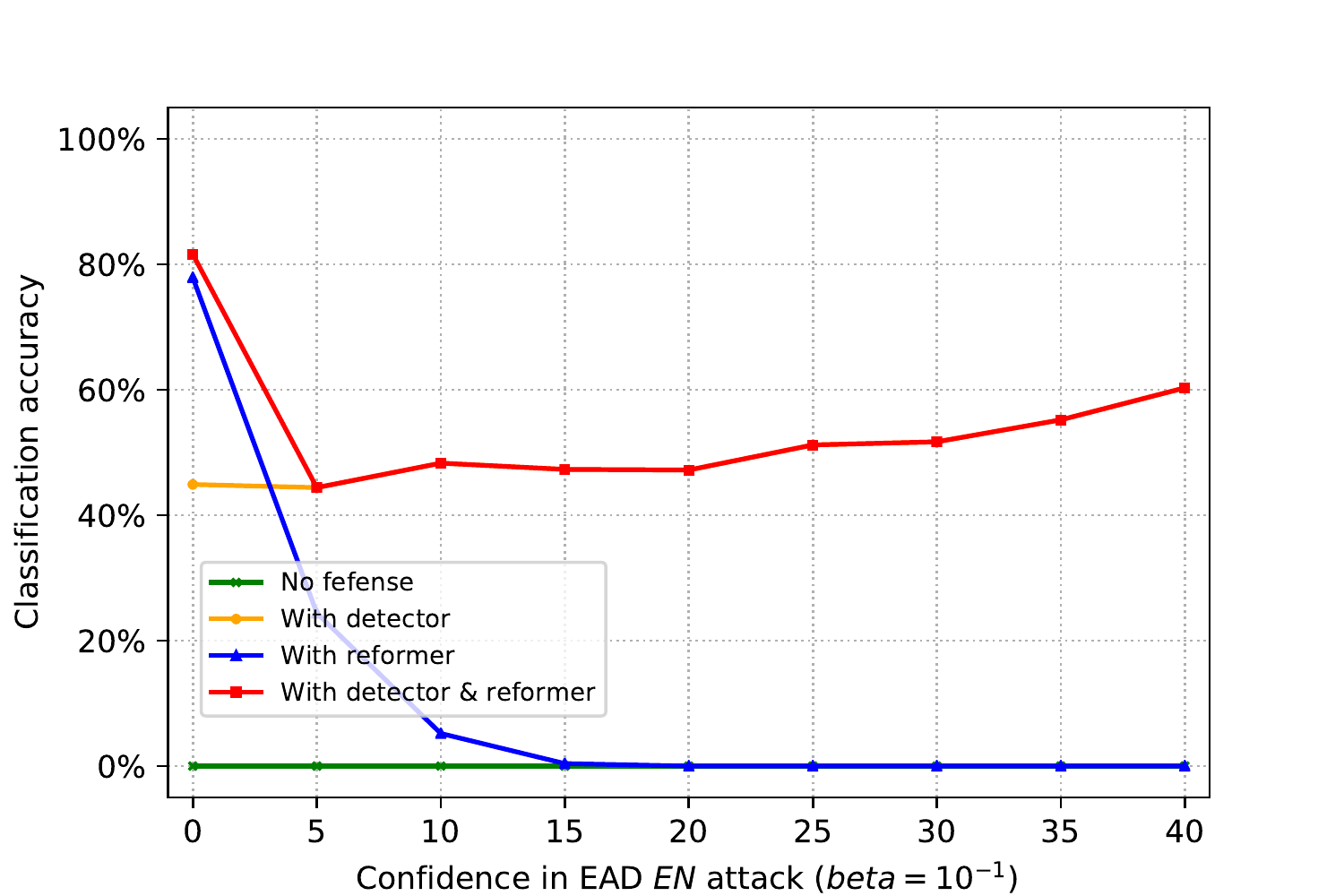}}
	\caption{EAD attacks on robust MagNet under different $\beta$ and different decision rules on MNIST  with varying confidence. Two JSD detectors are added into MagNet.}
	\label{fig:MNIST_default_JSD}
\end{figure*}

\begin{figure*}
	\centering
	\subfloat[$L_1$ decision rule $\beta=10^{-3}$]{
		\label{fig:L1_e-3_256}
		\includegraphics[scale=0.5]{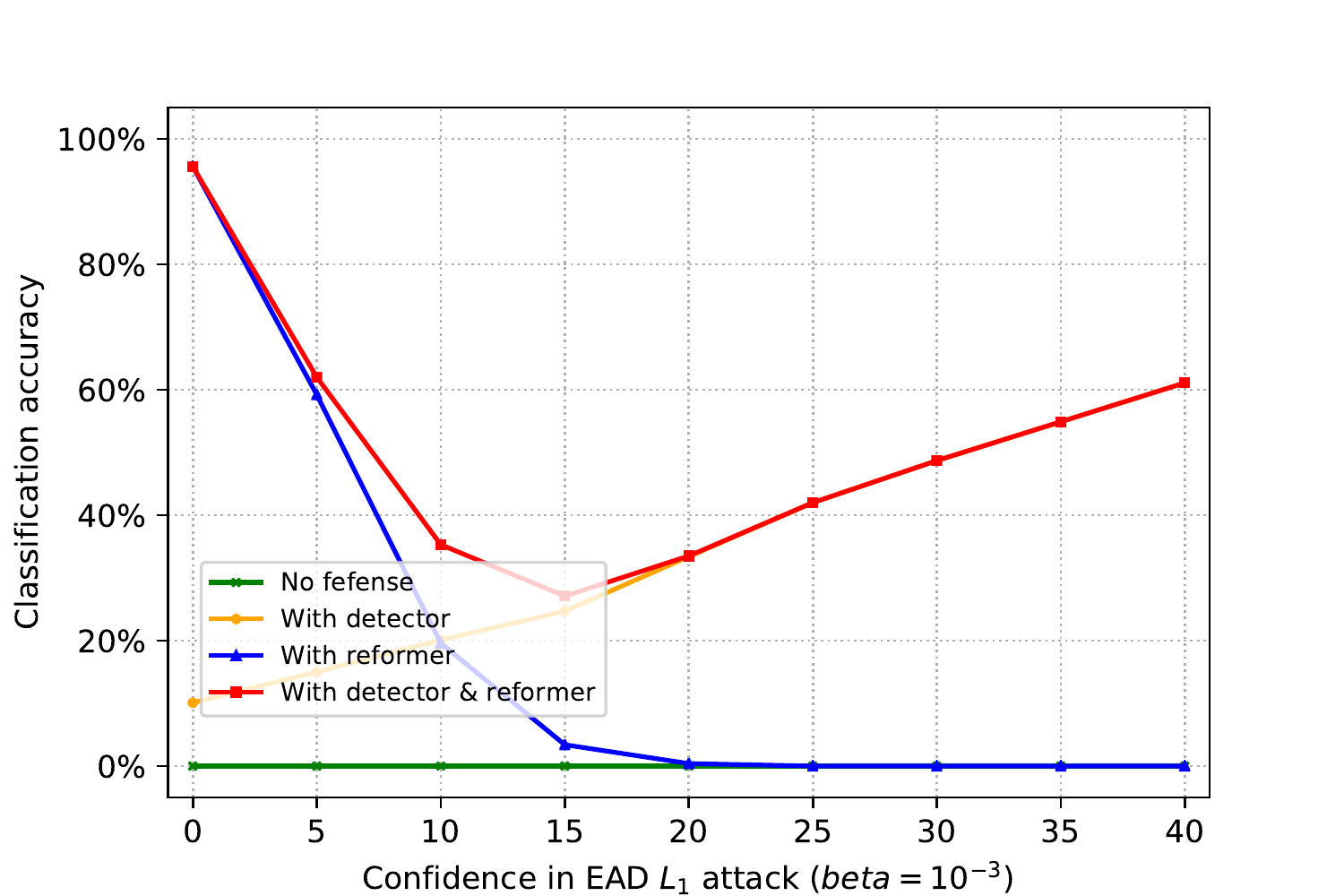}}
	\subfloat[EN decision rule $\beta=10^{-3}$]{
		\label{fig:EN_e-3_256}
		\includegraphics[scale=0.5]{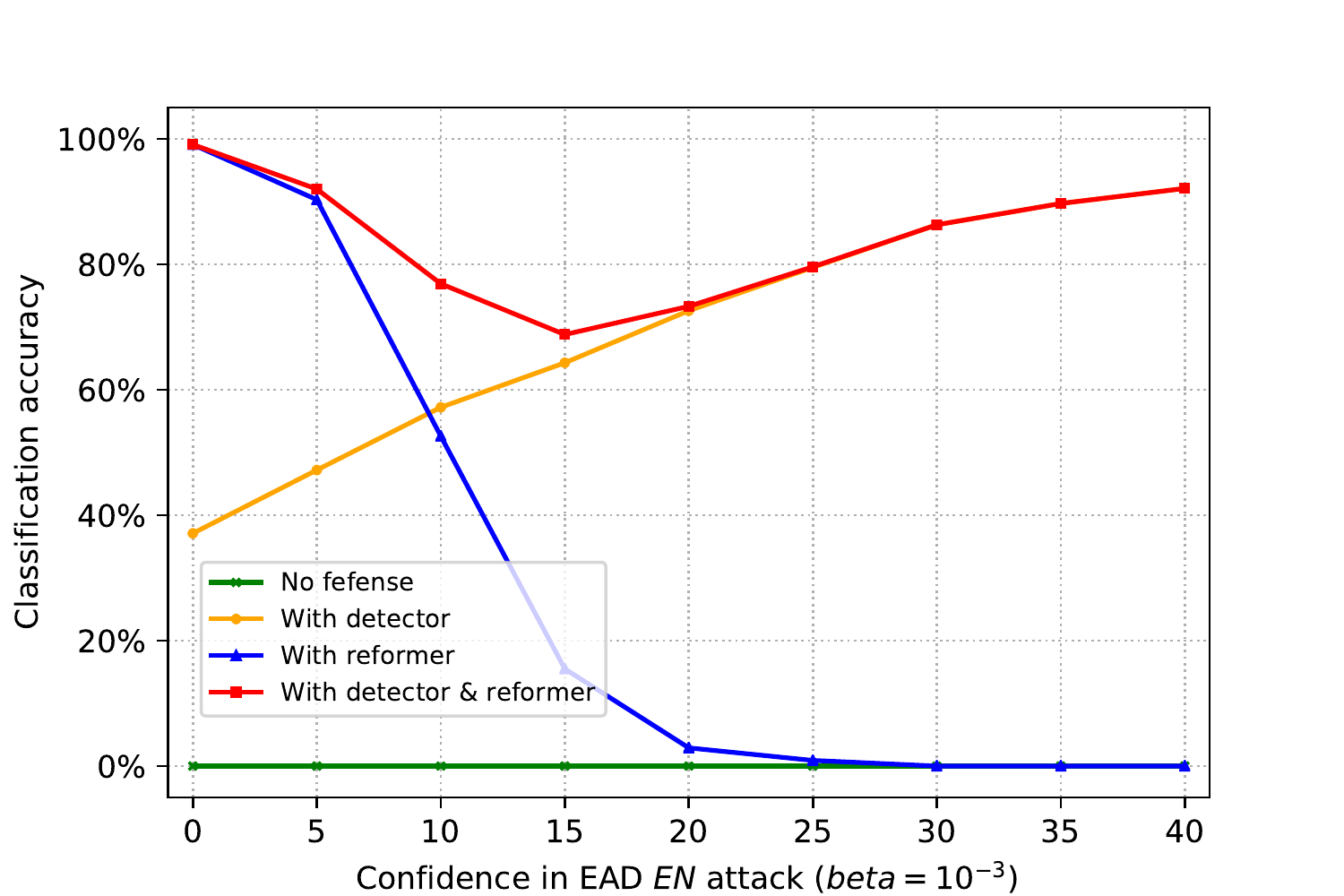}}
	\\
	\subfloat[$L_1$ decision rule $\beta=10^{-2}$]{
		\label{fig:L1_e-2_256}
		\includegraphics[scale=0.5]{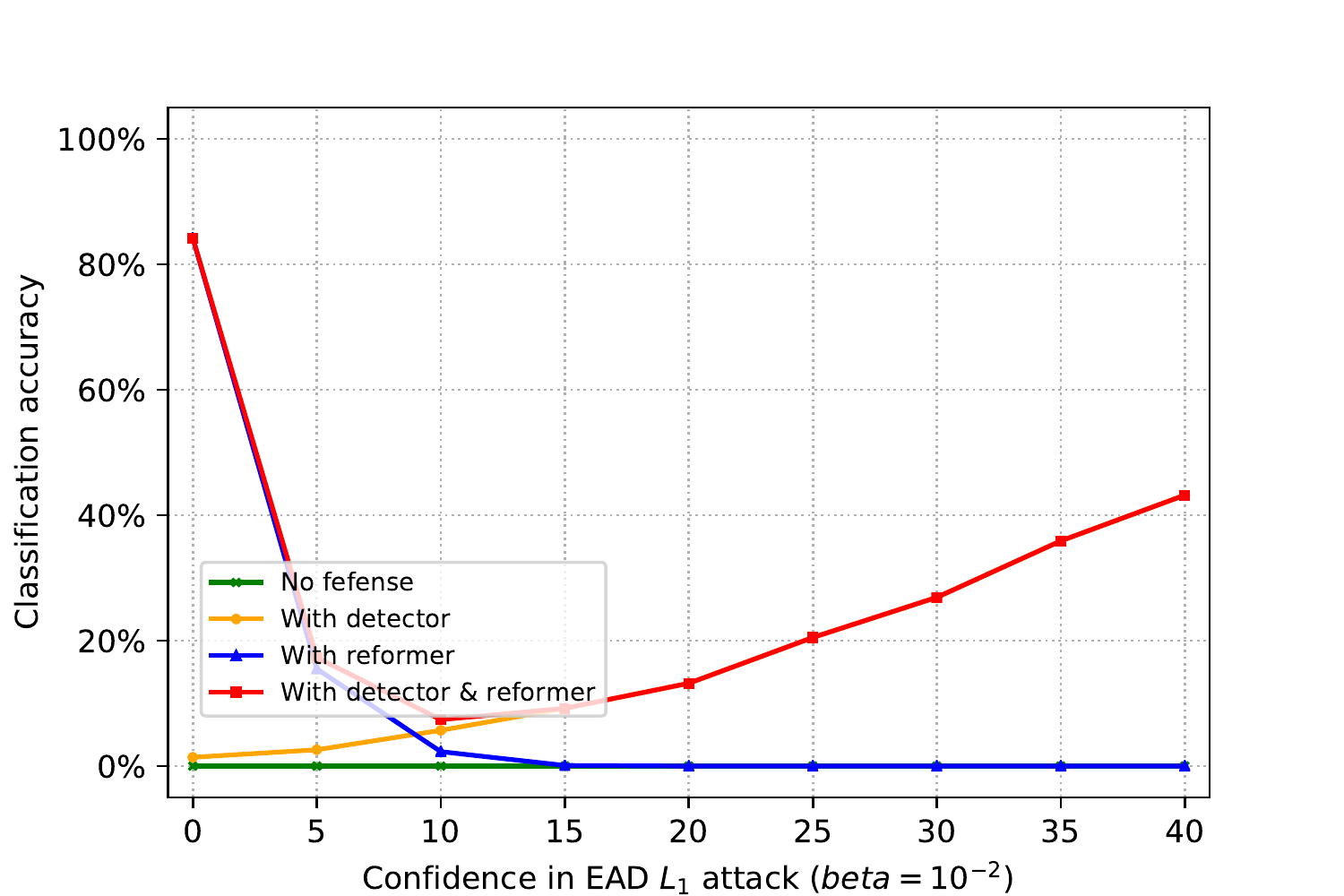}}
	\subfloat[EN decision rule $\beta=10^{-2}$]{
		\label{fig:EN_e-2_256}
		\includegraphics[scale=0.5]{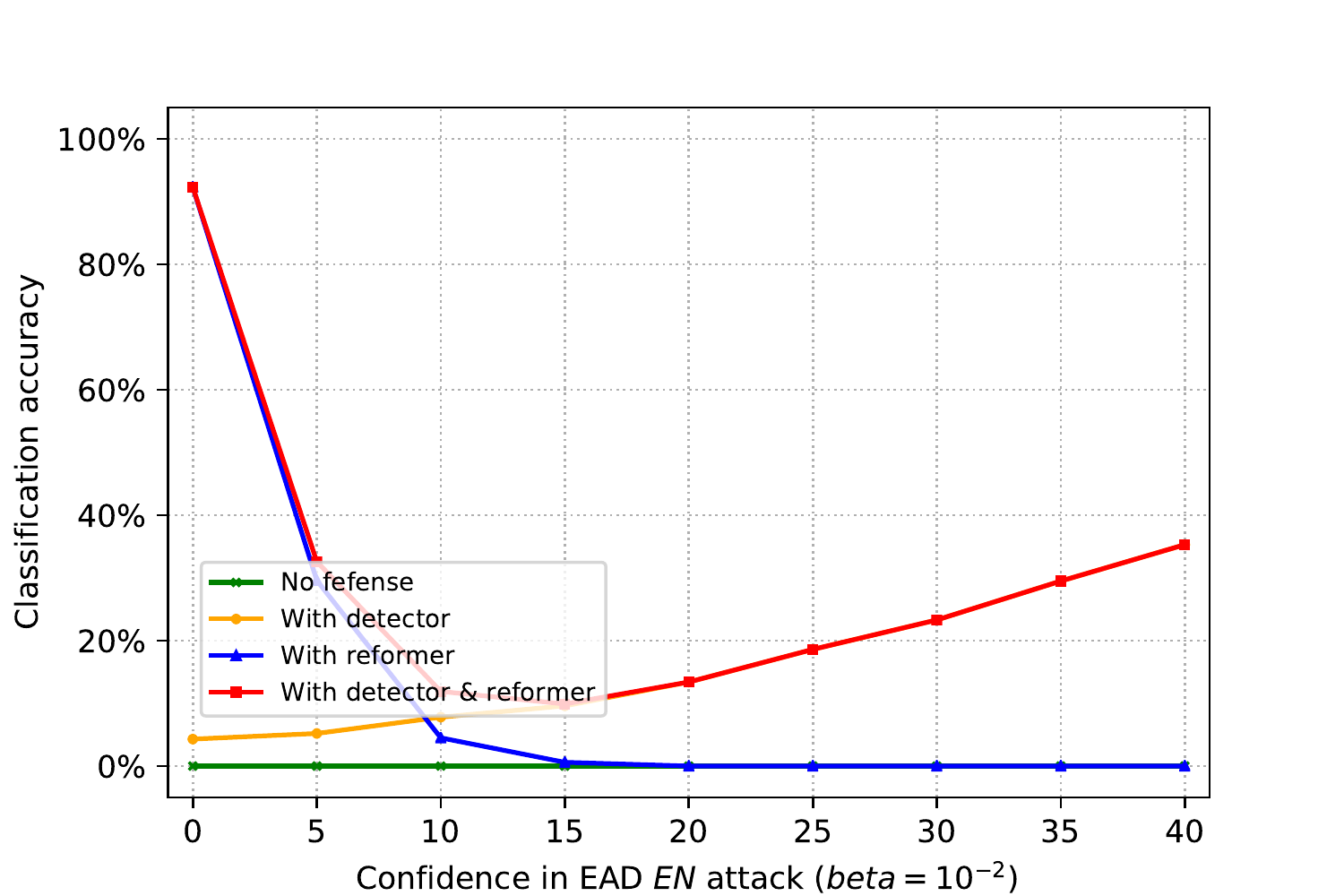}}
	\\
	\subfloat[$L_1$ decision rule $\beta=5\cdot10^{-2}$]{
		\label{fig:L1_5e-2_256}
		\includegraphics[scale=0.5]{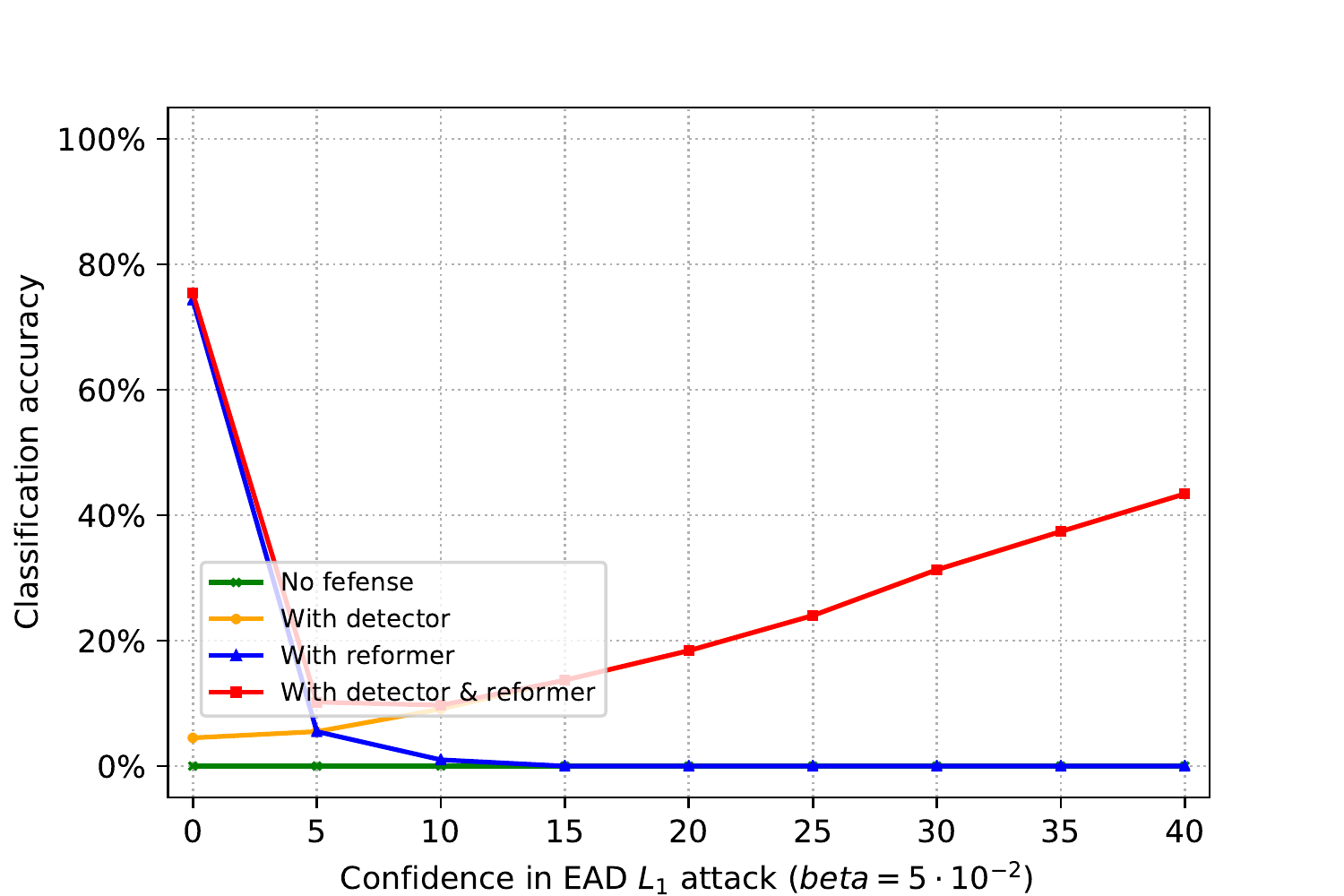}}
	\subfloat[EN decision rule $\beta=5\cdot10^{-2}$]{
		\label{fig:EN_5e-2_256}
		\includegraphics[scale=0.5]{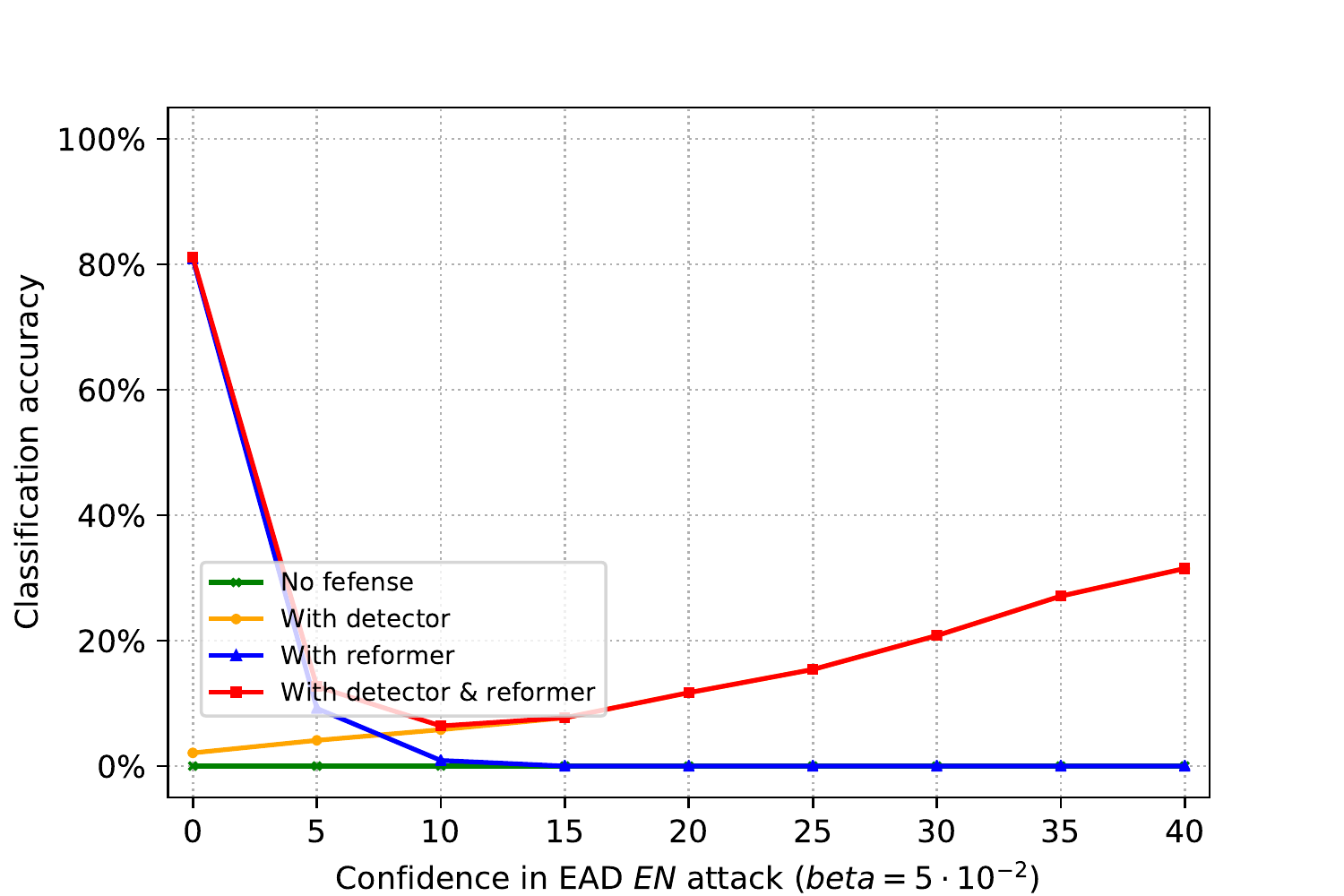}}
	\\	
	\subfloat[$L_1$ decision rule $\beta=10^{-1}$]{
		\label{fig:L1_e-1_256}
		\includegraphics[scale=0.5]{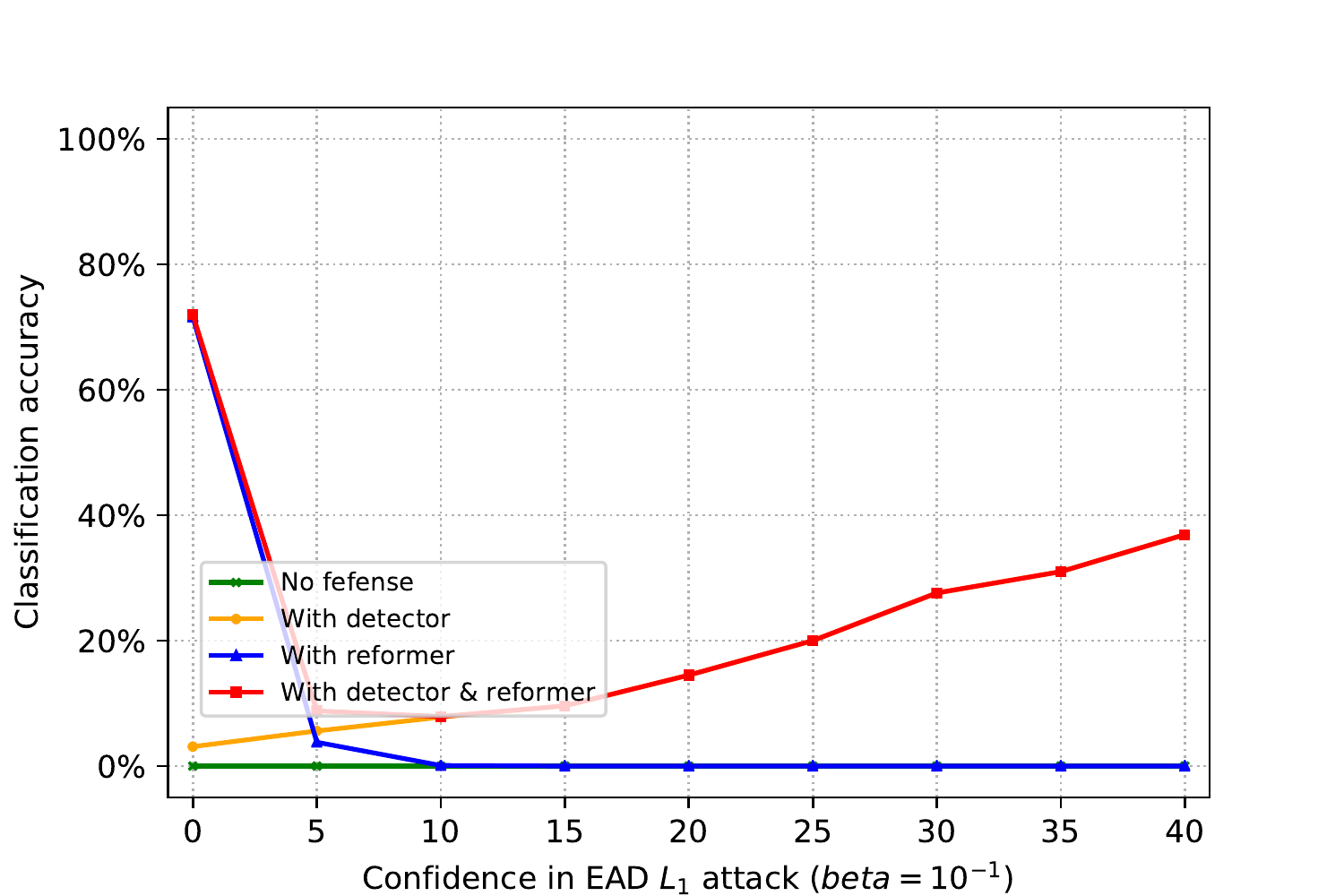}}
	\subfloat[EN decision rule $\beta=10^{-1}$]{
		\label{fig:EN_e-1_256}
		\includegraphics[scale=0.5]{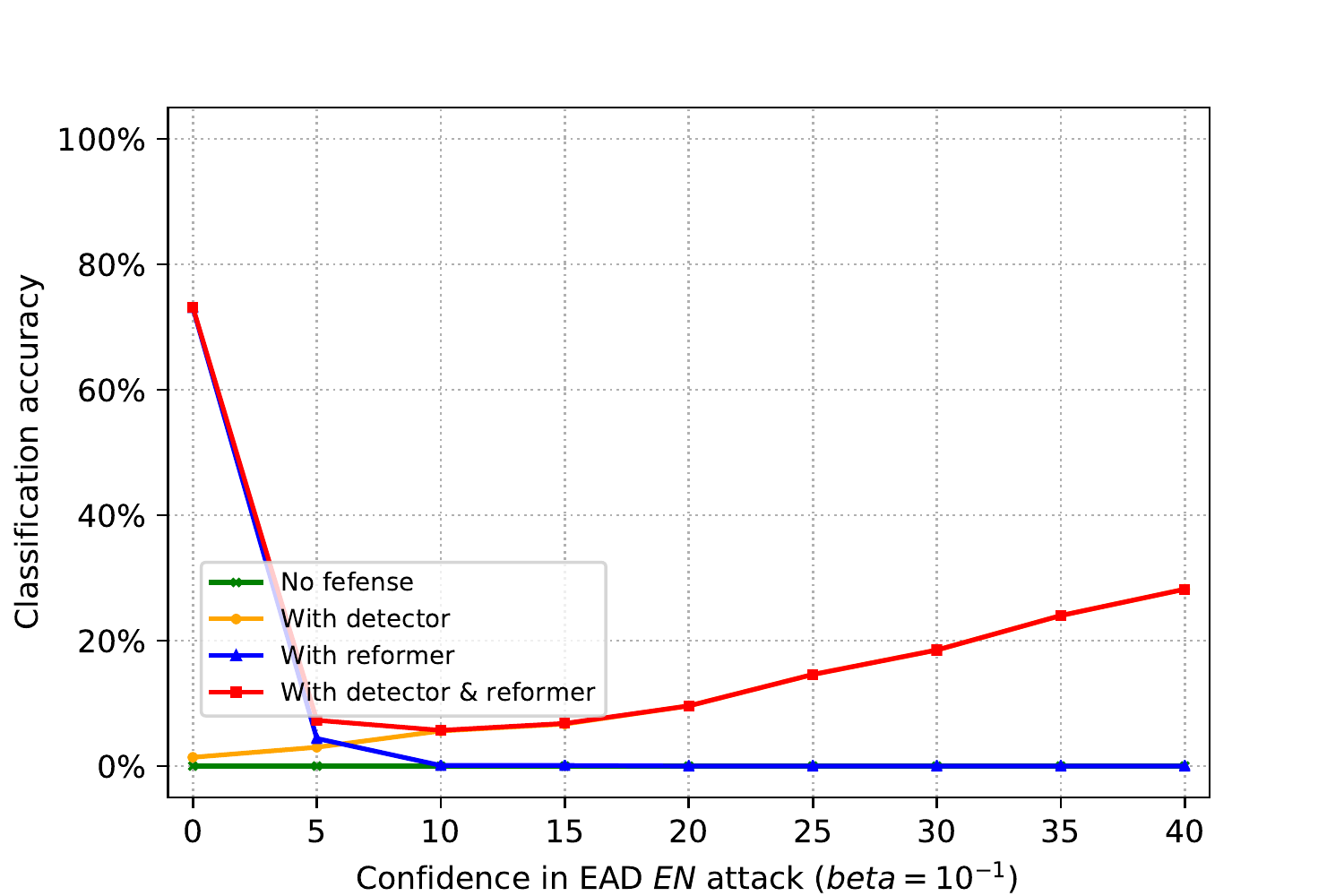}}
	\caption{EAD attacks on robust MagNet under different $\beta$ and different decision rules on MNIST dataset with varying confidence. The number of filters in a auto-encoder's convolution layer is increased to 256.}
	\label{fig:MNIST_256}
\end{figure*}

\begin{figure*}
	\centering
	\subfloat[$L_1$ decision rule $\beta=10^{-3}$]{
		\label{fig:L1_e-3_256_JSD}
		\includegraphics[scale=0.5]{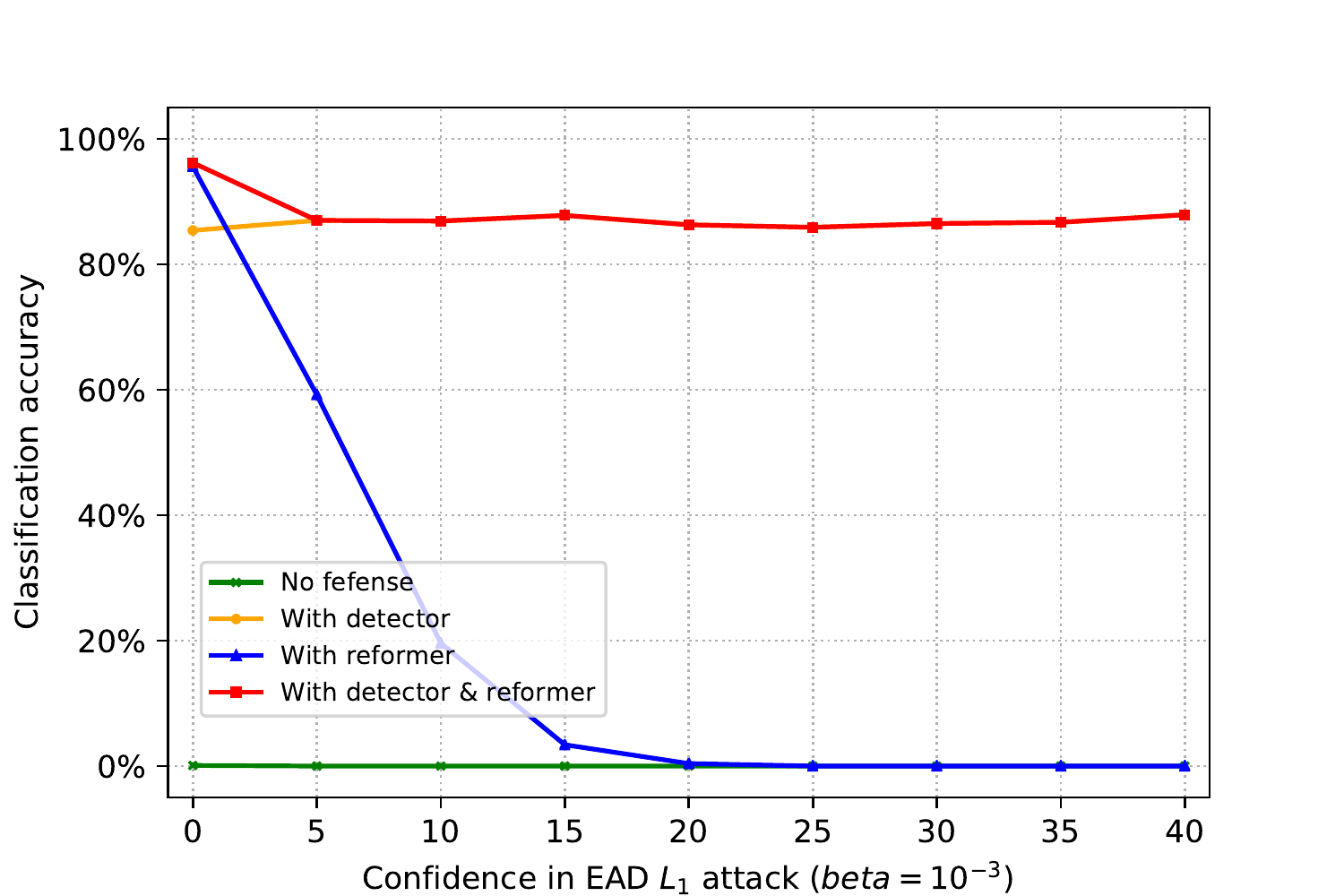}}
	\subfloat[EN decision rule $\beta=10^{-3}$]{
		\label{fig:EN_e-3_256_JSD}
		\includegraphics[scale=0.5]{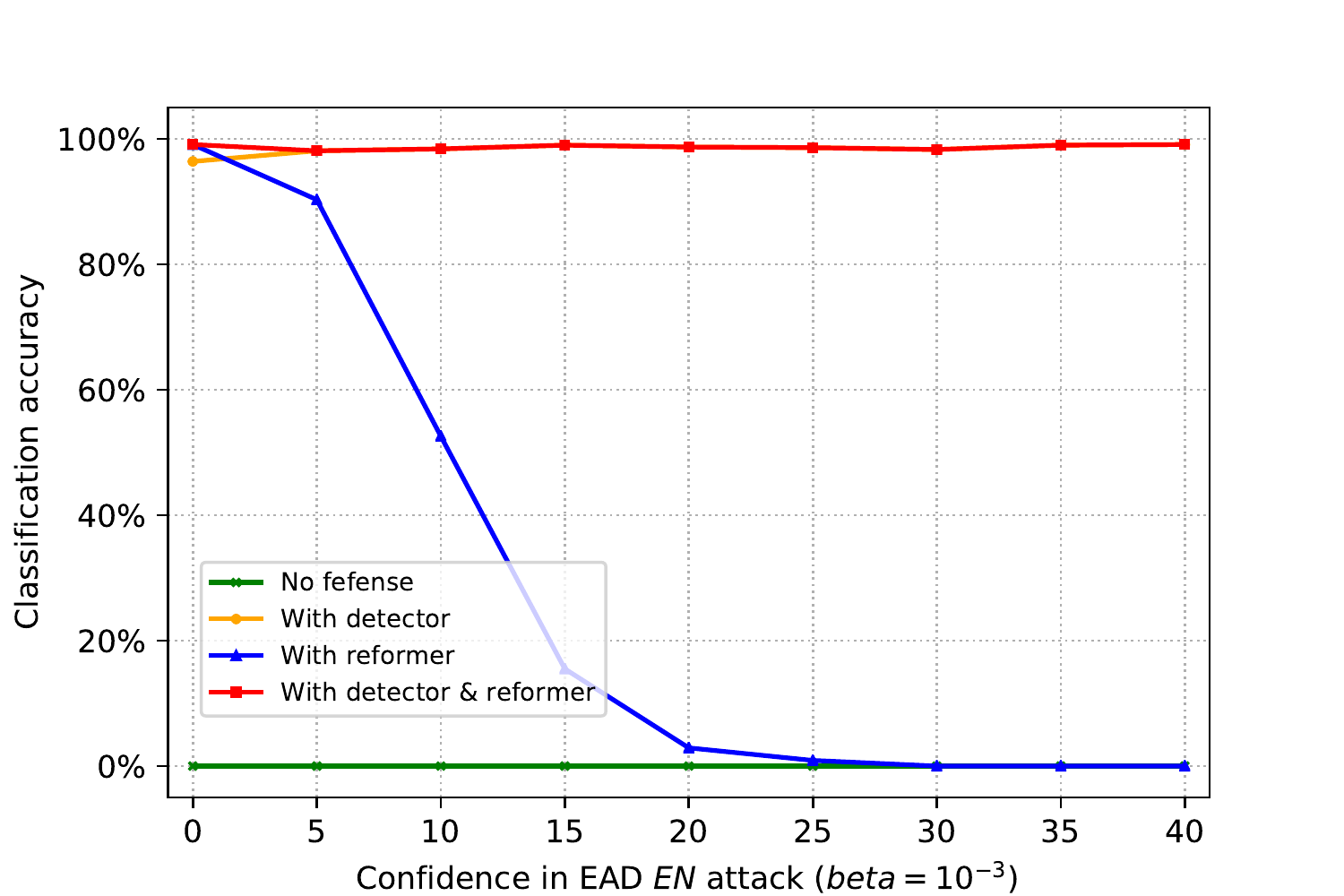}}
	\\
	\subfloat[$L_1$ decision rule $\beta=10^{-2}$]{
		\label{fig:L1_e-2_256_JSD}
		\includegraphics[scale=0.5]{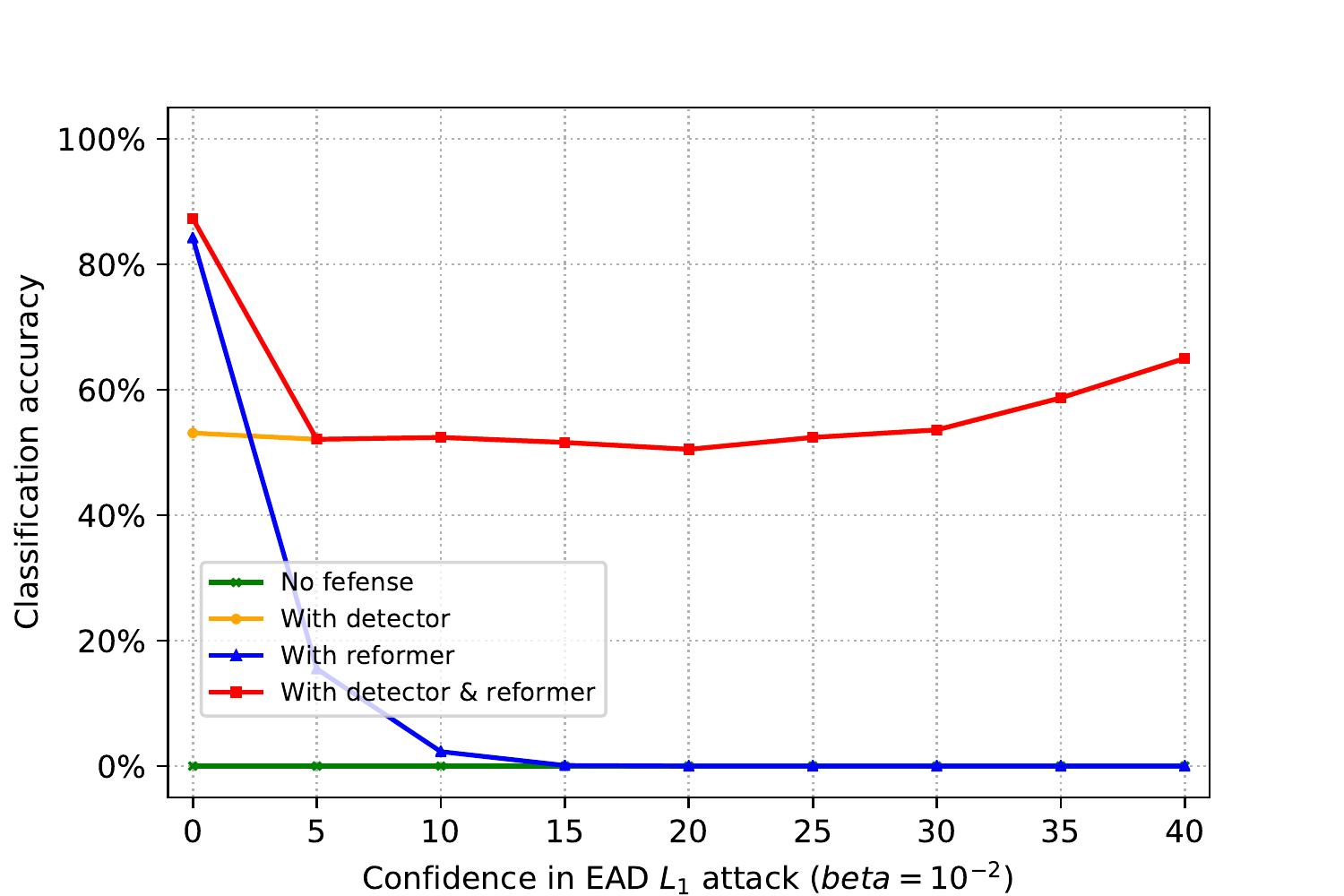}}
	\subfloat[EN decision rule $\beta=10^{-2}$]{
		\label{fig:EN_e-2_256_JSD}
		\includegraphics[scale=0.5]{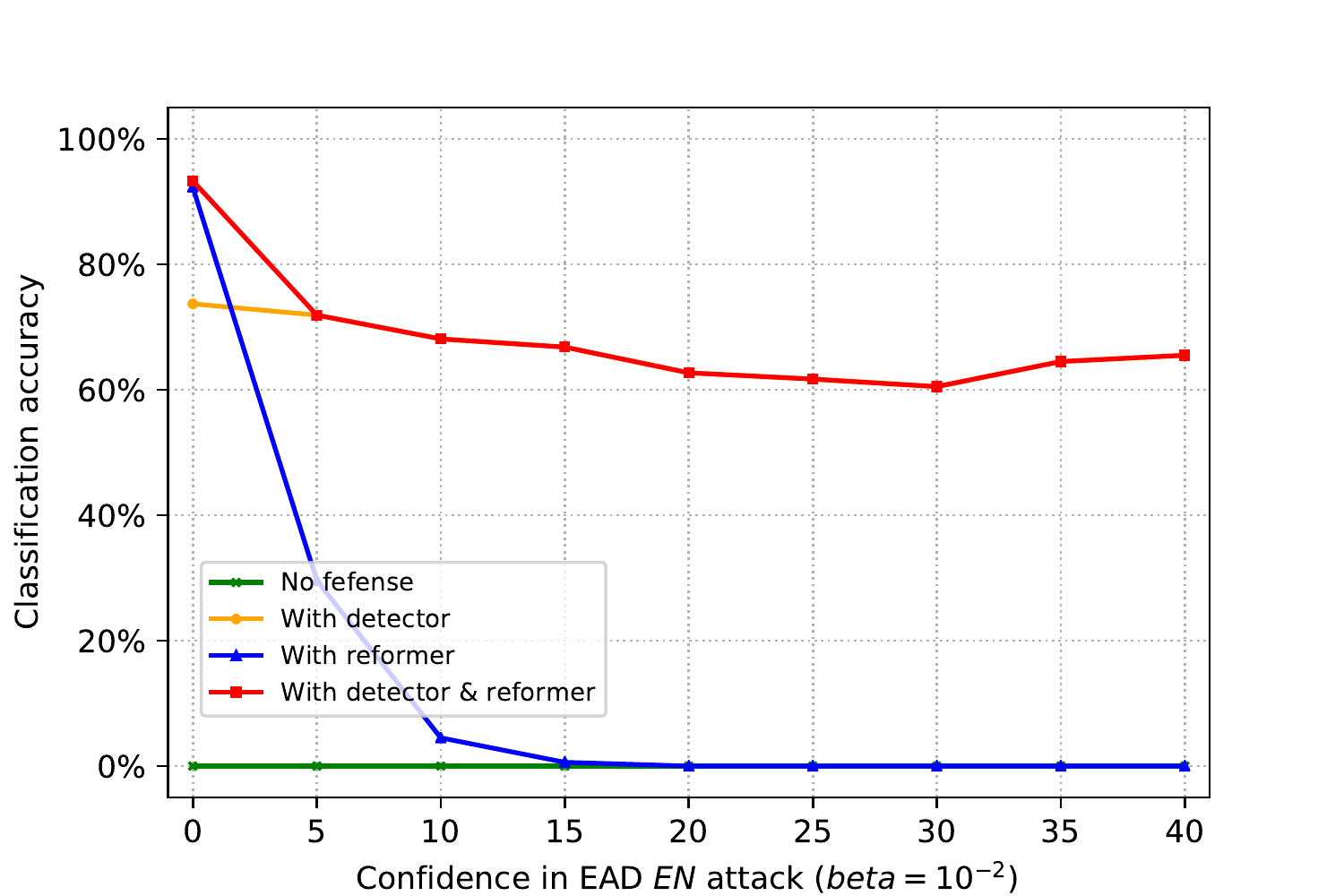}}
	\\
	\subfloat[$L_1$ decision rule $\beta=5\cdot10^{-2}$]{
		\label{fig:L1_5e-2_256_JSD}
		\includegraphics[scale=0.5]{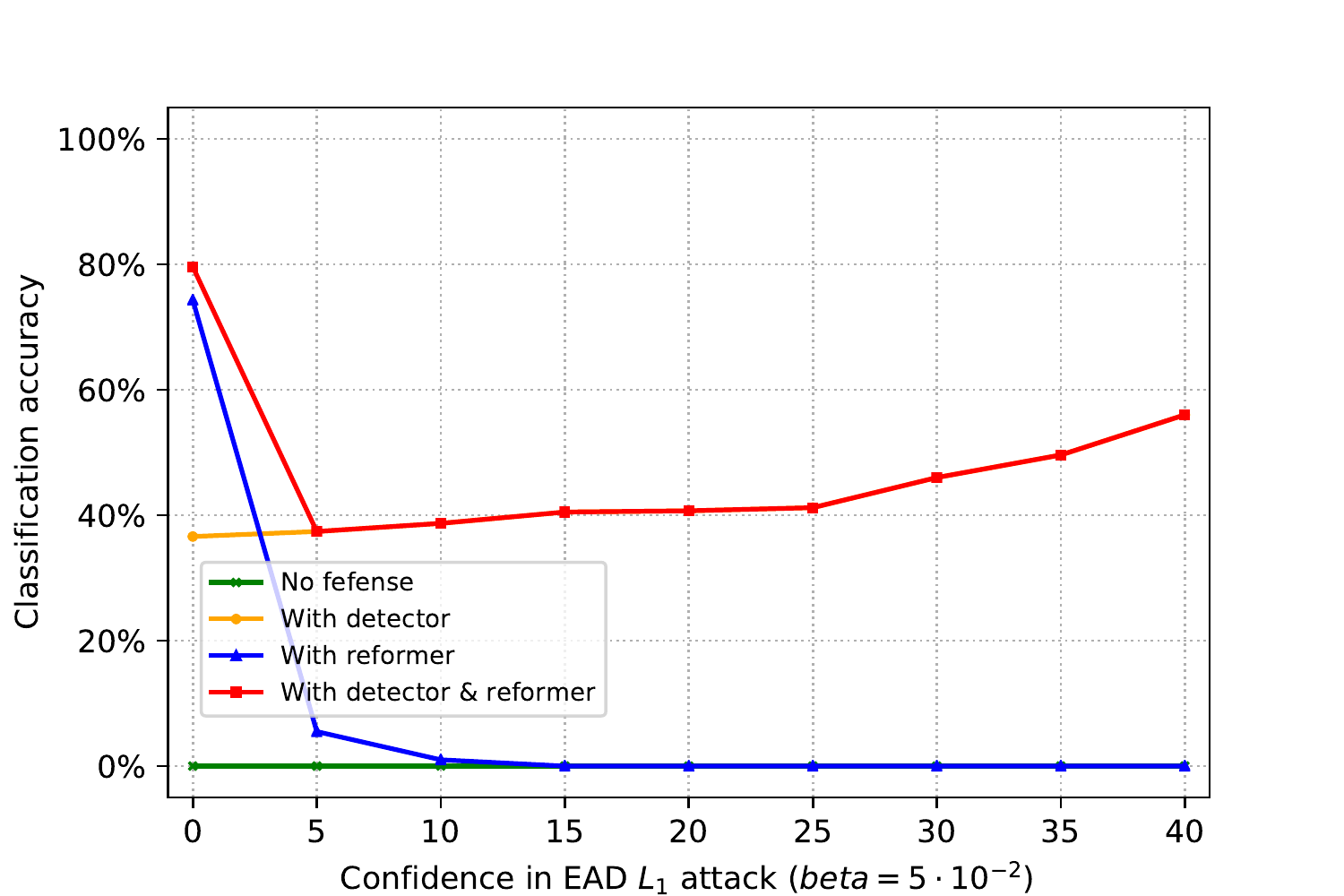}}
	\subfloat[EN decision rule $\beta=5\cdot10^{-2}$]{
		\label{fig:EN_5e-2_256_JSD}
		\includegraphics[scale=0.5]{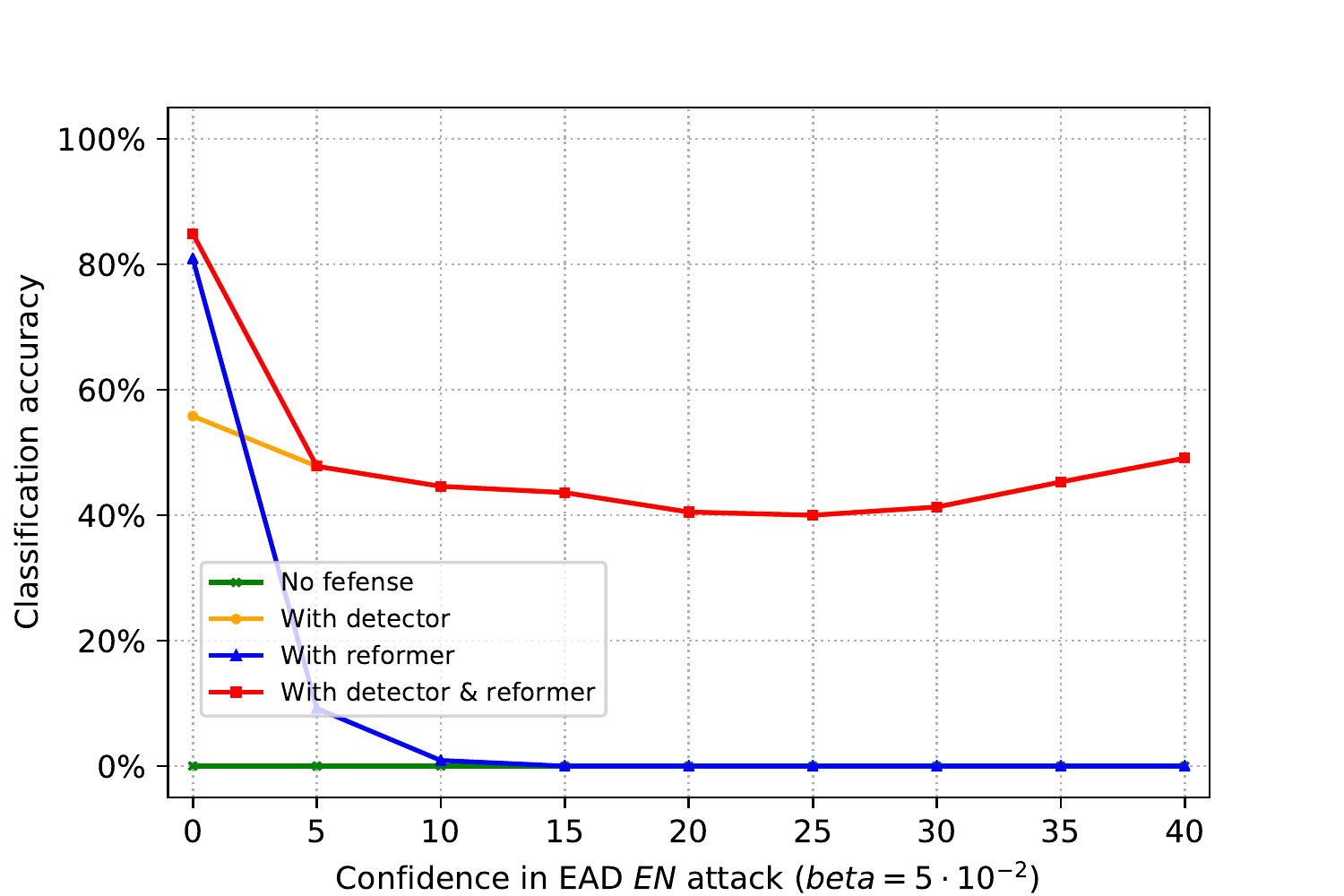}}
	\\	
	\subfloat[$L_1$ decision rule $\beta=10^{-1}$]{
		\label{fig:L1_e-1_256_JSD}
		\includegraphics[scale=0.5]{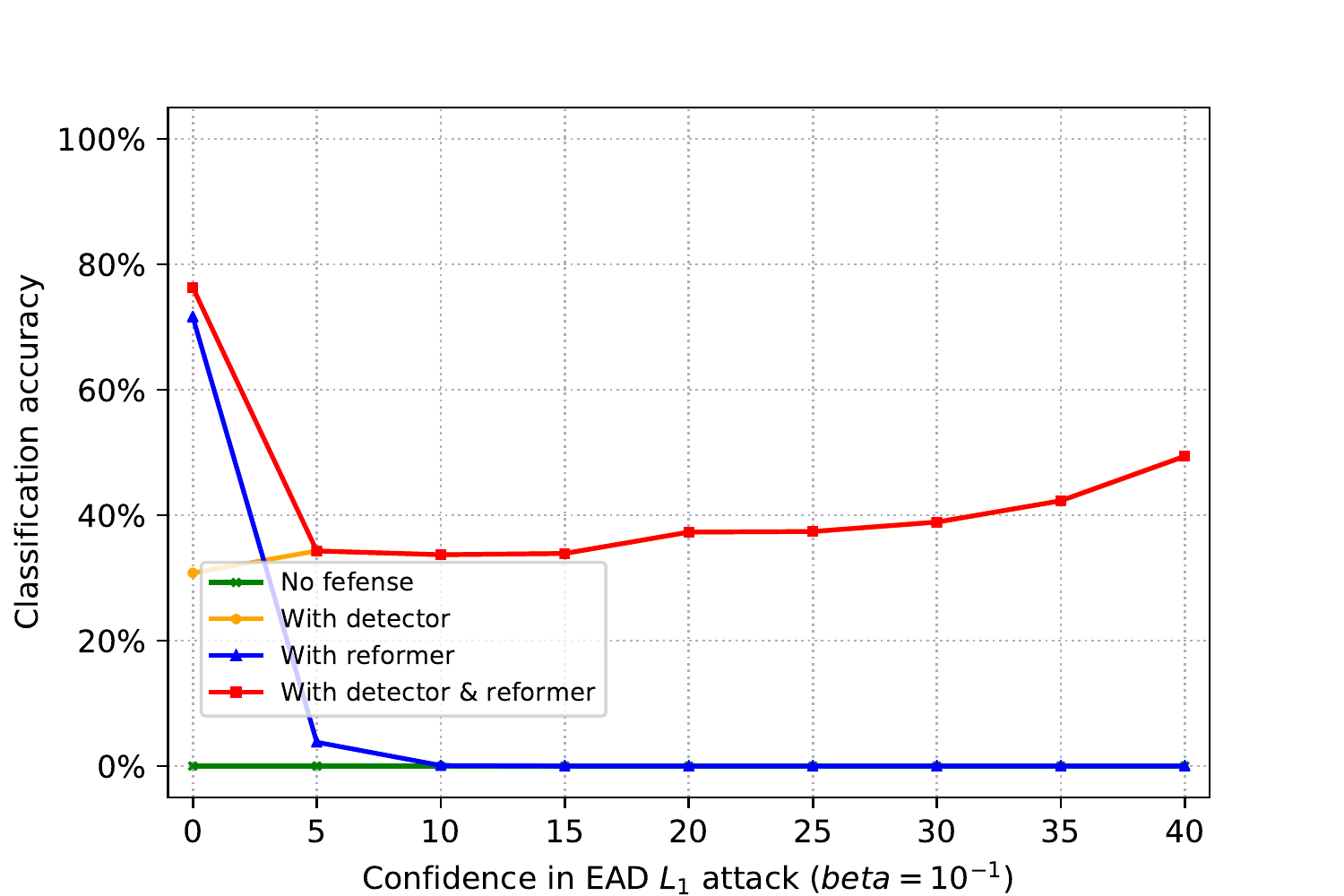}}
	\subfloat[EN decision rule $\beta=10^{-1}$]{
		\label{fig:EN_e-1_256_JSD}
		\includegraphics[scale=0.5]{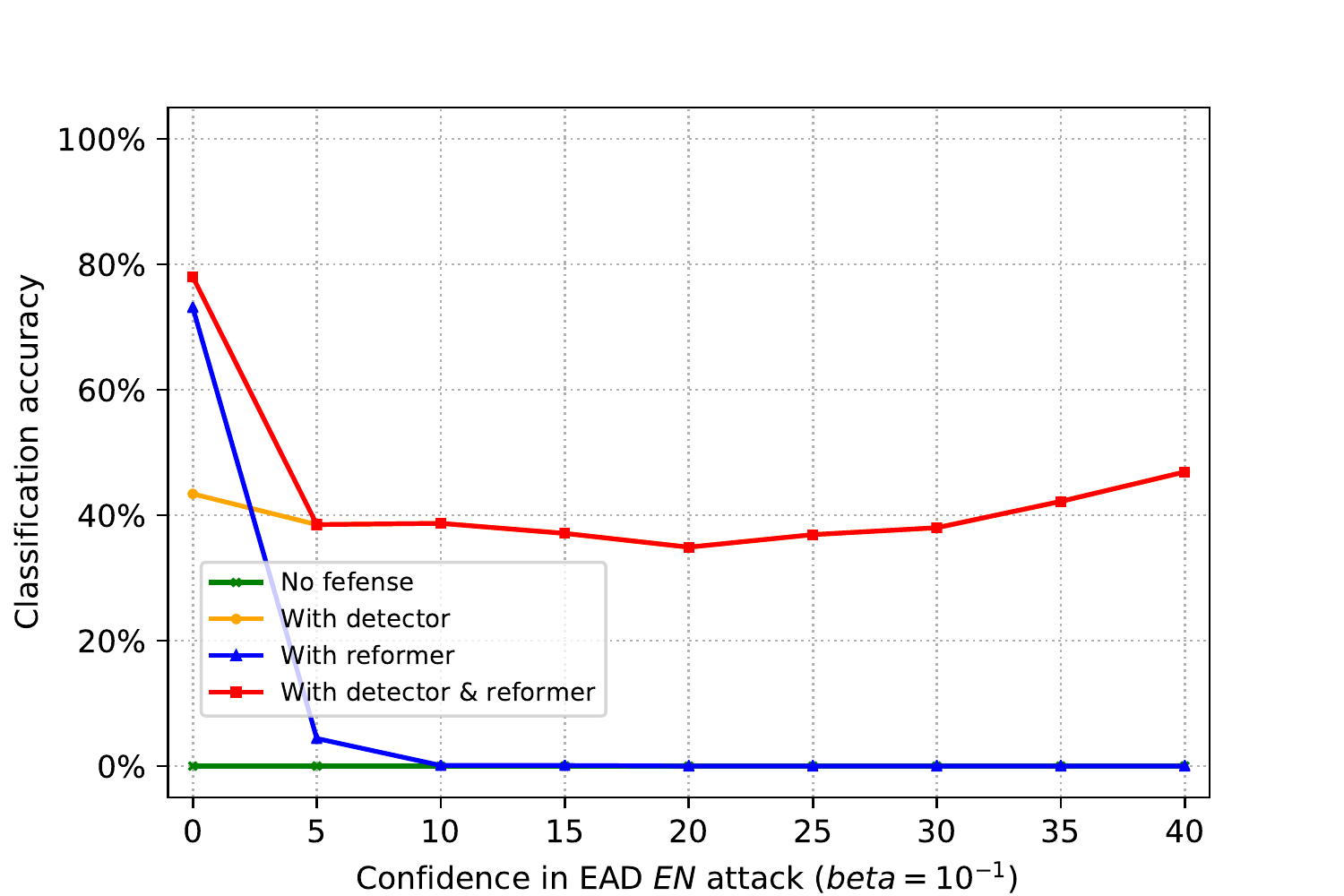}}
	\caption{EAD attacks on robust MagNet under different $\beta$ and different decision rules on MNIST with varying confidence. The number of filters in a auto-encoder's convolution layer is increased to 256 and two JSD detectors are added into MagNet.}
	\label{fig:MNIST_256_JSD}
\end{figure*}

\begin{figure*}
	\centering
	\subfloat[$L_1$ decision rule $\beta=10^{-3}$]{
		\label{fig:CIFAR_L1_e-3_256}
		\includegraphics[scale=0.5]{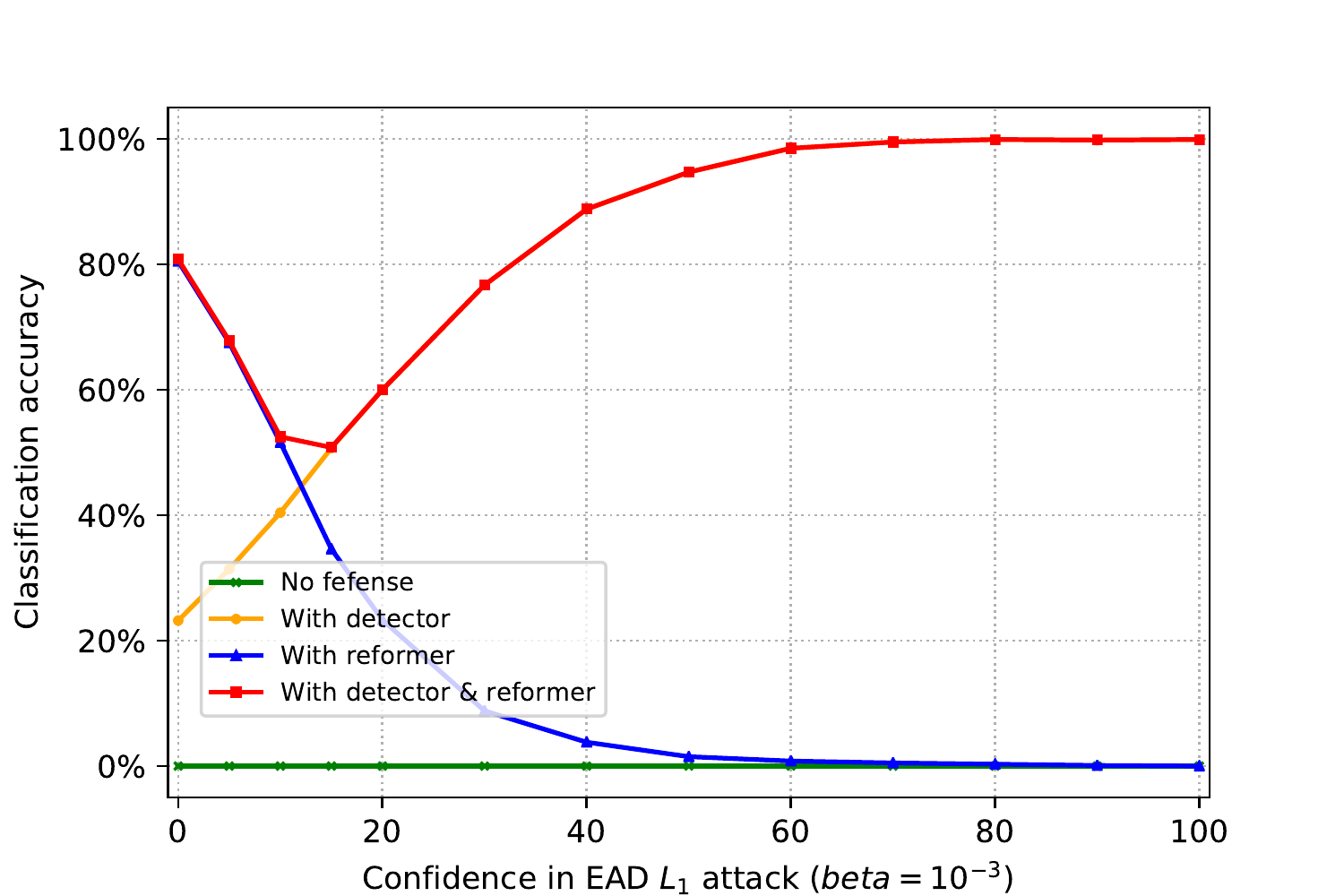}}
	\subfloat[EN decision rule $\beta=10^{-3}$]{
		\label{fig:CIFAR_EN_e-3_256}
		\includegraphics[scale=0.5]{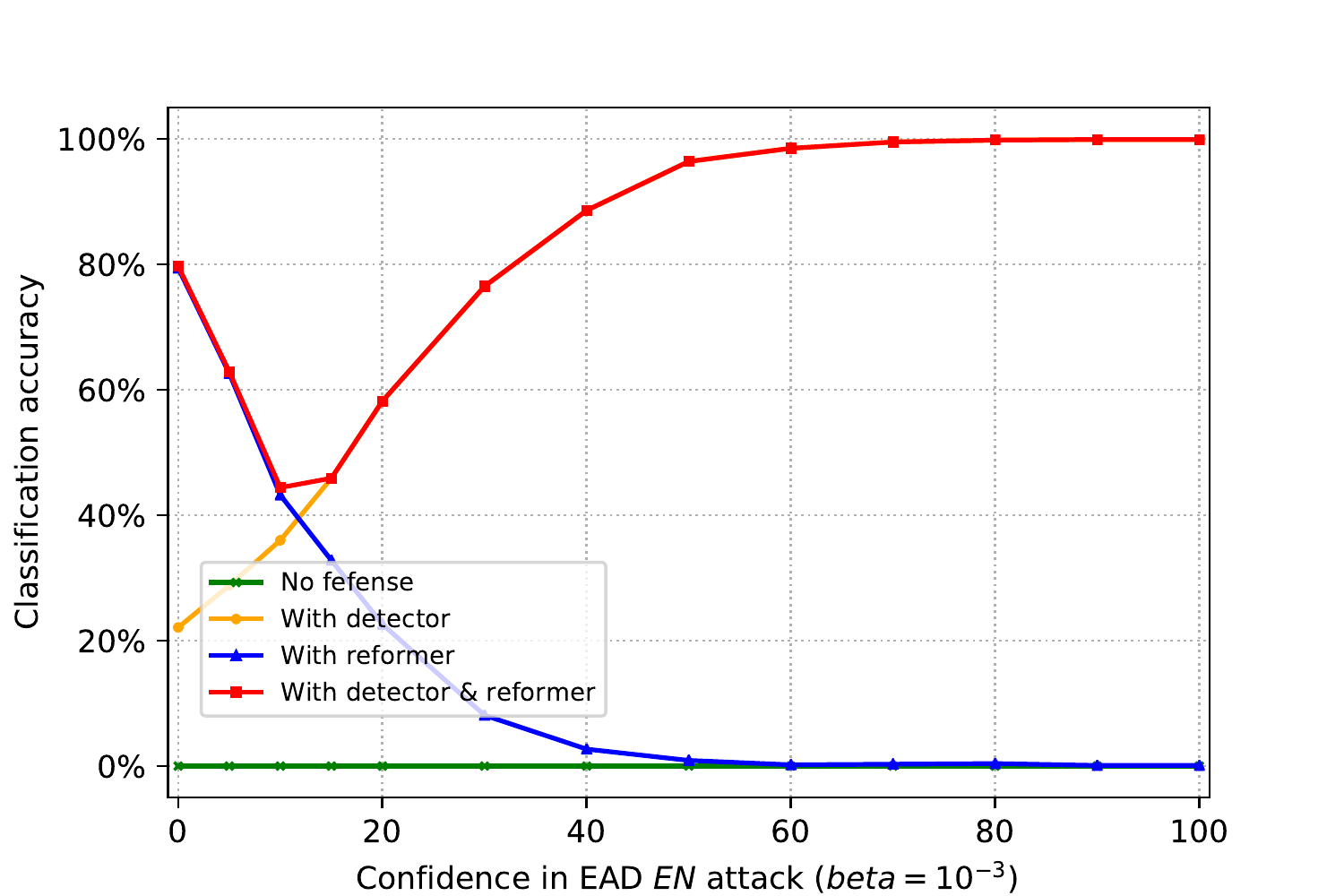}}
	\\
	\subfloat[$L_1$ decision rule $\beta=10^{-2}$]{
		\label{fig:CIFAR_L1_e-2_256}
		\includegraphics[scale=0.5]{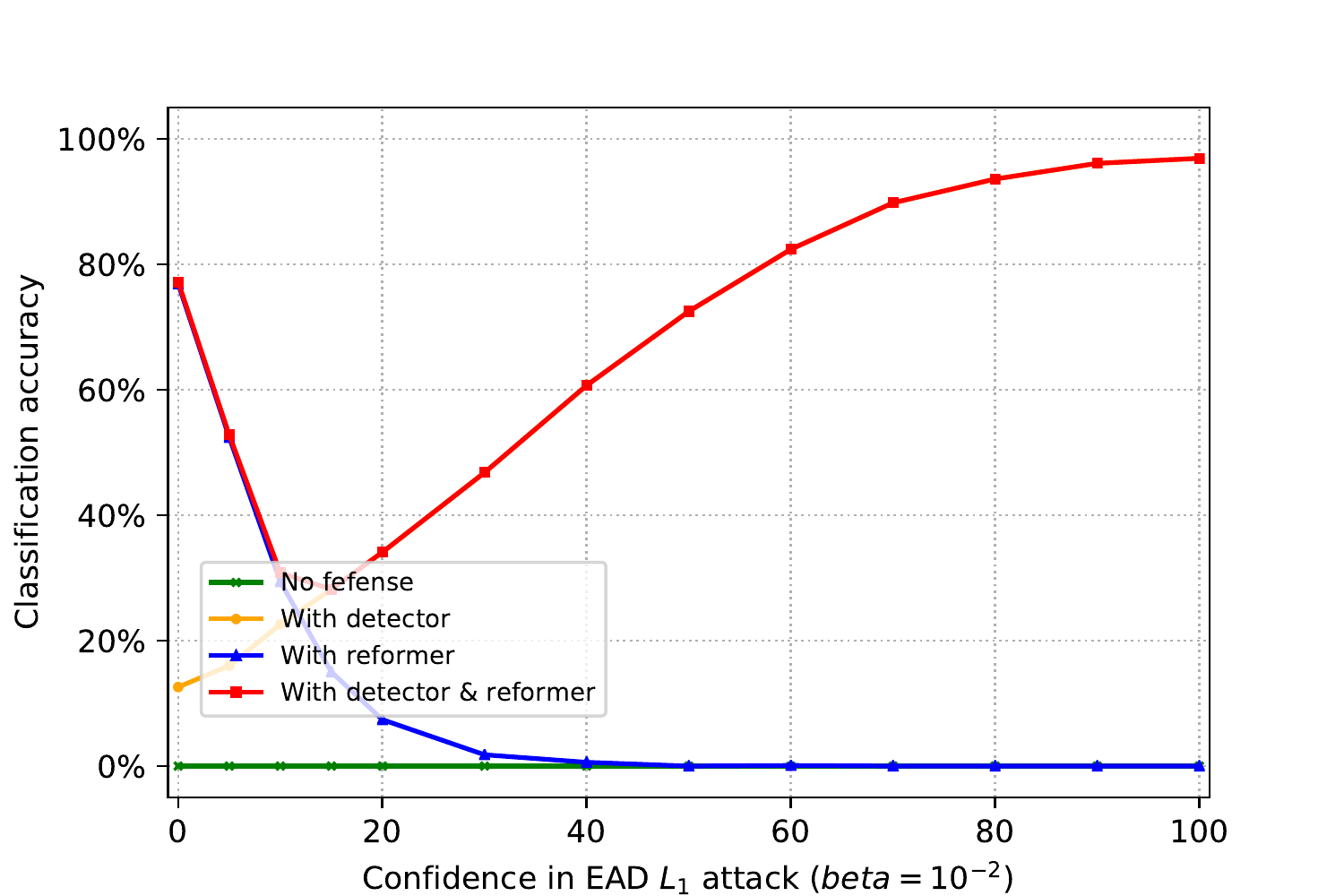}}
	\subfloat[EN decision rule $\beta=10^{-2}$]{
		\label{fig:CIFAR_EN_e-2_256}
		\includegraphics[scale=0.5]{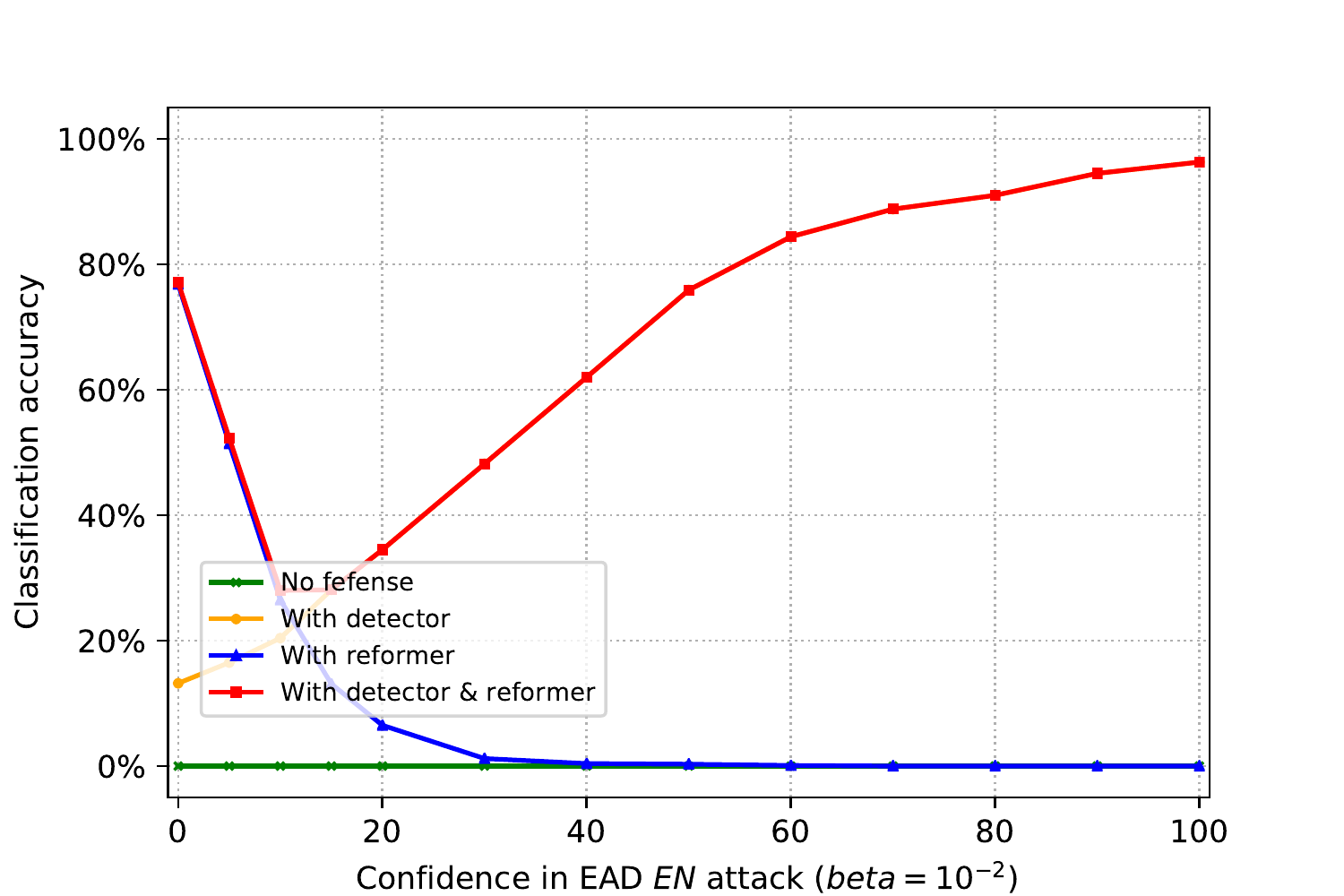}}
	\\
	\subfloat[$L_1$ decision rule $\beta=5\cdot10^{-2}$]{
		\label{fig:CIFAR_L1_5e-2_256}
		\includegraphics[scale=0.5]{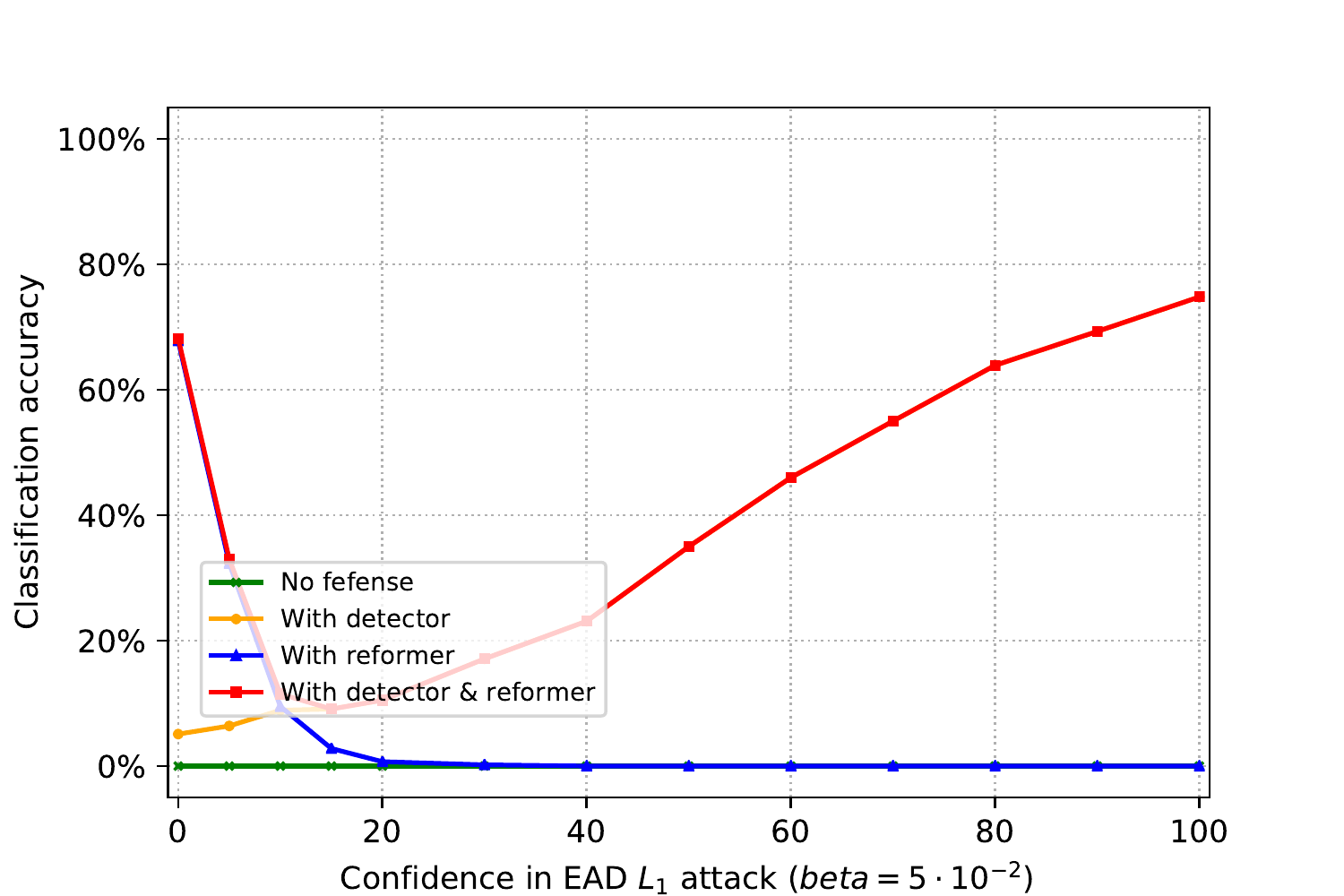}}
	\subfloat[EN decision rule $\beta=5\cdot10^{-2}$]{
		\label{fig:CIFAR_EN_5e-2_256}
		\includegraphics[scale=0.5]{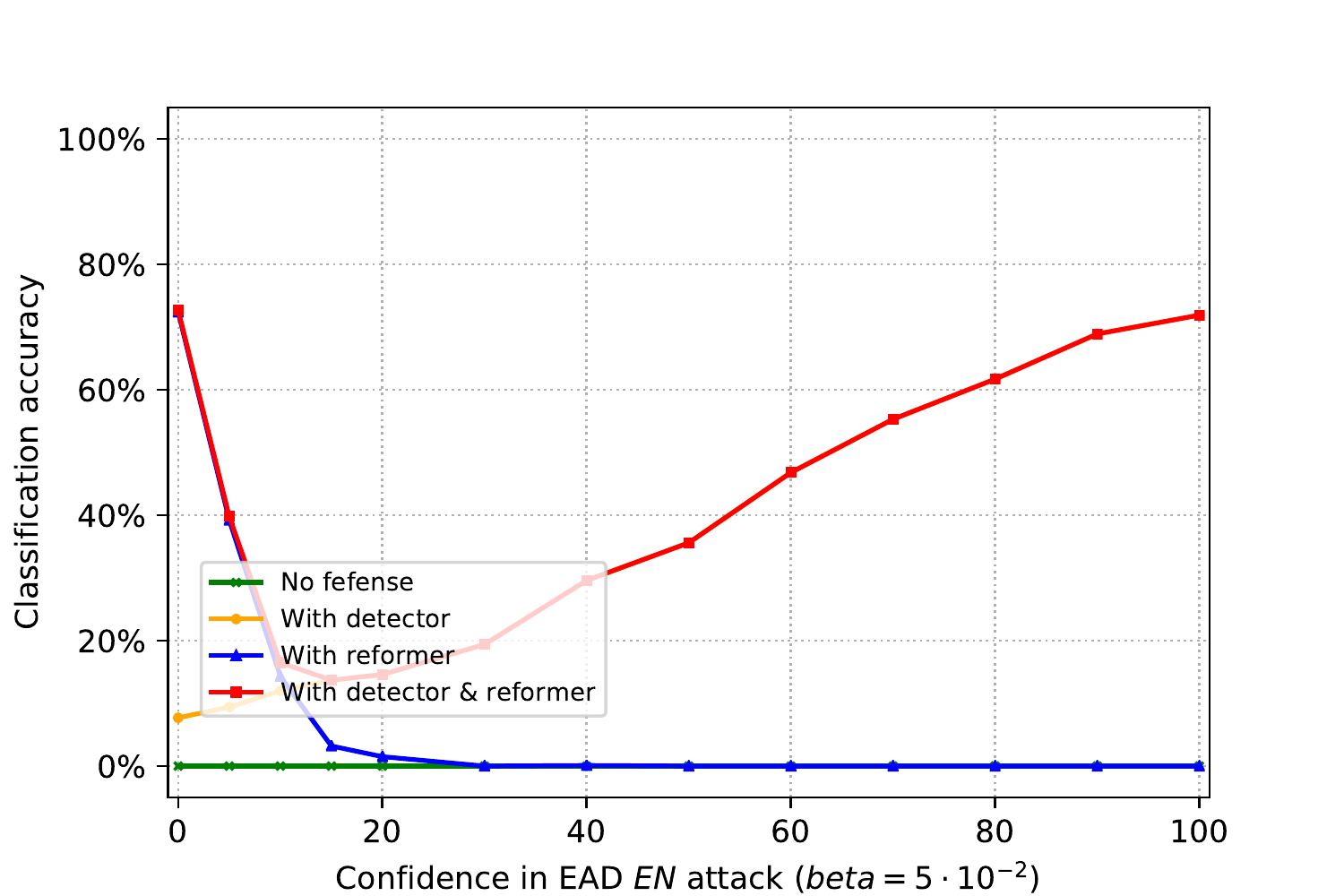}}
	\\	
	\subfloat[$L_1$ decision rule $\beta=10^{-1}$]{
		\label{fig:CIFAR_L1_e-1_256}
		\includegraphics[scale=0.5]{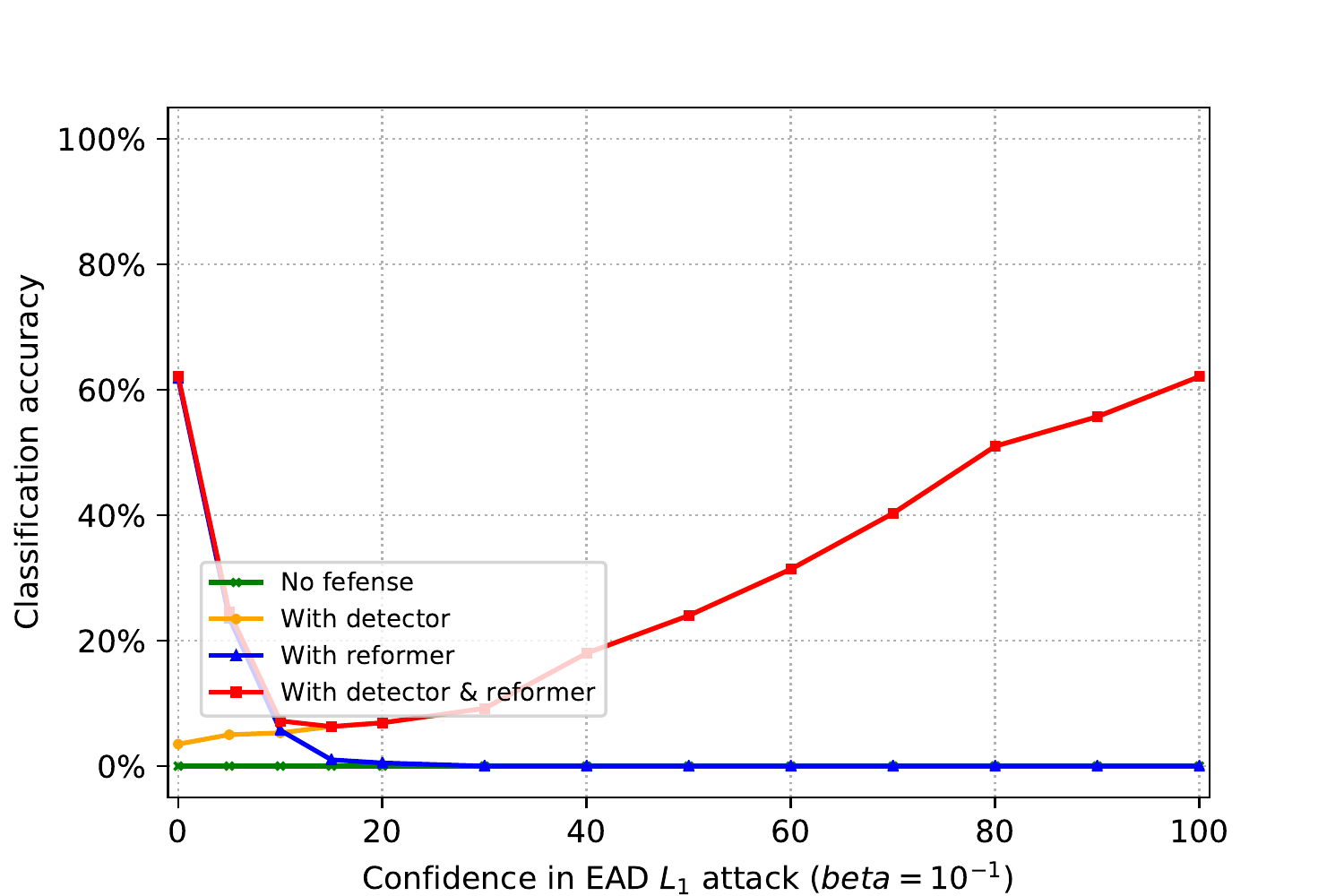}}
	\subfloat[EN decision rule $\beta=10^{-1}$]{
		\label{fig:CIFAR_EN_e-1_256}
		\includegraphics[scale=0.5]{CIFAR_default//defense_performance_CIFAR_EN_e-1}}
	\caption{EAD attacks on robust MagNet under different $\beta$ and different decision rules on CIFAR-10 with varying confidence. The number of filters in a auto-encoder's convolution layer is increased to 256.}
	\label{fig:CIFAR_256}
\end{figure*}

\begin{figure*}
	\centering
	\subfloat[mean squared error]{
		\label{fig:MAGNET_MNIST}
		\includegraphics[scale=0.5]{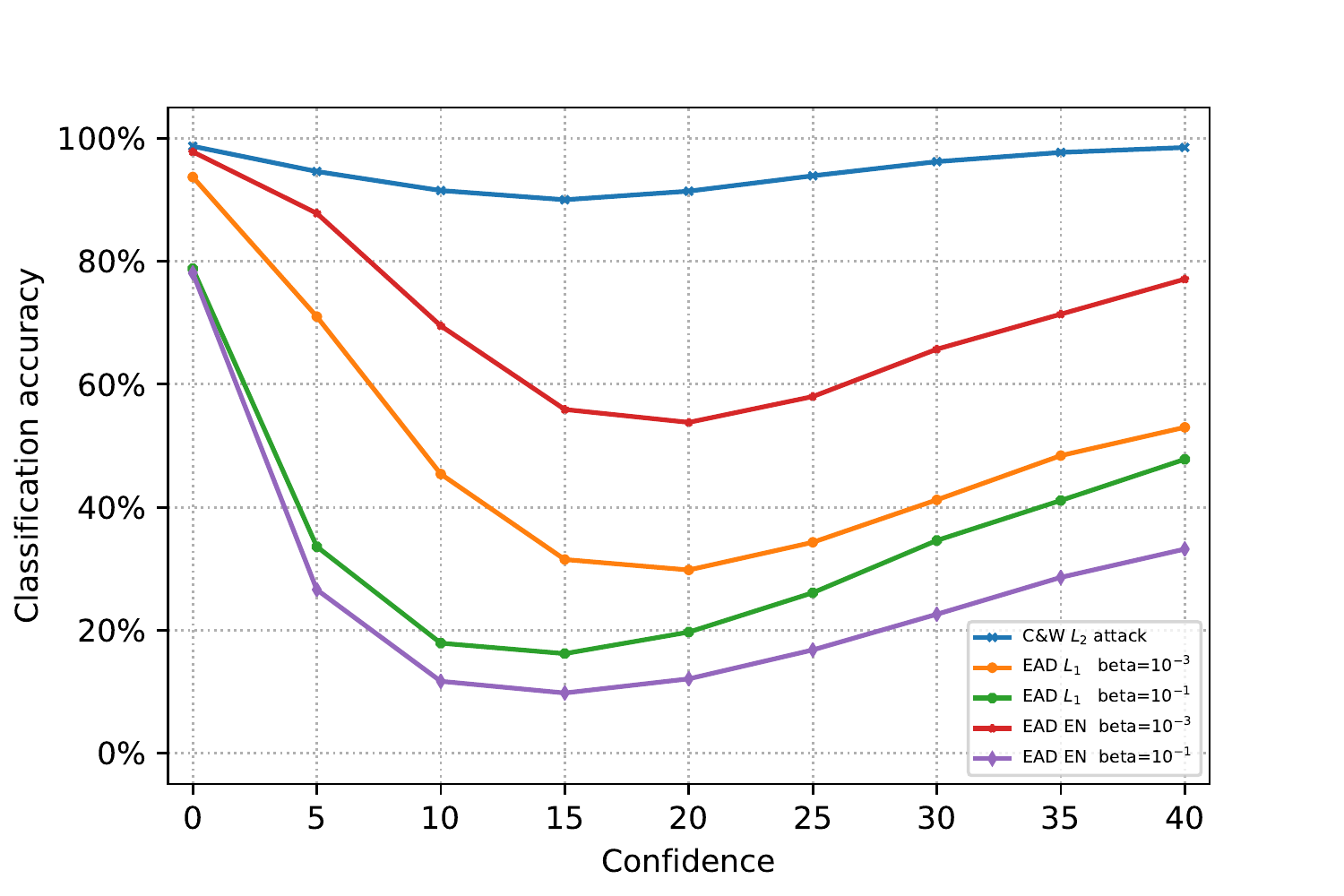}}
	\subfloat[mean absolute error]{
		\label{fig:L1_MNIST}
		\includegraphics[scale=0.5]{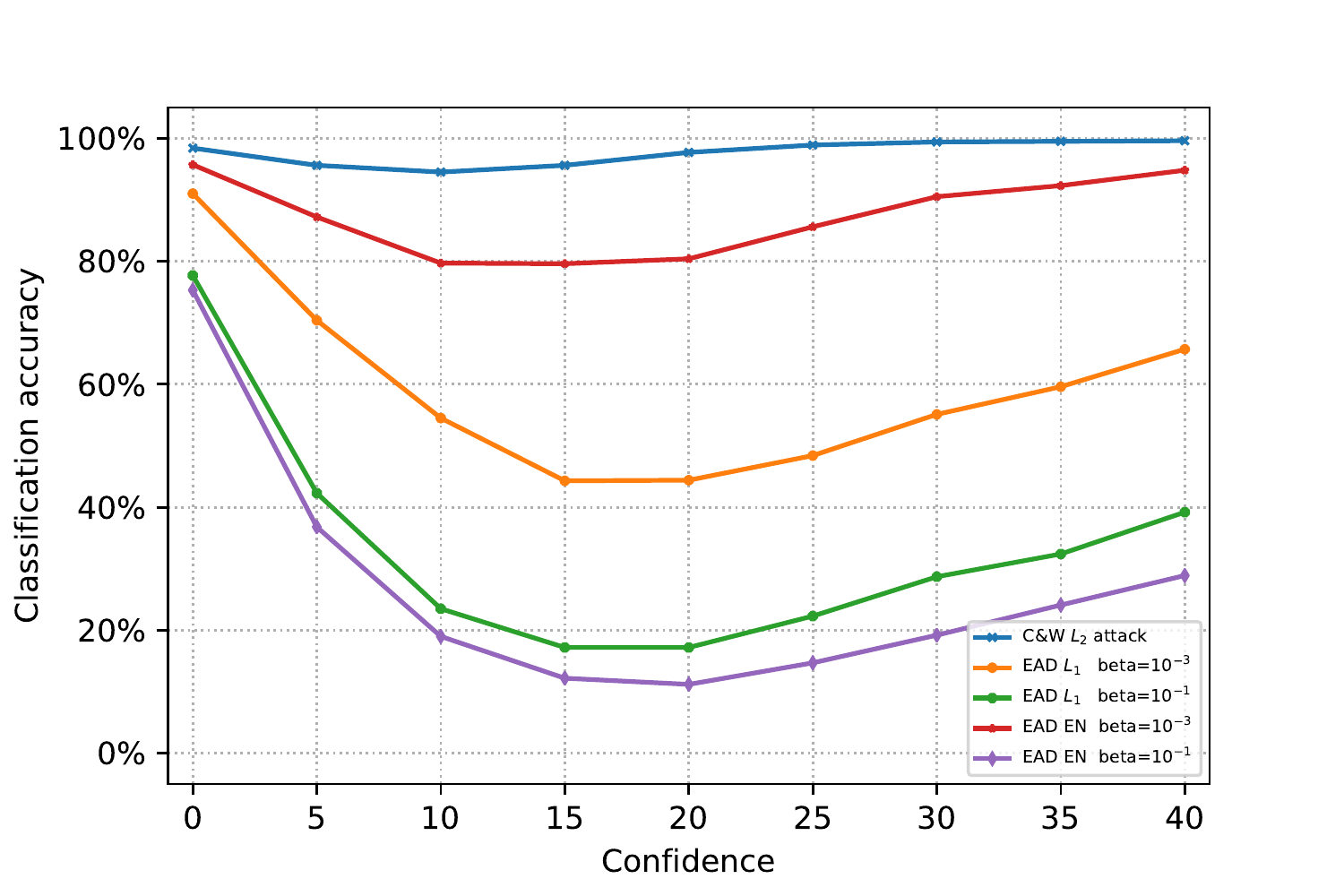}}
	\caption{Defense comparison when training the auto-encoders with different loss functions on MNIST using default MagNet.}
	\label{fig:L1vsL2_MNIST}
\end{figure*}

\begin{figure*}
	\centering
	\subfloat[mean squared error]{
		\label{fig:MAGNET_CIFAR}
		\includegraphics[scale=0.5]{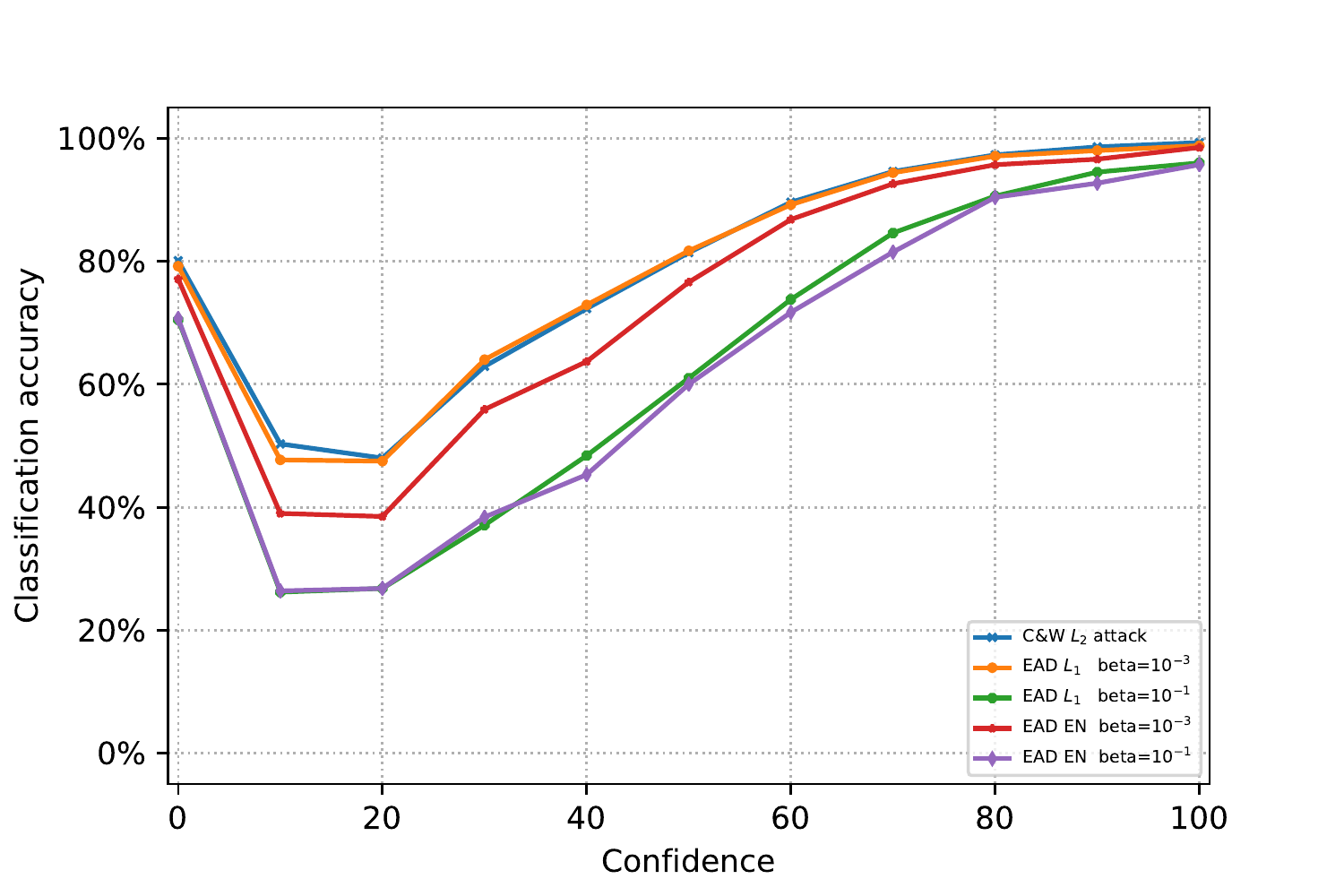}}
	\subfloat[mean absolute error]{
		\label{fig:L1_CIFAR}
		\includegraphics[scale=0.5]{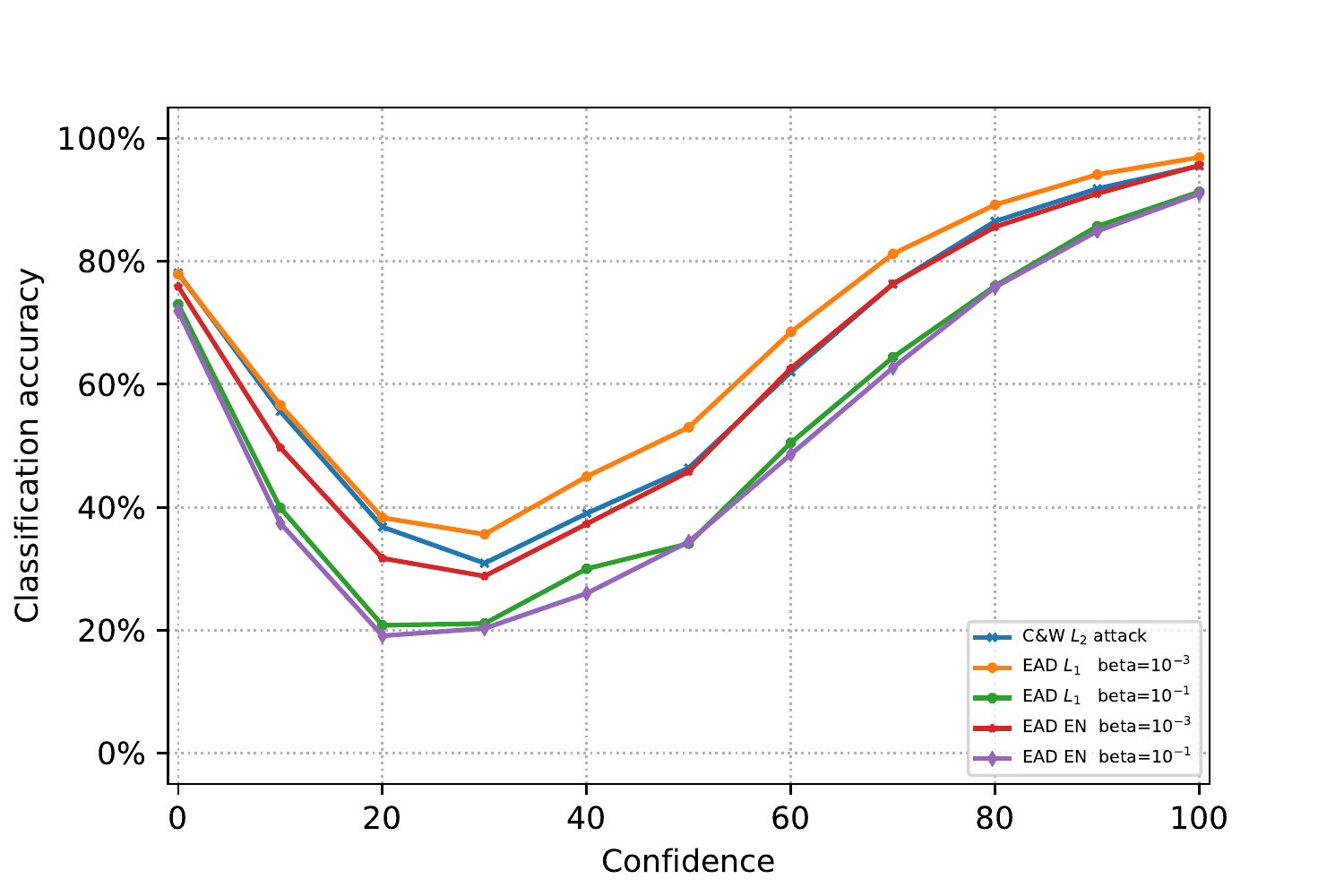}}
	\caption{Defense comparison when training the auto-encoders with different loss functions on CIFAR-10  using default MagNet.}
	\label{fig:L1vsL2_CIFAR}
\end{figure*}

\begin{itemize}
	\item 	Figure \ref{fig:MNIST} shows the defense performance of 4 different variants of MagNet against C\&W attack   on MNIST. 
	\item	Figure \ref{fig:CIFAR} shows the defense performance of 2 different variants of MagNet against C\&W attack on CIFAR-10.
	\item   Figure \ref{fig:MNIST_default} shows the defense performance of the default MagNet against EAD attacks with different values of the $L_1$ regularization parameter $\beta$ and different decision rules on MNIST.
	\item   Figure \ref{fig:CIFAR_default} shows the defense performance of the default MagNet against EAD attacks with different values of the $L_1$ regularization parameter $\beta$ and different decision rules on CIFAR-10.
	\item Figure \ref{fig:MNIST_default_JSD}	shows the defense performance of the robust MagNet (by including 2 additional JSD detectors) against EAD attacks with different values of the $L_1$ regularization parameter $\beta$ and different decision rules on MNIST.
	\item Figure \ref{fig:MNIST_256}	shows the defense performance of the robust MagNet (by increasing the number of filters to be 256) against EAD attacks with different values of the $L_1$ regularization parameter $\beta$ and different decision rules on MNIST.
	\item Figure \ref{fig:MNIST_256_JSD}	shows the defense performance of the robust MagNet (by increasing the number of filters to be 256 and including 2 additional JSD detectors ) against EAD attacks with different values of the $L_1$ regularization parameter $\beta$ and different decision rules on MNIST.	
	\item Figure \ref{fig:CIFAR_256}	shows the defense performance of the robust MagNet (by increasing the number of filters to be 256) against EAD attacks with different values of the $L_1$ regularization parameter $\beta$ and different decision rules on CIFAR-10.	
	\item Figure \ref{fig:L1vsL2_MNIST} shows the defense performance of the default MagNet using either $L_1$ (mean absolute error) or $L_2$ (mean squared error) reconstruction loss when training the auto-encoder on MNIST.  
	\item Figure \ref{fig:L1vsL2_CIFAR} shows the defense performance of the default MagNet using either $L_1$ (mean absolute error) or $L_2$ (mean squared error) reconstruction loss when training the auto-encoder on CIFAR-10.	
\end{itemize}

%


\end{document}